%% file: iclr2027_conference.tex
\documentclass{article} %
\usepackage{iclr2027_conference,times}

\input{math_commands.tex}

\usepackage{hyperref}
\usepackage{url}
\usepackage{graphicx}
\usepackage{booktabs}
\usepackage{subcaption}
\usepackage{wrapfig}
\usepackage{float}   %
\usepackage{caption}
\usepackage[T1]{fontenc}
\usepackage{xspace}

\title{SpeedrunBench: Challenging LLM Agents with Video Game Speedrunning}

\author{Yoshinari Fujinuma, Keisuke Kamahori, Ryuto Koike, Abdelrahman Madkour, \\ {\bf Varun Prashant Gangal, Monty Bichouna, Martyna Markiewicz, Shivani Jain,} \\ {\bf Duncan Curtis, Rebecca Qian, Anand Kannappan} \\
\includegraphics[width=0.3cm]{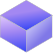} Patronus AI \; University of Washington \; Institute of Science Tokyo \\
\small{\texttt{\{yoshinari.fujinuma,varun.gangal,abdelrahman.madkour,anand\}@patronus.ai}}\\
\small{\texttt{kamahori@cs.washington.edu} \; \texttt{ryuto.koike@nlp.comp.isct.ac.jp}}}

\newcommand{\datasetname}{\textsc{SpeedrunBench}\xspace}

\iclrfinalcopy %
\usepackage{caption}
\begin{document}

\maketitle

\begin{abstract}

 Frontier LLM agents have been shown to be capable of solving increasingly complex tasks for which humans have measurable solutions. This begs the pertinent question of whether LLM agents can go beyond what humans have already solved. The ability to develop sophisticated strategies to tackle consequential problems becomes paramount as well-trodden, human-developed solutions become insufficient for problems for which we lack context or enough training data. We study agents' capability of such strategy formation through the communal practice of video game speedrunning. In speedrunning, practitioners compete to find the fastest way to complete a video game under certain conditions, and in so doing uncovering interesting unorthodox play styles that require a thorough understanding and mastery of the underlying game mechanics. We introduce {\bf \datasetname}, a benchmark that evaluates frontier LLM agents across 9 different games. To perform well in this benchmark, agents must repeatedly improve their strategy, reflect on their performance, exploit their gained knowledge, and reason across a long-horizon of actions to improve on an increasingly difficult problem: being faster than themselves and everyone else. Our experiments show that while frontier agents approach human world records in simple platformer games, they remain behind human performance on longer, more complex games under practical budgets. These results suggest that \datasetname is a useful testbed for studying agents' strategy formation capabilities as well as being a saturation-resistant evaluation measure, as there is almost always a faster completion time waiting to be discovered.\footnote{Code: \url{https://github.com/patronus-ai/speedrun_bench}}

\end{abstract}

\section{Introduction}

Through the rich history of AI research, games have served as a common testbed for developing \citep{samuel1959checkers,tesauro1995tdgammon}, testing \citep{bellemare2013ale,machado2018ale} and measuring \citep{mnih2015dqn,badia2020agent57} various techniques, from search \citep{campbell2002deepblue,silver2016alphago} to state representation \citep{mnih2015dqn,hafner2023dreamerv3}, and reinforcement learning \citep{silver2017alphagozero,vinyals2019alphastar,berner2019dota}. Games serve multiple desiderata for AI researchers: they are cheap, reproducible simulations with a clear ground truth \citep{bellemare2013ale,kurach2020football,kuttler2020nethack}, have  human baselines \citep{mnih2015dqn,toromanoff2019superhuman}, require multiple different skills across various modalities \citep{bellemare2013ale,hafner2022crafter,reed2022gato}, and many games require long-horizon reasoning under partial observability \citep{vinyals2017sc2le,berner2019dota,kuttler2020nethack}. In being such a fruitful playground for AI development, games have been a driver of progress in their own right: developing AI to outplay humans in complex games has produced techniques that found uses elsewhere in AI research \citep{silver2018alphazero,schrittwieser2020muzero,mandhane2022muzero}.

\begin{figure}[t]
\begin{center}
\includegraphics[width=0.95\linewidth]{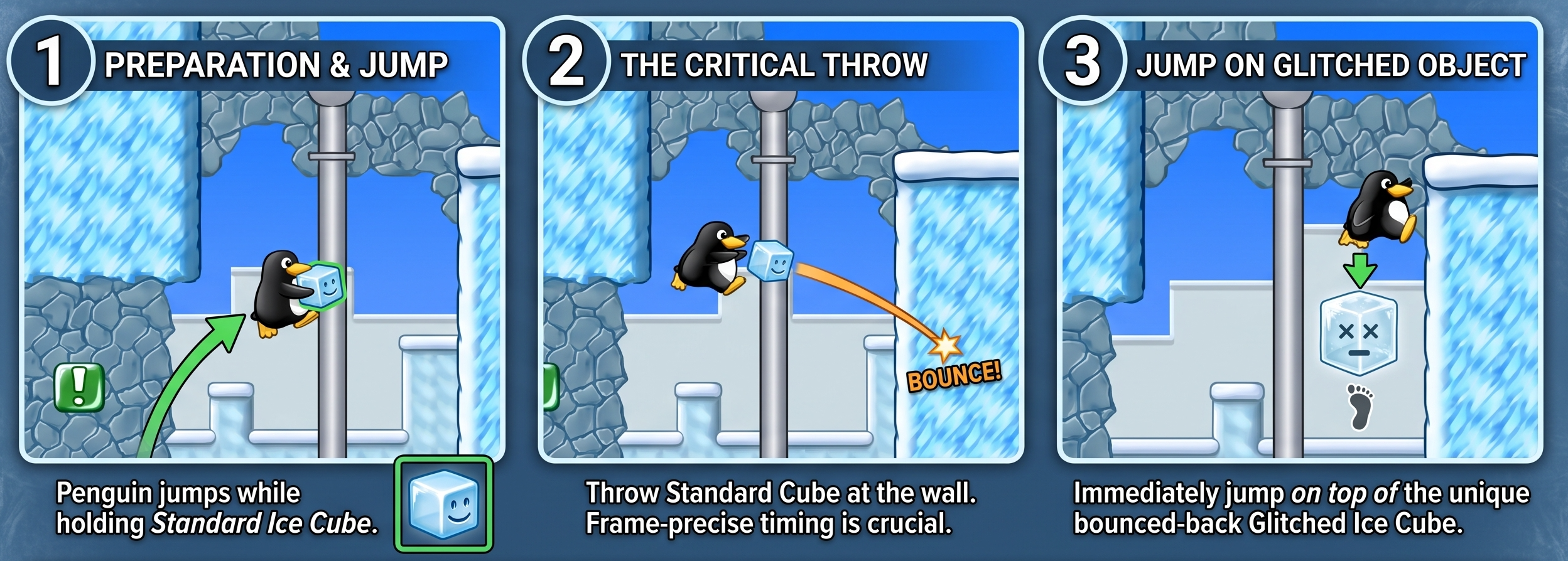}
\end{center}
\caption{One example of a complex technique necessary for speedrunning SuperTux, an open-source 2D scroll action game. This technique is unnecessary to complete the game itself, but for speedrunning, this is critical to cut down the frame to reach the goal.  }

\label{fig:speedrun_technique}
\end{figure}

It is no surprise, then, that games are once again an epicenter of AI evaluation. Their desirable qualities still hold as the landscape has shifted towards LLM agents, and prior work has leveraged many game genres \citep{cote2018textworld,hausknecht2020jericho,paglieri2025balrog,wu2024smartplay,hu2026lmgamebench, guertler2025textarena,zhang2025videogamebench}, and purpose-built game environments \citep{arcprize2026arcagi3} in assessing model performance. However, benchmarking for game-playing agents has so far focused on \emph{completion}: progress toward, or attainment of, the end of the game \citep{paglieri2025balrog,zhang2025videogamebench,hu2026lmgamebench}. While this remains a difficult task and, at the time of writing, a challenging one for frontier models \citep{zhang2025videogamebench,arcprize2026arcagi3}, it does not provide researchers with an evergreen lens by which to continue to push and analyze model capabilities. No finite benchmark is immune to saturation, but some are more resistant, and we posit that \emph{speedrunning} is a better fit for the current landscape of LLM agents evaluation. It is one thing for an agent to finish a complex game; it is another to be the fastest in the world to do so. Speedrunning is a community practice whereby players attempt to complete a game under specified criteria (e.g., with or without glitches) in the fastest possible time \citep{scullyblaker2014practiced}. Players compete to beat the record, and discover strategies the game's designers never anticipated (e.g., Figure~\ref{fig:speedrun_technique}). These strategies arise from the underlying dynamics of the game systems and speedrunners develop a more complete model of a game's mechanics than the people that built it \citep{scullyblaker2018museum}. Game developers have responded in the past to speedrun records by patching their games, by deliberately preserving exploits, and by consulting speedrunners during development, with several studios reporting that speedrun analysis shaped design decisions \citep{scullyblaker2018museum,francis2016speedrunners}. Thus, speedrunning is a proven mechanism by which humans have developed a cooperative culture of strategy formulation to optimize a shared goal (finding the fastest possible route to the end) that had downstream consequences for other objectives.

As prior work has noted \citep{vinyals2019alphastar,berner2019dota}, games require long-horizon reasoning. Speedrunning stresses this further, as one must reason beyond what constitutes an acceptable solution to finding the fastest one. The state and action space of a game, even if held to be deterministic under a fixed seed, is finite but exponential in the machine's memory \citep{hearn2009games}, and it grows with every frame a run requires. If we take Super Mario Bros.\ for instance it has been shown to be NP-hard \citep{aloupis2015nintendo} to complete, and speedrunning requires an additional optimality constraint, which makes it a more complex problem \citep{krentel1988optimization}. Proving optimality in this context practically requires either a structural lower bound (e.g., the run must contain at least $n$ unavoidable frames) or exhaustion of the reachable space. This the source of speedrunning's resistance to saturation, a game can be observed to be unsaturated whenever a new run exceeds the known best time, but saturation itself cannot be proven without solving the optimization problem outright. Thus, in practical terms, we should expect LLM agents to find speedrunning substantially harder than completion, as the task does not saturate when the agent does a successful valid run, every new record is an upper bound upon which to improve, and this improvement requires counterfactual reasoning about alternative routes and mechanics that no single episode reveals.
In the same way that beating the world's experts at Go and Chess led to ground-breaking research in reinforcement learning \citep{silver2016alphago,silver2018alphazero}, we believe speedrunning can invoke similar innovation in LLM agents. 

\begin{figure}[t]
\begin{center}
\includegraphics[width=\linewidth]{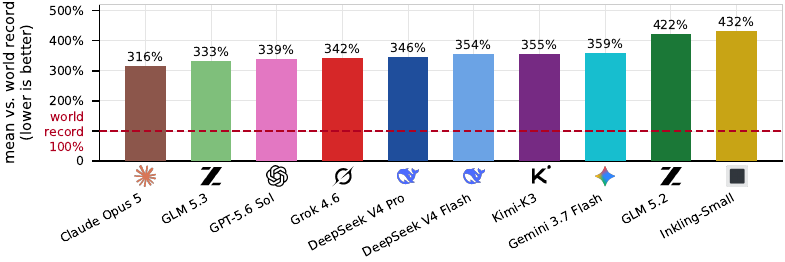}
\end{center}
\caption{With the world record as an anchor (100\%), we average the time the model took to reach the goal across the games and compare on how much longer it took than the world record as percentage (lower is better). Detailed game-level breakdown is in Figure~\ref{fig:pergame}.}
\label{fig:best-vs-record}
\end{figure}

In this paper we introduce \datasetname, a benchmark for evaluating LLM agents on speedrunning games, covering 10+ frontier agents and 9 different games.
We evaluate agents in three settings: an \textsc{online} computer-use setting and two new offline settings. In \textsc{online}, an agent observes the current frame and selects the next action or short sequence of actions while the game runs. In both offline settings, the agent repeatedly edits and replays a complete frame-indexed action trace for a fixed number of iterations. 
\textsc{offline-seed} begins with a goal-reaching reference trajectory and focuses on improving it, whereas \textsc{offline-scratch} requires the agent to first construct a successful trajectory. Each setting highlights a different capability: \textsc{online} requires multimodal real-time perception, \textsc{offline-scratch} requires route discovery, as the agent begins without a successful trajectory, and \textsc{offline-seed} focuses on improving an existing solution, as the agent begins with a successful reference trajectory. All settings demand long-horizon coherence, since the agent has to keep improving a route it already holds, over many turns, without breaking it.
Our experiments (Figure~\ref{fig:best-vs-record}) show that while recent frontier agents approach human world records on simpler platformer games, they remain slower on long, complex games such as role-playing, racing, and strategy games, given a fixed budget. Even when given a successful trace as a starting point, agents achieve only marginal improvements, showing a gap between finding a valid solution and improving it.

\section{Related Work}

\paragraph{Benchmarks for LLM Agents Seeing \& Playing Games} Games have been serving as evaluation benchmarks for machine learning (ML) models well before LLMs became popular, from text adventures~\citep{cote2018textworld} to 2D
navigation~\citep{multinet_v2_preview_technical_report_2026} and abstract reasoning
tasks~\citep{arcprize2026arcagi3}. Games were also the setting for some of
the field's most visible results before LLMs, from Atari from
pixels~\citep{mnih2015dqn} to superhuman play in Go~\citep{silver2016alphago} and
StarCraft~II~\citep{vinyals2019alphastar}.
Recent benchmarks extend the setting to let frontier LLMs play games
directly. LMGame-Bench~\citep{hu2026lmgamebench} supplies perception and memory scaffolds,
standardizes prompting, and screens for training data contamination, while
VideoGameBench~\citep{zhang2025videogamebench} asks whether
vision-language models (VLMs) can complete popular titles at all. 
\citet{taesiri2024glitchbenchlargemultimodalmodels} pairs images with
questions to evaluate whether a VLM can identify what is wrong in a rendered screen. Individual
titles have also served as milestones in their own right, with Gemini 2.5 Pro clearing Pok\'emon\footnote{Pok\'emon is a trademark of Nintendo Co., Ltd., Creatures Inc., and Game Freak Inc.}
in 406.5 hours over two attempts~\citep{comanici2025gemini25pushingfrontier}, and
\citet{hu2024pokellmonhumanparityagentpokemon} testing LLM capability on Pok\'emon battles
specifically. 
Minecraft is the main setting where the objective is not merely completing something:
MineDojo~\citep{fan2022minedojo} and Voyager~\citep{wang2024voyager} score an agent on how much it
acquires in an open world, by items obtained and skills learned. 
In contrast, \datasetname{} scores how
quickly it was completed, which continues to separate models after completion is reached by
challenging them to learn and discover speedrun-related skills.

\paragraph{Speedrunning as a Task} 
Outside games, related work studies speedrunning LLM training~\citep{zhao2025automated} and using agents to optimize LLM serving systems~\citep{kamahori2026vibeserveaiagentsbuild}.
The closest work to ours is the Pokeagent Challenge from
NeurIPS 2025~\citep{karten2026pokeagentchallengecompetitivelongcontext}, in which speedrunning is
one of the tracks, followed by~\citet{karten2026continualharnessonlineadaptation} on online
adaptation. Both evaluate an agent that observes a game frame and predicts the next action or a
short chunk of actions, and both focus on a single title: Pok\'emon. \datasetname{} differs in
scope and in setting: our suite spans several genres, and beyond the online
setting it evaluates an offline one in which the agent edits a complete frame-indexed trajectory,
replays it, and scores on the resulting frame count, which turns the task from finding a
working trajectory into improving one that already works.

\section{Speedrunning with an LLM Agent}
We first review what agent capabilities are required to perform well on speedrunning (\S\ref{sec:capabilities}), then introduce different setups for LLM agents speedrunning~(\S\ref{sec:online_offline}).
Using \datasetname, we explore the following research questions:
\begin{itemize}
    \item  RQ1: Is \datasetname{} challenging for modern LLM agents? Under what setup can a modern LLM-based agent speedrun, and under what setup can it not?
    \item  RQ2: Can LLM agents optimize their gameplay with an autoresearch-like loop? Even for a near-optimal Tool-Assisted Speedrun (TAS) run?
    \item  RQ3: What are the critical features for an LLM agent to speedrun? 
\end{itemize}

\subsection{Agent Capabilities Measured in \datasetname}
\label{sec:capabilities}

{\bf Exploration}: The agent needs to ``think outside the box'' to explore new paths and approaches to minimize the frames to the goal. This capability is tested especially in an  \textsc{offline-scratch} setup where given seed gameplay trajectories are not necessarily fast gameplays. 

{\bf Discovery}: For example, a glitch in Super Mario Brothers was discovered 40 years after the original game is released by a speedrunner~\citep{polygon2026smbace}. %

{\bf Long-horizon Coherence}: Traces run large number of actions, and a single wrong edit anywhere in one costs the whole run. The agent has to keep
what already works while changing part of it, and to do so over hundreds of iterations.

\begin{figure}[t]
\begin{center}
\includegraphics[width=\linewidth]{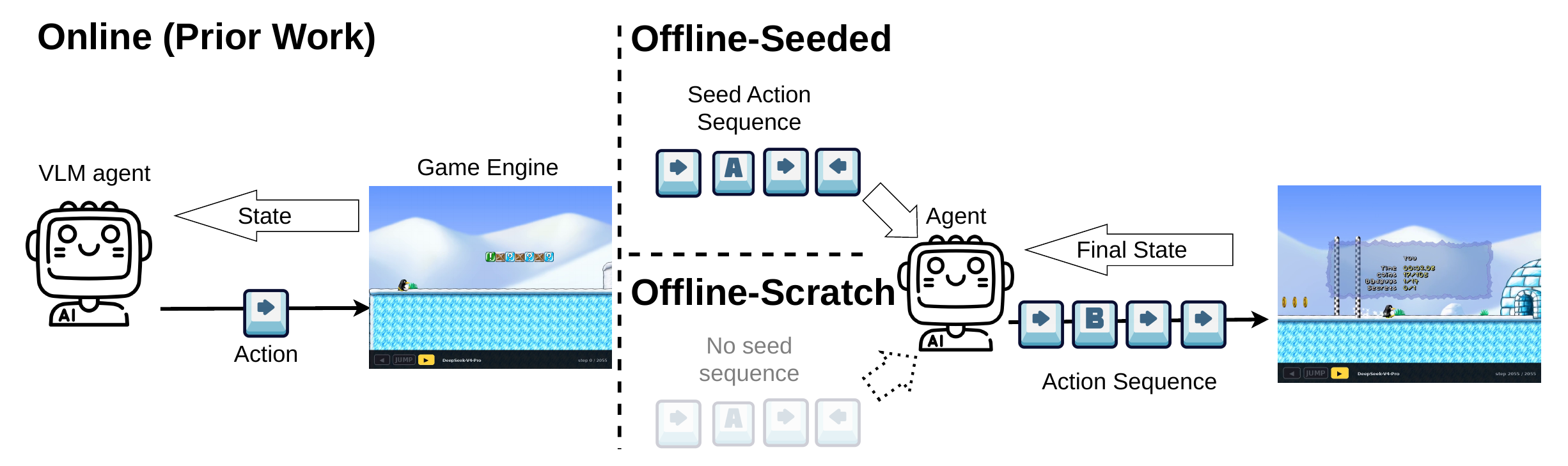}
\end{center}
\caption{Overview of the different settings explored in \datasetname: \textsc{online}, \textsc{offline-seed}, and \textsc{offline-scratch}. \textsc{online} setup predicts next action (or small action chunks) from a current state, whereas \textsc{offline} setup predicts the full action trace from the start to the goal. 
}

\label{fig:overview}
\end{figure}

\subsection{Online vs. Offline Setting for Speedrunning}
\label{sec:online_offline}
We define a game as a deterministic environment with state space $\mathcal S$
and a finite action space $\mathcal A$, the set of controller inputs available on that console.
Play begins from a fixed initial state $s_0 \in \mathcal S$, and each action
advances the game by one frame, so a sequence of actions in $\mathcal A$ fully determines a run.
We explore three settings per game: {\it Online}, {\it Offline-Scratch}, {\it Offline-Seed} for measuring different capabilities of each model (Figure~\ref{fig:overview}).
We further explain these three different setups in the following paragraphs.

\paragraph{Online (Action prediction from screenshots)}

Following prior work~\citep{hu2026lmgamebench, zhang2025videogamebench}, given a current state $s$ of a game, the agent decides the next action $a \in \mathcal A$.
A limited subset of frontier models (Opus 5, GPT-5.6-Sol, Grok 4.6) are capable of completing the game, but this setup only measures one aspect agent's capability such as multimodal understanding. Since this setup can produce a baseline seed reference trajectory, we optimize it in the offline or trajectory editing setting. 

\paragraph{Offline (Trace Editing)}
This setting mimics human speedrunning by playing the game over and over again throughout the decades. 
A run is represented as a frame-indexed action trace: a sequence of $n$ actions
$\{a_i\}_{i=1}^{n} \in \mathcal A$ covering an entire action sequence, from start to goal. 
On each turn the agent submits an edited trace, evaluated, and the agent receives the final state $s_{\text{final}}$ of the trace such as 1) outcome $o \in  \{\text{true}, \text{false}\}$ on whether it reached the goal, 2) the number of frames $f$ to reach the goal if $o=\text{true}$, or the maximum progress the trace made if $o=\text{false}$ i.e., the edited trace no longer reached the goal.
The agent repeats this for a fixed number of turns (200 unless specified). 
This can be regarded as action chunking i.e., predicting next $k$ actions~\citep{Zhao-RSS-23} with an extremely long chunk. 
We explore the following two variant settings:

    \noindent\textbf{\textsc{offline-scratch}:} The agent has to come up with a valid action trave from scratch i.e., without a reference trace. This is harder than seeded setting due to a working trace not given in advance. 
    \noindent\textbf{\textsc{offline-seed}:} The agent is given a seed reference trajectory to optimize from. 
    Seed reference trajectory are either from humans, scripted by breadth-first-search (BFS), from online setup, which are all far from the best run for each game except for \S\ref{sec:tas_run} where we analyze the effect of number of times calling the scorer.

Note that vision inputs are not required under \textsc{offline} settings, and therefore, more open-weight models can be evaluated under these settings. This is because, especially for simple games, no vision inputs are needed, smaller open-weight models can optimize. 

\subsection{Auto-research Loop-based Optimization}

For \textsc{offline} setup, we use a self-improving loop inspired by auto-research~\citep{karpathy2026autoresearch} to mimic humans speedrunning by playing the game multiple times, see where time was lost, change one thing, and run it again, for as long as the route keeps giving. The \textsc{offline}
settings reproduce that structure. At each iteration the agent selects its current best trace and the record of what it has already tried, proposes an edit, and receives a verdict. 
We fix both the number of iterations and how many candidate traces the agent may evaluate within one iteration before committing to the edit it submits. These hyperparameters are left fixed across models.
(\S\ref{sec:budget}).

\begin{table}[t]
\caption{Games explored in \datasetname, spanning several platforms and genres.}
\label{tab:games}
\begin{center}
\begin{tabular}{lll}
\toprule
\bf GAME & \bf PLATFORM & \bf GENRE \\
\midrule
Super Mario Brothers        & NES              & Platformer \\
Super Mario Land            & Game Boy         & Platformer \\
Mario Kart 64               & Nintendo 64      & Racing     \\
Pok{\'e}mon Blue/Red            & Game Boy         & RPG        \\
Civilization I              & MS-DOS/SNES      & Strategy   \\
Astray                      & Browser          & Puzzle     \\
SuperTux                    & C++/Browser      & Platformer   \\
Tuxemon                     & Python/Browser             & RPG   \\
SuperTuxKart               & C++/Browser             & Racing   \\
\bottomrule
\end{tabular}
\end{center}
\end{table}

\section{Experiments}

\subsection{Games} Table~\ref{tab:games} lists the nine titles in \datasetname, spanning six
genres and multiple consoles, and assembled along two axes. The first is
genre, since the capability required for speedrunning depends on each game: a platformer requires frame-precise movement, an RPG rewards routing across a long horizon, and a puzzle game
rewards search over a discrete state space.
Alongside
widely known commercial titles we include open-source games of the same genre, so that each of
the three most popular and documented genres has a counterpart whose routes and records are far less
likely to appear in a pretraining corpus.\footnote{Commercial titles listed here are named just to identify evaluation setting. Users must obtain lawful copies themselves and nothing here grants rights beyond those already allowed by relevant licenses, TOS, \& law.
}

\noindent\textbf{Super Mario Brothers, Super Mario Land, and SuperTux~\citep{supertux2021}} are 2D side-scrolling platformers, which demand precise movement and route discovery toward a level
exit. The goal is finishing Level 1-1. Both Mario titles have decades of documented routes and world records, which supplies an external speedrun reference, and they are also the titles which are most likely to have been exposed to LLMs during pretraining. SuperTux is the open-source counterpart that makes the extent of
exposure measurable: it has similar gameplay as Mario for the agent while carrying none of the same pretraining exposure. We use its ``Welcome to Antarctica'' level, referred to as SuperTux 1-1 throughout.

\noindent\textbf{Pok\'emon Blue/Red and Tuxemon~\citep{tuxemon}} are turn-based RPGs, and the longest-horizon tasks in this benchmark. A run reaches the first gym badge only after 13 milestones, and the frame counts involved are one to two orders of magnitude larger than in the platformers, which stresses long-context tracking. Tuxemon is the open-source counterpart, matched in genre and structure but estimated to be less representative in the pretraining data in general.

\noindent\textbf{Mario Kart 64 and SuperTuxKart~\citep{pringle2021supertuxkart}} are racing games, where the goal is lap time. We use Black Forest track for SuperTuxKart and Wario Stadium for Mario Kart 64.

\noindent\textbf{Astray} is a browser-based maze puzzle.
We use the implementation from~\citet{ouyang2026gameworldstandardizedverifiableevaluation}.

\noindent\textbf{Civilization~I} is a strategy game. These broaden genre coverage beyond the platformer and RPG settings that dominate prior game benchmarks.

\subsection{Setup}

\noindent\textbf{Models:} We use frontier models both from proprietary models such as Claude Opus 5~\citep{anthropic2026opus5}, GPT-5.6-sol~\citep{openai2026gpt56}, Grok 4.6~\citep{spacexai2026grok46}, Gemini-3.7-Flash~\citep{deepmind2026gemini37flash}, and open-weight models such as Kimi-K3~\citep{kimiteam2026kimik3openfrontier}, GLM 5.3~\citep{zai2026glm53}, GLM 5.2~\citep{zai2026glm52}, DeepSeek-V4-Pro-0813~\citep{deepseek2026pro0813}, DeepSeek-V4-Flash-0731~\citep{deepseek2026flash0731}, Inkling-Small~\citep{thinkingmachines2026inklingsmall}.
All models ran with Opencode~\citep{opencode_software} for experimenting on the same harness. Reasoning effort is set to the maximum available for every model in this paper. 

\noindent\textbf{Reference Times:} Every world record we compare against is a human real-time attack (RTA) from verified
runs~\citep{sinned2025pkmnredblue,eip25_2026_sml1,dylcat2025supertux,pringle2021supertuxkart,abney3172026mk64}.

\noindent\textbf{Seed Reference Trajectory for \textsc{offline-seed}:} For traces to seed the agent in \textsc{offline-seed}, GPT-5.6-sol/Opus 5 online run, scripted, and human traces are used. All the seed trajectories are far from the optimal speedrun play for leaving the space for agents optimize their play.

\subsection{Online Play Results}
\begin{figure}[h]
\begin{center}
\begin{subfigure}[b]{0.49\linewidth}
    \centering
    \includegraphics{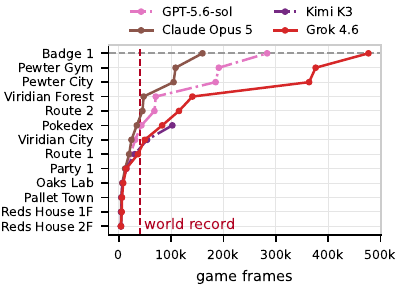}
    \caption{\textsc{online}: milestones reached over game frames.}
    \label{fig:pokemon-only}
\end{subfigure}
\hfill
\begin{subfigure}[b]{0.49\linewidth}
    \centering
    \includegraphics{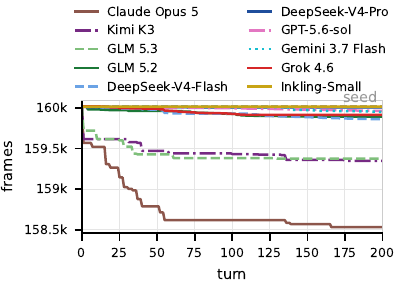}
    \caption{\textsc{offline-seed}: frames to the first badge.}
    \label{fig:pokemon-offline}
\end{subfigure}
\end{center}
\caption{Pokemon Blue results. \textbf{(a)} \textsc{online} run on models reaching Pokemon Blue
milestones. Opus 5 reaches Badge 1 in 160,021 frames, whereas the human world record is $\sim$40,435 frames~\citep{sinned2025pkmnredblue}: the record is $\sim4\times$ faster than the best model result. Kimi K3 did not reach Route 2. 
\textbf{(b)} \textsc{offline-seed} speedrun until the first badge, starting from the
reference trajectory of Opus 5's online run and given 200 iterations. }
\label{fig:pokemon-pair}
\end{figure}

Figure~\ref{fig:pokemon-only} shows \textsc{online} Pok\'emon play. Only the three
vision-capable proprietary models reach the first badge at all, while Kimi K3 stalls at the prior point in the game where Reds(Ash) gets the Pok\'edex from Prof. Oak.
All runs are still far off the from the human world record of $\sim$40,435: Opus 5 takes
160,021 frames, GPT-5.6-sol 283,474, and Grok 4.6 477,140, so
the best agent is roughly $4\times$ slower than the record and the worst nearly $12\times$. 
Under a completion metric, these three runs are indistinguishable, which shows the edge of \datasetname{} at providing finer-grained discrimination.

\subsection{Offline Speedrunning Results}

As shown in Figure~\ref{fig:best-vs-record}, no agent reaches a world record under the current budget, and the best mean result is more than twice slower than the world record.
Figure~\ref{fig:pergame} in Appendix~\ref{app:pergame} gives the per-game breakdown. 
The closest results to the world record are the from-scratch SuperTux runs at 118\%, while SuperTuxKart result are in between 256\% and 261\% and every model on Pok\'emon sits near 395\%.
A completion metric scores all of these runs identically.

\begin{wraptable}{r}{0.5\linewidth}
\centering
\small
\vspace{-1.2em}
\begin{tabular}{lrr}
\toprule
\bf Model & \bf \begin{tabular}[b]{@{}r@{}}\textsc{offline-}\\\textsc{scratch}\end{tabular}
          & \bf \begin{tabular}[b]{@{}r@{}}\textsc{offline-}\\\textsc{seed}\end{tabular} \\
\midrule
\multicolumn{3}{l}{\emph{SuperTux 1-1}} \\
DeepSeek-V4-Pro   & \textbf{1{,}918} & 9{,}428 \\
Gemini 3.7 Flash  & \textbf{1{,}951} & 9{,}094 \\
DeepSeek-V4-Flash & \textbf{1{,}980} & 9{,}123 \\
GLM 5.3           & \textbf{2{,}087} & 8{,}412 \\
Kimi K3           & \textbf{2{,}432} & 9{,}280 \\
\midrule
\multicolumn{3}{l}{\emph{Super Mario Land 1-1}} \\
Claude Opus 5     & \textbf{2{,}835} & 5{,}385 \\
GPT-5.6-sol       & \textbf{3{,}271} & 5{,}896 \\
Grok 4.6          & \textbf{3{,}417} & 5{,}730 \\
GLM 5.3           & \textbf{4{,}113} & 5{,}534 \\
\bottomrule
\end{tabular}
\caption{Frames to the goal under \textsc{offline-scratch} and \textsc{offline-seed} for the
same model which found a trace to reach the goal (lower is better, with the faster in bold). 
SuperTux runs are seeded with a 9{,}498-frame trace, 
and Super Mario Land runs seed with a 6{,}017 trace. Results are largely influenced by the seed traces.}
\label{tab:scratch-vs-seed}
\vspace{-1em}
\end{wraptable}

In Table~\ref{tab:scratch-vs-seed}, every agent is run both from scratch and from a common 9498-frame seed produced by GPT-5.6-sol's \textsc{online} run.
All seven
\textsc{offline-seed} runs end within 12\% of the seed, between 8{,}412 and 9{,}428 frames. In \textsc{offline-scratch}, five of six models clear the level
3.8--4.9$\times$ faster
than the same model from the seed. GLM 5.2 is the exception: it 
reaches the goal only when handed the seed. \textsc{offline-scratch} thus separates agents that can construct a route from those that can merely refine one, whereas \textsc{offline-seed} shows how strongly a working but slow route through the game primes the agents.

On Super Mario Land 1-1, Claude Opus 5, GPT-5.6-sol, Grok 4.6 and GLM 5.3 cleared under
\textsc{offline-scratch}, at 2{,}835, 3{,}271, 3{,}417 and 4{,}113 frames, while the best
\textsc{offline-seed} run, from a 6{,}017-frame human seed, reached 5{,}145.
On SuperTuxKart, under \textsc{offline-seed} from a 168-second bot-generated lap, every model ends
between 165 and 168 seconds. On Pok\'emon, again without the scorer, no \textsc{offline-seed} run improves on the
seed by more than 1\% (Figure~\ref{fig:pokemon-offline}). No model reaches the record on any of
the three games.

\begin{figure}[h]
\begin{center}
\includegraphics{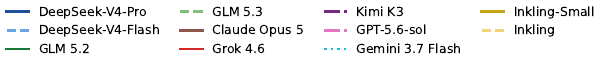}%

\begin{subfigure}[b]{0.49\linewidth}
    \centering
    \includegraphics{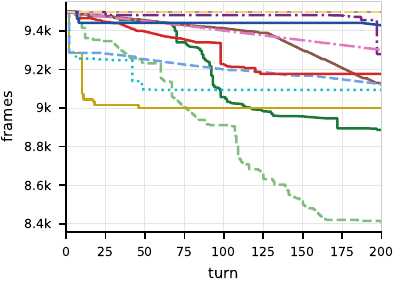}
    \caption{\textsc{offline-seed}, from a 9498-frame seed.}
    \label{fig:supertux-offline}
\end{subfigure}
\hspace{0.1em}
\begin{subfigure}[b]{0.49\linewidth}
    \centering
    \includegraphics{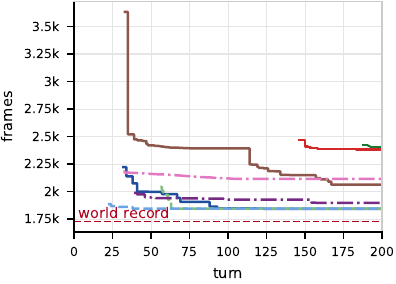}
    \caption{\textsc{offline-scratch}, no seed. }
    \label{fig:supertux-scratch}
\end{subfigure}
\end{center}
\caption{SuperTux 1-1 speedrun results with an agent backed with various LLMs. Lines in \textsc{offline-scratch} start from the turn an agent finds a valid trace to complete the goal.}
\label{fig:supertux-pair}
\end{figure}

\section{Analysis}

\subsection{Expanding the Evaluation Budget Per Turn}
\label{sec:budget}

\begin{figure}[h]
\begin{center}
\begin{subfigure}[b]{0.49\linewidth}
    \centering
    \includegraphics{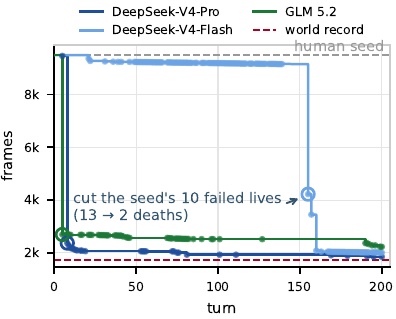}
    \caption{SuperTux 1-1, top 3 models under this setup.}
    \label{fig:supertux-offline-unlimited}
\end{subfigure}
\hfill
\begin{subfigure}[b]{0.49\linewidth}
    \centering
    \includegraphics{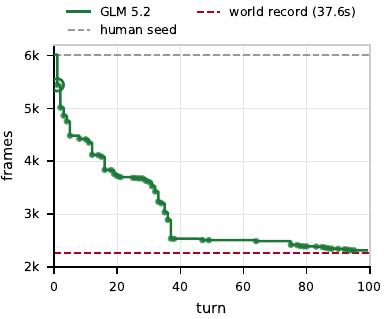}
    \caption{Super Mario Land 1-1 with in-turn scorer.}
    \label{fig:sml-offline-unlimited}
\end{subfigure}
\end{center}
\caption{\textsc{offline-seed} speedrun results with an agent allowed to evaluate up to 200 traces per turn. Super Mario Land world record is the 1-1 portion from~\cite{eip25_2026_sml1}. With this budget, GLM 5.2 on Super Mario Land \& DeepSeek-V4-Pro on SuperTux come close to the world record.}
\label{fig:offline-unlimited}
\end{figure}

\begin{figure}[t]
\begin{minipage}[t]{0.52\linewidth}
\vspace{0pt}
\centering
\includegraphics{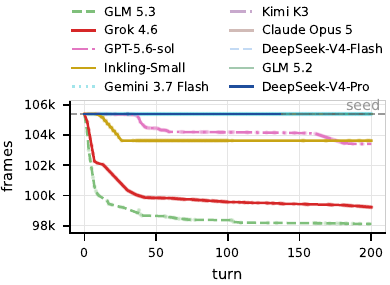}
\caption{\textsc{offline-seed} results on Tuxemon scored on frames to the first gym. All ten models start from the same 105{,}400-frame trace.}
\label{fig:tuxemon-offline}
\end{minipage}\hfill
\begin{minipage}[t]{0.44\linewidth}
\vspace{0pt}
\centering
\small
\begin{tabular}{lrr}
\toprule
\bf Model & \bf no scorer & \bf scorer \\
\midrule
\multicolumn{3}{l}{\emph{SuperTux 1-1}} \\
DeepSeek-V4-Pro   & 9{,}428 & \textbf{1{,}847} \\
Kimi-K3           & 9{,}280 & \textbf{1{,}930} \\
DeepSeek-V4-Flash & 9{,}123 & \textbf{2{,}009} \\
GLM 5.2           & 8{,}887 & \textbf{2{,}228} \\
\midrule
\multicolumn{3}{l}{\emph{Super Mario Land 1-1}} \\
GLM 5.2           & 5{,}643 & \textbf{2{,}314} \\
\bottomrule
\end{tabular}
\captionof{table}{Frames to the goal under \textsc{offline-seed} for the same model and seed,
without the in-turn scorer and with up to 200 evaluations per turn (lower is better).}
\label{tab:budget}
\end{minipage}
\end{figure}

In the experiments so far, the agent holds no in-turn scorer: it submits one trace per turn and gets one verdict. Here, we instead let the agent evaluate its own candidates, up to 200 per turn, and hold all else fixed. On SuperTux 1-1 this gives a controlled comparison, since the same models start from the same 9{,}498-frame seed and run for the same 200 turns in both settings
(Table~\ref{tab:budget}). With the scorer, every model escapes the seed and ends 4.0--5.1$\times$
below where it ended without it. DeepSeek-V4-Pro reaches 1{,}847 frames, within 7\% of the world
record and about two seconds slower, while DeepSeek-V4-Flash and GLM 5.2 remain 16\% and 29\% above
it (Figure~\ref{fig:supertux-offline-unlimited}). On Super Mario Land 1-1, GLM 5.2 with the scorer
comes within 3\% of the record after 220K evaluated traces over $\sim$114 hours
(Figure~\ref{fig:sml-offline-unlimited}), against 5{,}643 frames from the same human seed without
it. For these models, access to a scorer decides whether an agent escapes a non-optimal seeded trace, and
given enough evaluations the strongest agent on each short 2D level comes close to its world record.

\subsection{Case Study on Improving Near-Optimal Speedrun Trace}
\label{sec:tas_run}

The \textsc{offline-seed} settings above start from seed traces far off the optimum, which leaves large room to improve for an agent. 
We report two experiments that instead start from a near-optimal trace on Super Mario Brothers: the first asks an agent to improve one which is same as \textsc{offline-seed}
 By seeding instead from an 18834-frame full-game Super Mario Brothers trace~\citep{bisqwit2003smb} and given 200 turns,
GPT-5.6-sol returned a trace three frames faster. The second is merging two near-optimal traces\footnote{
\url{https://tasvideos.org/UserFiles/Info/638242850875487965} and \url{https://tasvideos.org/UserFiles/Info/638586818800241580}
} which saved 26 frames from the fastest trajectory out of the two. These results suggest seeding the optimization loop with multiple near-optimal trajectory has the potential to further improve those.

\subsection{Contamination from Pretraining: Case Study on Pokemon vs. Tuxemon}
\label{sec:contamination}

That famous titles are memorized in model parameters is well known: 
\cite{hu2026lmgamebench} reports game content memorised during pretraining, and
Gemini 2.5 Pro has been observed confusing items across Pok\'emon versions, expecting tea where the game requires soda~\citep{zhang2025gemini}. 
Hence, we add a controlled pair: Pok\'emon and Tuxemon holding genre and setting fixed while varying how much a title is likely to appear in a pretraining corpus. Both are turn-based RPGs, run under \textsc{offline-seed} with the same iteration budget and scored on frames to the first gym.
Figure~\ref{fig:tuxemon-offline} shows that lower popularity does not make the open-source title harder to improve on when compared to the Pok\'emon run is the one in Figure~\ref{fig:pokemon-offline}. No model improves on the Pok\'emon seed by more than 1\%, whereas GLM 5.3 and Grok 4.6 improve on the Tuxemon seed by 6.9\% and 5.9\%. Tuxemon is harder to make any progress on since every model on Pok\'emon submits at least one trace faster than the seed, while four of ten never submit a trace that reaches the Tuxemon gym. 
Across ten models run on both titles, the rank correlation observed b/w titles is slightly negative at $-0.18$. Hence, we see that results on a popular title are not predictive of performance on an unfamiliar one, which is consistent with popular titles being present in pretraining corpora.

\section{Conclusion}
We introduce \datasetname{}, a benchmark that measures agents' capability by assessing how fast an agent can finish a game and comparing it with the best known completion time. \datasetname{} evaluates agents across three settings: an \textsc{online} setting where agents get frame by frame screen shots and must take an action, and two \textsc{offline} settings where agents must optimize their game plan for the run either with a seeded trace or without. 
We show that our benchmark remains challenging for current models (RQ1), as no agent reaches human world record on any game under a practical budget, with the best performing agent being on average \emph{more than twice} as slow as the world record set by human players. Additionally we show that agents can improve their gameplay in an autoresearch-like loop (RQ2), even when the start seed is a near-optimal tool-assisted speedrun trace. However, giving the agents a slow seed as a starting point holds them back.
Agents perform better when they can construct the routes of their own accord, and expanding the agents' ability to evaluate a candidate with a larger per-turn budget, as seen with how close the platformer games came to the world records. 
We hope that \datasetname{} inspires others to consider speedrunning as an avenue for pushing model abilities such as strategic thinking and long horizon coherence.

\subsection*{AI use statement}

We used generative AI tools to design or provide feedback on research methodology or experiments, to implement methods, to assist with translation, to clean and reformat datasets, and to support qualitative and thematic data analysis.
We have not used generative AI tools to generate synthetic data sets, to help develop theoretical models or conceptual frameworks, to formulate mathematical claims, to provide critical ingredients for proving mathematical claims, to assist in the writing of proofs, or to propose or refine hypotheses.
We have reviewed all AI-assisted work manually and take responsibility for the final content of this work, including text, claims or artifacts produced with the aid of generative AI.

\subsection*{Ethics Statement}
\label{subsec:ethics}

\datasetname{} will be released for research and benchmarking purposes only and it will soley be limited to open-source games and licensed codes for public release to the community for reproducibility purpose. None of the copyrighted assets will be publicly released. This paper and associated assets conveys no right to access, reproduce, distribute, modify, or commercially exploit any third-party title or its assets beyond what applicable licenses, terms of service, and law already permit. Users are responsible for lawfully obtaining the titles they evaluate on and for securing any permissions their intended use requires.

\bibliography{iclr2027_conference}
\bibliographystyle{iclr2027_conference}

\appendix
\section{Appendix}

\subsection{Opensource Game Code Example}
Our anonymized code for the patched version of Tuxemon is publicly available at  \url{https://anonymous.4open.science/r/tuxemon-benchmark-anon-A2A4} under GPL-3.0 (code), CC-BY-SA (assets) following the original Tuxemon code and assets~\citep{tuxemon}.

\subsection{Per-Game Results}
\label{app:pergame}

\begin{figure}[H]
\begin{center}
\includegraphics[width=\linewidth]{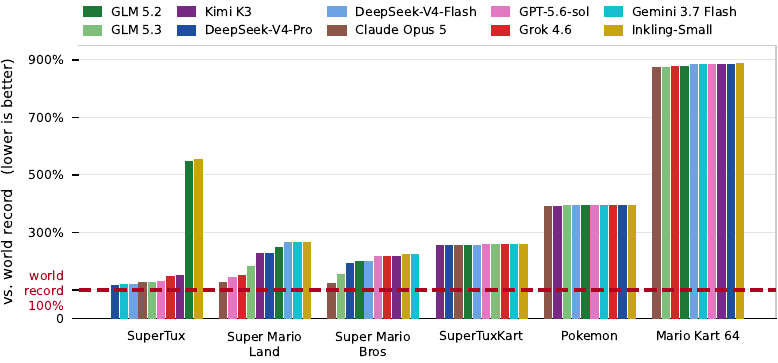}
\end{center}
\caption{Per-game results: best result per model and title, as a percentage of the human
world record (lower is better, and the dashed line is the record). Figure~\ref{fig:best-vs-record}
averages every title here. Each bar is the better of that model's \textsc{offline-scratch}
and \textsc{offline-seed} run. Bars from runs that never beat the seed they were handed report the
seed's score rather than the model's work. These are the Inkling-Small on SuperTux, the three
right-most models in Super Mario Land, and the right-most models in SuperTuxKart. GLM 5.2 never cleared
SuperTux from scratch, so its bar shows its seeded run. Mario Kart 64 has seeded runs only since no model cleared from scratch runs, and
every model drives without using shortcuts, while the record uses a shortcut that skips most of the track.
}
\label{fig:pergame}
\end{figure}

Figure~\ref{fig:pergame} shows the breakdown of results per game.

\begin{figure}[H]
\centering
\begin{subfigure}{\linewidth}
    \centering
    \includegraphics{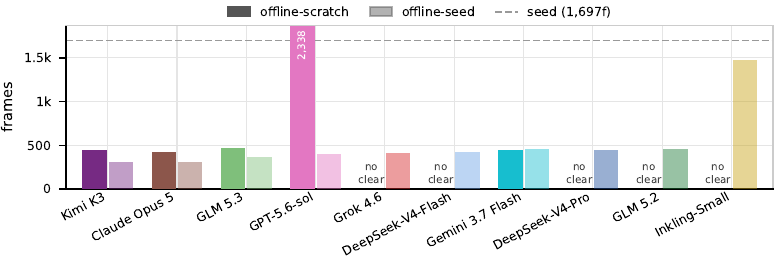}
    \caption{Game, browser, scorer and documentation all withheld.}
    \label{fig:astray-banned}
\end{subfigure}

\vspace{0.6em}
\begin{subfigure}{\linewidth}
    \centering
    \includegraphics{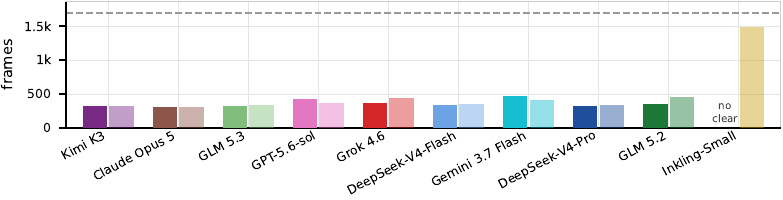}
    \caption{Scorer withheld, game and documentation accessible.}
    \label{fig:astray-accessible}
\end{subfigure}
\caption{Final results on Astray level 1 under \textsc{offline-scratch} (solid) and
\textsc{offline-seed} (lighter shade of the same colour), after 50 turns per run (lower is better,
and the dashed line is the 1{,}697-frame seed). Both runs withhold the in-turn scorer. In (a) the
agent also has no access to the game, a browser or the repository documentation. In (b) the game
and documentation are available, so agents can load the maze and simulate it locally. Models are in
the same order in both panels, ranked by (a), and each cell is a single run. GPT-5.6-sol's
\textsc{offline-scratch} run in (a) reaches 2{,}338 frames and is clipped.}
\label{fig:astray-runs}
\end{figure}

\paragraph{Astray} Figure~\ref{fig:astray-runs} shows the Astray results.

\paragraph{Civilization} Across all models given 100 turns with no in-trun evaluations enabeled, none of the models improve the number of frames to reach the goal.

\begin{figure}[H]
\begin{center}
\includegraphics{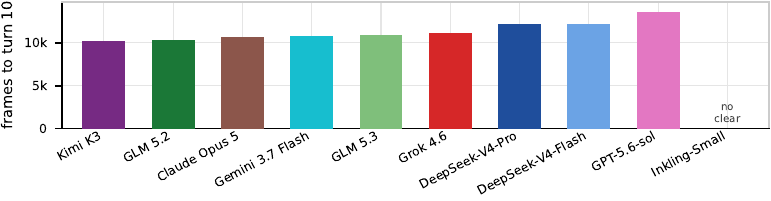}
\end{center}
\caption{Frames to reach turn 10 of Civilization  I, best verified run per model (lower is
better). Every run had the in-turn scorer withheld and used its full 100 turns. 
Inkling-Small never reached turn 10.}
\label{fig:civ}
\end{figure}

\end{document}

%% file: math_commands.tex
\usepackage{amsmath,amsfonts,bm}

\def\eqref#1{equation~\ref{#1}}

\def\1{\bm{1}}

\DeclareMathAlphabet{\mathsfit}{\encodingdefault}{\sfdefault}{m}{sl}
\SetMathAlphabet{\mathsfit}{bold}{\encodingdefault}{\sfdefault}{bx}{n}

%% file: iclr2027_conference.bbl
\begin{thebibliography}{69}
\providecommand{\natexlab}[1]{#1}
\providecommand{\url}[1]{\texttt{#1}}
\expandafter\ifx\csname urlstyle\endcsname\relax
  \providecommand{\doi}[1]{doi: #1}\else
  \providecommand{\doi}{doi: \begingroup \urlstyle{rm}\Url}\fi

\bibitem[{abney317}(2026)]{abney3172026mk64}
{abney317}.
\newblock {Mario Kart 64} -- all cups (skips) in 22:24.640.
\newblock Speedrun.com, 2026.
\newblock URL \url{https://www.speedrun.com/mk64/runs/y84r7gdz}.

\bibitem[Aloupis et~al.(2015)Aloupis, Demaine, Guo, and
  Viglietta]{aloupis2015nintendo}
Greg Aloupis, Erik~D. Demaine, Alan Guo, and Giovanni Viglietta.
\newblock Classic {Nintendo} games are (computationally) hard.
\newblock \emph{Theoretical Computer Science}, 586:\penalty0 135--160, 2015.
\newblock \doi{10.1016/j.tcs.2015.02.037}.

\bibitem[{Anthropic}(2026)]{anthropic2026opus5}
{Anthropic}.
\newblock System card: {Claude} {Opus} 5.
\newblock
  \url{https://www-cdn.anthropic.com/ceaf5c7ff2783855203fde8208ec311252dced5b/Claude\%20Opus\%205\%20System\%20Card.pdf},
  July 2026.
\newblock Accessed: 2026-08-24.

\bibitem[{ARC Prize Foundation}(2026)]{arcprize2026arcagi3}
{ARC Prize Foundation}.
\newblock {ARC-AGI-3}: A new challenge for frontier agentic intelligence.
\newblock Technical report, ARC Prize Foundation, 2026.
\newblock arXiv:2603.24621.

\bibitem[Badia et~al.(2020)Badia, Piot, Kapturowski, et~al.]{badia2020agent57}
Adri{\`a}~Puigdom{\`e}nech Badia, Bilal Piot, Steven Kapturowski, et~al.
\newblock Agent57: Outperforming the {Atari} human benchmark.
\newblock In \emph{Proceedings of the 37th International Conference on Machine
  Learning (ICML)}, 2020.

\bibitem[Bellemare et~al.(2013)Bellemare, Naddaf, Veness, and
  Bowling]{bellemare2013ale}
Marc~G. Bellemare, Yavar Naddaf, Joel Veness, and Michael Bowling.
\newblock The arcade learning environment: An evaluation platform for general
  agents.
\newblock \emph{Journal of Artificial Intelligence Research}, 47:\penalty0
  253--279, 2013.
\newblock \doi{10.1613/jair.3912}.

\bibitem[Berner et~al.(2019)Berner, Brockman, Chan, et~al.]{berner2019dota}
Christopher Berner, Greg Brockman, Brooke Chan, et~al.
\newblock Dota 2 with large scale deep reinforcement learning.
\newblock \emph{arXiv preprint arXiv:1912.06680}, 2019.

\bibitem[{Bisqwit}(2003)]{bisqwit2003smb}
{Bisqwit}.
\newblock {NES} {Super Mario Bros.} ``warps'' in 05:15.650.
\newblock TASVideos publication \#665, December 2003.
\newblock URL \url{https://tasvideos.org/665M}.
\newblock Tool-assisted speedrun; Famtasia 5.1; obsoleted.

\bibitem[Campbell et~al.(2002)Campbell, Hoane, and Hsu]{campbell2002deepblue}
Murray Campbell, A.~Joseph Hoane, and Feng-hsiung Hsu.
\newblock Deep blue.
\newblock \emph{Artificial Intelligence}, 134\penalty0 (1--2):\penalty0 57--83,
  2002.
\newblock \doi{10.1016/S0004-3702(01)00129-1}.

\bibitem[Colantonio(2026)]{polygon2026smbace}
Giovanni Colantonio.
\newblock Super mario bros. players uncover the biggest glitch in the game's
  40-year history.
\newblock \emph{Polygon}, April 2026.
\newblock URL
  \url{https://www.polygon.com/super-mario-bros-ace-trick-discovered/}.

\bibitem[Comanici et~al.(2025)Comanici, Bieber, Schaekermann, Pasupat,
  Sachdeva, Dhillon, Blistein, Ram, Zhang, Rosen, Marris, Petulla, Gaffney,
  Aharoni, Lintz, Pais, Jacobsson, Szpektor, Jiang, Haridasan, Omran, Saunshi,
  Bahri, Mishra, Chu, Boyd, Hekman, Parisi, Zhang, Kawintiranon, Bedrax-Weiss,
  Wang, Xu, Purkiss, Mendlovic, Deutel, Nguyen, Langley, Korn, Rossazza, Ramé,
  Waghmare, Miller, Byrd, Sheshan, Hadsell, Bhardwaj, Janus, Rissa, Horgan,
  Abdagic, Belenki, Allingham, Singh, Guidroz, Srinivasan, Schmit, Chiafullo,
  Elisseeff, Jha, Kolhar, Berrada, Ding, Si, Mallick, Och, Erell, Ni, Latkar,
  Yang, Sirkovic, Feng, Leland, Hornung, Wu, Blundell, Alvari, Huang, Yip,
  Deur, Liu, Surita, Duque, Damen, Jia, Guez, Mircea, Sinha, Magni, Stradomski,
  Marian, Galić, Chen, Husain, Singhal, Grewe, Aubet, Song, Blanco, Rechis,
  Ho, Munoz, Zheng, Hamrick, Mather, Taitelbaum, Rutherford, Lei, Chen, Shukla,
  Moreira, Doi, Isik, Shabat, Rogozińska, Kolipaka, Chang, Vušak,
  Venkatachary, Noghabi, Bharti, Jun, Zaks, Green, Challagundla, Wong,
  Mohammad, Hirsch, Cheng, Naim, Proleev, Vincent, Singh, Krikun, Krishnan,
  Ghahramani, Atias, Aggarwal, Kirov, Vytiniotis, Koh, Chronopoulou, Dogra,
  Ion, Tyen, Lee, Weissenberger, Strohman, Balakrishna, Rae, Velic,
  de~Liedekerke, Elyada, Yuan, Liu, Shani, Kishchenko, Alessio, Li, Song, Kwei,
  Jankowski, Pappu, Namiki, Ma, Tripuraneni, Cherry, Ikonomidis, Ling, Ji,
  Westberg, Wright, Yu, Parkinson, Ramaswamy, Connor, Yeganeh, Grover,
  Kenwright, Litchev, Apps, Tomala, Halim, Castro-Ros, Li, Boral, Sho, Yarom,
  Malmi, Klinghoffer, Lin, Ansell, S, Zhao, Zuo, Santoro, Cheng, Demmessie,
  Liu, Brichtova, Culp, Braun, Graur, Ng, Mehta, Phillips, Sundberg, Godbole,
  Liu, Katariya, Rim, Seyedhosseini, Ammirati, Valfridsson, Malihi, Knight,
  Toor, Lampe, Ittycheriah, Chiang, Yeung, Fréchette, Rao, Wang, Srivastava,
  Zhang, Rhodes, Brand, Weesner, Figotin, Gimeno, Fellinger, Marcenac, Leal,
  Marcus, Cotruta, Cabrera, Luo, Garrette, Axelrod, Baltateanu, Barker, Chen,
  Toma, Ingram, Riesa, Kulkarni, Zhang, Liu, Wang, Polacek, Wu, Hui, Reyes, Su,
  Barnes, Malhi, Siddiqui, Feng, Damaschin, Pighin, Steiner, Yang, Boppana,
  Ivanov, Kandoor, Shah, Mujika, Huang, Choquette-Choo, Patel, Yu, Creswell,
  Jerry, Liu, Barros, Razeghi, Roy, Culliton, Xiong, Pan, Strohmann, Powell,
  Seal, DeCarlo, Shyam, Katircioglu, Wang, Hardin, Odisho, Broder, Chang, Nair,
  Shtefan, O'Brien, Agarwal, Potluri, Goyal, Jhindal, Thakur, Stuken, Lyon,
  Toutanova, Feng, Wu, Horn, Wang, Cullum, Taubman, Shrivastava, Shi,
  Tomlinson, Patel, Tu, Oflazer, Pongetti, Yang, Taïga, Perot, Pierse, Han,
  Drori, Iturrate, Chakrabarti, Yeung, Dopson, ting Chen, Kulshreshtha, Guo,
  Pham, Schuster, Chen, Polozov, Xing, Zhou, Kacham, Kukliansky, Miech,
  Yaroshenko, Chi, Douglas, Fei, Blondel, Myla, Madmoni, Wu, Keysers, Kjems,
  Albuquerque, Yu, D'sa, Plantan, Ionescu, Elias, Gupta, Vuyyuru, Alcober,
  Zhou, Ji, Hartmann, Puttagunta, Song, Amid, Stefanoiu, Lee, Pucciarelli,
  Wang, Raul, Petrov, Tian, Anklin, Nti, Gomes, Schumacher, Vesom,
  Panagopoulos, Bousmalis, Andor, Jacob, Zhang, Rosgen, Kecman, Tung, Belias,
  Goodman, Covington, Wieder, Saxena, Davoodi, Huang, Maddineni, Roulet,
  Campbell-Ajala, Sessa, Xintian, Wu, Lai, Collins, Haig, Sakenas, Xu,
  Giustina, Shafey, Charoenpanit, Garg, Ainslie, Severson, Arenas, Pathak,
  Rajayogam, Feng, Bakker, Li, Wichers, Rogers, Geng, Li, Jagerman, Jia,
  Olmert, Sharon, Mauger, Mariserla, Ma, Mohabey, Kim, Andreev, Pollom, Love,
  Jain, Agrawal, Schroecker, Fortin, Warmuth, Liu, Leach, Blok, Girirajan,
  Aharoni, Uria, Sozanschi, Goldberg, Ionita, Ribeiro, Zlocha, Birodkar,
  Lachgar, Yuan, Choudhury, Ginsberg, Zheng, Dibb, Graves, Lokhande, Rasskin,
  Muraru, Quick, Tata, Sermanet, Chawla, Karo, Wang, Zhang, Keller, Dragan, Su,
  Chou, Liu, Tao, Prabhakara, Wilson, Liu, Wang, Evans, Du, Castaño, Prasad,
  Mahdy, Gerlach, Reid, Kahn, Zait, Pillai, Ulrich, Wang, Wassenberg, Farkash,
  Yalasangi, Wang, Bauza, Bucher, Liu, Yan, Leung, Sindhwani, Barnes, Singh,
  Jurin, Chang, Bhumihar, Eiger, Citovsky, Withbroe, Li, Xue, Santo, Stoyanov,
  Raimond, Zheng, Gao, Listík, Kwasiborski, Saputro, Ozturel, Mallya,
  Majmundar, West, Caron, Wei, Castrejon, Vikram, Ramachandran, Dhawan, Park,
  Smoot, van~den Driessche, Blau, Malik, Liang, Hirsch, dos Santos, Weinstein,
  van~den Oord, Lall, FitzGerald, Jiang, Yang, Webster, Elqursh, Pope, Rotival,
  Raposo, Zhu, Dean, Alabed, Tran, Gupta, Gleicher, Austin, Rosseel, Umekar,
  Das, Sun, Chen, Misiunas, Zhou, Di, Loo, Newlan, Li, Ramasesh, Xu, Chen,
  Gandhe, Soricut, Gupta, Hu, El-Sayed, Garcia, Brusilovsky, Chen, Bolt, Huang,
  Gurney, Zhang, Pritzel, Wilkiewicz, Seybold, Shamanna, Fischer, Dean, Gill,
  Mcilroy, Bhowmick, Selier, Yang, Cheng, Magay, Tan, Varma, Walder, Kocisky,
  Nakashima, Natsev, Kwong, Gog, Zhang, Dieleman, Jimma, Ryabtsev, Brahma,
  Steiner, Du, Žužul, Žanić, Raghavachari, Gierke, Zheng, Petrova, Dauphin,
  Liu, Kessler, Hand, Duvarney, Kim, Lee, Hussenot, Hui, Smith, Jain, Xia,
  Tomar, Amiri, Phan, Fuchs, Weyand, Tomasev, Cordell, Liu, Mallinson, Joshi,
  Crawford, Suggala, Chien, Fernando, Sanchez-Vargas, Williams, Crone, Luo,
  Karpov, Shan, Thurk, Strudel, Voigtlaender, Patil, Dozat, Khodaei, Singla,
  Ambroszczyk, Wu, Chang, Roark, Hegde, Ding, Filos, Wu, Pinto, Liu, Khanna,
  Pandey, Mcloughlin, Li, Haves, Zhou, Buchatskaya, Leal, de~Boursac, Akazawa,
  Anderson, Chen, Somandepalli, Liang, Goenka, Winkler, Grushetsky, Ding,
  Smith, Ye, Pont-Tuset, Li, Li, Golany, Wegner, Jiang, Barak, Shangguan,
  Vértes, Wong, Bornschein, Tudor, Bevilacqua, Schaul, Rawat, Zhao, Axiotis,
  Meng, McLean, Lai, Beattie, Kushman, Liu, Kutzman, Lang, Ye, Netrapalli,
  Mishra, Khan, Goel, Willoughby, Tian, Zhuang, Chen, Tsai, Kementsietsidis,
  Khare, Keeling, Xu, Waters, Altché, Popat, Mittal, Saxton, Badawy, Mathieu,
  Zheng, Zhou, Ranka, Shin, Duan, Salimans, Mihailescu, Shaham, Chang, Assael,
  Dikkala, Izzard, Cohen-Addad, Graves, Feinberg, Chung, Strouse, Karmon,
  Sharifzadeh, Ashwood, Pham, Blanton, Vasiloff, Barber, Geller, Zhou, Zubach,
  Huang, Zhang, Gupta, Young, Proskurnia, Votel, Gabeur, Barcik, Tripathi, Yu,
  Yan, Changpinyo, Pavetić, Coyle, Fujii, Mendez, Zhou, Rajamani, Hechtman,
  Cao, Juan, Tan, Dalibard, Du, Clay, Yao, Jia, Vijaykumar, Zhou, Bai, Hung,
  Pecht, Todorov, Khadke, Gupta, Lahoti, Autef, Duddu, Lee-Thorp, Bykovsky,
  Misiunas, Flennerhag, Thangaraj, McGiffin, Nado, Kunesch, Noever, Hertz,
  Liang, Stone, Palmer, Daruki, Pramanik, Põder, Kyker, Khan, Sluzhaev,
  Ritter, Ruderman, Zhou, Nagpal, Vodrahalli, Necula, Barham, Pavlick,
  Hartford, Shafran, Zhao, Mikuła, Eccles, Shimokawa, Garg, Vilnis, Chen,
  Shumailov, Lee, Abdelhamed, Xie, Cohen, Hlavnova, Malkin, Sitawarin, Lottes,
  Coquinot, Yu, Kumar, Zhang, Mahendru, Ahmed, Martens, Chen, Boag, Peng,
  Devin, Klimovskiy, Phuong, Vainstein, Xie, Ramabhadran, Howard, Yu, Goswami,
  Cui, Shleifer, Pinto, Yeh, Yang, Javanmardi, Ethier, Lee, Orbay, Kotecha,
  Bromberg, Shaw, Thornton, Rosenthal, Gu, Thomas, Gemp, Ayyar, Ushio, Selvan,
  Wee, Liu, Majzoubi, Yu, Abernethy, Liechty, Pan, Nguyen, Qiong, Hu, Perrin,
  Arora, Pitler, Wang, Shivakumar, Prost, Limonchik, Wang, Gao, Cour, Buch,
  Gui, Ivanova, Neubeck, Chan, Kim, Chen, Goyal, Chung, Liu, Su, Petrushkina,
  Shen, Joulin, Xu, Lin, Kulizhskaya, Chelba, Vasudevan, Collins, Bashlovkina,
  Lu, Fritz, Park, Zhou, Su, Tanburn, Sushkov, Rasquinha, Li, Prendki, Li, LV,
  Sharma, Fitoussi, Huang, Dai, Dao, Burrows, Prior, Qin, Pundak, Sjoesund,
  Khurshudov, Zhu, Webson, Kemp, Tan, Agrawal, Sargsyan, Cheng, Stephan,
  Kwiatkowski, Reid, Byravan, Michaely, Heess, Zhou, Goenka, Carpenter,
  Levskaya, Wang, Roberts, Leblond, Chikkerur, Ginzburg, Chang, Riachi,
  Chuqiao, Xu, Borsos, Pliskin, Pawar, Lustman, Kirkwood, Anand, Chaudhary,
  Kalb, Milan, Augenstein, Goldie, Prince, Raman, Sun, Xia, Cohen, Huo, Camp,
  Ellis, Zilka, Torres, Patel, Arora, Chan, Adler, Ayoub, Liang, Jamil, Jiang,
  Baumgartner, Sun, Karov, Akulov, Zheng, Cai, Fantacci, Rubin, Acha, Wang,
  D'Souza, Sathyanarayana, Dai, Rowe, Simanovsky, Goldman, Kuang, Pan,
  Rosenberg, Rojas-Esponda, Dutta, Zeng, Jurenka, Farquhar, Bansal, Iqbal,
  Roelofs, Joung, Beak, Ryu, Poplin, Wu, Alayrac, Buthpitiya, Ronneberger,
  Habtegebriel, Li, Cavallaro, Wei, Bensky, Denk, Ganapathy, Stanway, Joshi,
  Bertolini, Lo, Ma, Charles, Sampemane, Sahni, Chen, Askham, Gaddy, Young,
  Tan, Eyal, Bražinskas, Zhong, Wu, Epstein, Bailey, Hard, Lee, Goldshtein,
  Ruiz, Badawi, Lochbrunner, Kearns, Brown, Pardo, Weber, Yang, Jiang, Akin,
  Fu, Wainwright, Zou, Gaba, Manzagol, Kan, Song, Zainullina, Lin, Ko,
  Deshmukh, Jindal, Svensson, Tyam, Zhao, Kaeser-Chen, Baird, Moradi, Hall,
  Guo, Tsang, Liang, Pereira, Ganesh, Korotkov, Adamek, Thiagarajan, Tran,
  Chen, Tar, Jain, Dasgupta, Bilal, Reitter, Zhao, Vezzani, Gehman, Mehta,
  Beltrone, Dotiwalla, Guadarrama, Abbas, Karp, Georgiev, Ferng, Brockschmidt,
  Peng, Hirnschall, Verma, Bi, Xiao, Dabush, Xu, Wallis, Parker, Wang, Xu,
  Safarli, Tewari, Zhang, Kim, Gesmundo, Thomas, Levi, Chowdhury, Rao, Garst,
  Conway-Rahman, Ran, McKinney, Xiao, Yu, Agrawal, Stjerngren, Ionescu, Chen,
  Sharma, Chiu, Liu, Franko, Sanford, Cai, Michel, Ganapathy, Labanowski,
  Garrett, Vargas, Sun, Gale, Buschmann, Desjardins, Ghelani, Jain, Verma,
  Asawaroengchai, Eisenschlos, Harlalka, Kazawa, Metzler, Howland, Jian, Ades,
  Shah, Gangwani, Lee, Ring, Hernandez, Reich, Sinha, Sathe, Kovac, Gill,
  Kannan, D'olimpio, Sevenich, Whang, Kim, Sim, Chen, Zhang, Lall, Matias, Jia,
  Friesen, Nasso, Thapliyal, Perozzi, Yu, Shekhawat, Huda, Grabowski, Wang,
  Sreevatsa, Dib, Hassen, Schuh, Milutinovic, Welty, Quinn, Shah, Wang,
  Barth-Maron, Frye, Axelsson, Zhu, Ma, Giannoumis, Sedghi, Ye, Luan, Aydin,
  Chandra, Sampathkumar, Huang, Lavrenko, Eleryan, Hong, Hansen, Carthy,
  Samanta, Ćevid, Wang, Li, Voznesensky, Hoffman, Terzis, Sehwag, Fidel, He,
  Cai, He, Feng, Nikoltchev, Phatale, Chase, Lawton, Zhang, Ouyang, Tragut,
  Manshadi, Narayanan, Shen, Gao, Bolukbasi, Roy, Li, Golovin, Panait, Qin,
  Han, Anthony, Kudugunta, Patraucean, Ray, Chen, Yang, Bhatia, Talluri,
  Morris, Ražnatović, Brownfield, An, Peng, Kane, Zheng, Duduta, Kessinger,
  Noraky, Liu, Rong, Veličković, Rush, Goldin, Wei, Garlapati, Pantofaru,
  Kwon, Ni, Noland, Trapani, Beaufays, Roy, Chow, Turker, Cideron, Mei, Clark,
  Dou, Bošnjak, Leith, Du, Yazdanbakhsh, Nasr, Kwak, Sheth, Kaskasoli, Anand,
  Lakshminarayanan, Jerome, Bieber, Chu, Senges, Shen, Sridhar, Ndebele,
  Beyret, Mohamed, Chen, Freitag, Guo, Liu, Roit, Chen, Yan, Stone, Co-Reyes,
  Cole, Scellato, Azizi, Hashemi, Jin, Iyer, Valentine, György, Ahuja, Diaz,
  Lee, Clement, Kong, Garmon, Watts, Bhatia, Gupta, Miecnikowski, Vallet, Taly,
  Loper, Joshi, Atwood, Chick, Collier, Iliopoulos, Trostle, Gunel,
  Leal-Cavazos, Hrafnkelsson, Guzman, Ju, Forbes, Emond, Chauhan, Caine, Xiao,
  Zeng, Moufarek, Murphy, Meng, Gupta, Riedel, Das, Lawal, Narayan, Sosea,
  Swirhun, Friso, Neyshabur, Lu, Girgin, Wunder, Yvinec, Pyne, Carbune,
  Rijhwani, Guo, Doshi, Briukhov, Bain, Hitron, Wang, Gupta, Chen, Du, Zhang,
  Shah, Akula, Dylla, Kachra, Kuo, Zou, Wang, Xu, Zhu, Snyder, Menon, Firat,
  Mordatch, Yuan, Ponomareva, Blevins, Moore, Wang, Chen, Scholz, Dwornik, Lin,
  Li, Antognini, I, Song, Miller, Kalra, Raveret, Akerlund, Wu, Nystrom,
  Godbole, Liu, DeBalsi, Zhao, Liu, Caciularu, Lax, Khandelwal, Langston,
  Bailey, Lattanzi, Wang, Kovelamudi, Mondal, Guruganesh, Hua, Roval,
  Wesołowski, Ingale, Halcrow, Sohn, Angermueller, Raad, Stickgold, Lu, Kosik,
  Xie, Lillicrap, Huang, Zhang, Paulus, Farabet, Wertheim, Wang, Joshi, ling
  Ko, Wu, Agrawal, Lin, Sheng, Sung, Breland-King, Butterfield, Gawde, Singh,
  Zhang, Apte, Shetty, Hutter, Li, Salesky, Lebron, Kanerva, Paganini, Nguyen,
  Vallu, Peter, Velury, Kao, Hoover, Bortsova, Bishop, Jakobovits, Agostini,
  Agarwal, Liu, Kwong, Tavakkol, Bica, Greve, GP, Marcus, Hou, Duerig,
  Moroshko, Lacey, Davis, Amelot, Wang, Kim, Strinopoulos, Wan, Lan, Krishnan,
  Tang, Humphreys, Bai, Shtacher, Machado, Pang, Burke, Liu, Aravamudhan, Song,
  Hirst, Singh, Jou, Bai, Piccinno, Fu, Alazard, Meiri, Winter, Chen, Zhang,
  Heitkaemper, Lambert, Lee, Frömmgen, Rogulenko, Nair, Niemczyk, Bulyenov,
  Xu, Shemtov, Zadimoghaddam, Toropov, Wirth, Dai, Gollapudi, Zheng, Kurakin,
  Lee, Bullard, Serrano, Balazevic, Li, Schalkwyk, Murphy, Zhang, Sequeira,
  Datta, Agrawal, Sutton, Attaluri, Chiang, Farhan, Thornton, Lin, Choma,
  Nguyen, Dasgupta, Robinson, Comşa, Riley, Pillai, Mustafa, Golan, Zandieh,
  Lespiau, Porter, Ross, Rajayogam, Agarwal, Venugopalan, Shahriari, Yan, Xu,
  Tobin, Dubov, Shi, Recasens, Kovsharov, Borgeaud, Dery, Vasanth, Gribovskaya,
  Qiu, Mahdieh, Skut, Nielsen, Zheng, Yu, Bostock, Gupta, Archer, Rawles,
  Davies, Svyatkovskiy, Tsai, Halpern, Reisswig, Wydrowski, Chang, Puigcerver,
  Taege, Li, Schnider, Li, Dena, Xu, Telang, Shi, Zen, Kastner, Ko,
  Subramaniam, Kumar, Blois, Dai, Wieting, Lu, Zeldes, Xie, Hauth, Ţifrea, Li,
  El-Husseini, Abolafia, Zhou, Ding, Ghalebikesabi, Guía, Maksai, Ágoston
  Weisz, Arik, Sukhanov, Świetlik, Jia, Yu, Wang, Brand, Bloxwich, Kirmani,
  Chen, Go, Sprechmann, Kannen, Carin, Sandhu, Edkins, Nooteboom, Gupta,
  Maggiore, Azizi, Pritch, Yin, Gupta, Tarlow, Smith, Ivanov, Babaeizadeh,
  Goel, Kambala, Chu, Kastelic, Liu, Soltau, Stone, Agrawal, Kim, Soparkar,
  Tadepalli, Bunyan, Soh, Kannan, Kim, Chen, Halumi, Roy, Wang, Sercinoglu,
  Gibson, Bhatnagar, Sano, von Dincklage, Ren, Mitrevski, Olšák, She,
  Doersch, Jilei, Wang, Liu, Tan, Yakar, Warkentin, Ramirez, Lebsack, Dillon,
  Mathews, Cobley, Wu, Chen, Simon, Nath, Sainath, Bendebury, Julian,
  Mankalale, Ćurko, Zacchello, Brown, Sodhia, Howard, Caelles, Gupta, Evans,
  Bulanova, Katzen, Goldenberg, Tsitsulin, Stanton, Schillings, Kovalev, Fry,
  Shah, Lin, Upadhyay, Li, Radpour, Maggioni, Xiong, Haas, Brennan, Kamath,
  Savinov, Nagrani, Yacovone, Kappedal, Andriopoulos, Lao, Li, Rozhdestvenskiy,
  Hashimoto, Audibert, Austin, Rodriguez, Ruoss, Honke, Karkhanis, Xiong, Wei,
  Huang, Leng, Premachandran, Bileschi, Evangelopoulos, Mensink, Pavagadhi,
  Teplyashin, Chang, Xue, Tanzer, Goldman, Patel, Li, Wiesner, Zheng,
  Stewart-Binks, Han, Li, Luo, Lenc, Lučić, Xue, Mullins, Guseynov, Chang,
  Galatzer-Levy, Zhang, Bingham, Hu, Hartman, Ma, Griffith, Irpan, Radebaugh,
  Yue, Fan, Ungureanu, Sorokin, Teufel, Li, Anil, Paparas, Wang, Lin, Peng,
  Shum, Petrovic, Brady, Nguyen, Macherey, Li, Singh, Yenugula, Iinuma, Chen,
  Kopparapu, Stern, Dave, Thekkath, Perot, Kumar, Li, Xiao, Bilotti, Bateni,
  Noble, Lee, Vázquez-Reina, Salazar, Yang, Wang, Gruzewska, Rao, Raghuram,
  Xu, Ben-David, Mei, Dalmia, Zhang, Liu, Bansal, Pankov, Schwarcz, Burns,
  Chan, Sanghai, Liang, Liang, He, Stuart, Narayanan, Zhu, Frank, Fatemi,
  Sabne, Lang, Bhattacharya, Settle, Wang, McMahan, Tacchetti, Soares, Hadian,
  Cabi, Chung, Putikhin, Li, Chen, Tarango, Michalewski, Kazemi, Masoom,
  Sheftel, Shivanna, Vadali, Comanescu, Reid, Moore, Neelakantan, Sander,
  Herzig, Rosenberg, Dehghani, Choi, Fink, Hayes, Ge, Weng, Ho, Karro, Krishna,
  Thiet, Skerry-Ryan, Eppens, Andreetto, Sarma, Bonacina, Ayan, Nawhal, Shan,
  Dusenberry, Thakoor, Gubbi, Nguyen, Tsarfaty, Albanie, Mitrović, Gandhi,
  Chen, Epasto, Stephanov, Jin, Gehman, Amini, Weber, Behbahani, Xu, Allamanis,
  Chen, Ott, Sha, Jastrzebski, Qi, Greene, Wu, Toki, Vlasic, Shapiro,
  Kotikalapudi, Shen, Saeki, Xie, Cassirer, Bharadwaj, Kiyono, Bhojanapalli,
  Rosenfeld, Ritter, Mao, Oliveira, Egyed, Bandemer, Parisotto, Kinoshita,
  Pluto, Maniatis, Li, Guo, Ghiasi, Tarbouriech, Chatterjee, Jin, Katrina, Xu,
  Palomaki, Arnold, Sewak, Piccinini, Sharma, Albrecht, Purser-haskell,
  Vaswani, Chen, Wisniewski, Cao, Aslanides, Phu, Sieb, Agubuzu, Zheng, Sohn,
  Selvi, Andreassen, Subudhi, Eruvbetine, Woodman, Mery, Krause, Ren, Ma, Luo,
  Chen, Fan, Griffiths, Schuler, Li, Zhang, Sarr, Luo, Patana, Watson,
  Naboulsi, Collins, Sidhwani, Hoogeboom, Silver, Caveness, Zhao, Rodriguez,
  Deines, Bai, Griffin, Tagliasacchi, Xue, Babbula, Pang, Ding, Shen, Peake,
  Crocker, Raghvendra, Swisher, Han, Singh, Wu, Pchelin, Munkhdalai, Alon,
  Bacon, Robles, Bulian, Johnson, Powell, Ferreira, Li, Benzing, Velimirović,
  Soyer, Kong, Tony, Nguyên, Yang, Liu, van Amersfoort, Gillick, Sun,
  Rauschmayr, Zhang, Zhan, Zhou, Frolov, Yang, Vnukov, Rouillard, Li, Mandhane,
  Fallen, Venkataraman, Hu, Brennan, Lee, Chang, Sundermeyer, Pan, Ke, Tong,
  Fabrikant, Bono, Gu, Foley, Mao, Delakis, Bhaswar, Frostig, Li, Zipori, Hope,
  Kozlova, Mishra, Djolonga, Schiff, Merey, Briakou, Morgan, Wan, Hassidim,
  Skerry-Ryan, Sengupta, Jasarevic, Kallakuri, Kunkle, Brennan, Lieber,
  Mansoor, Walker, Zhang, Xie, Žužić, Chukwuka, Druinsky, Cho, Yao, Naeem,
  Butt, Kim, Jia, Jordan, Lelkes, Kurzeja, Wang, Zhao, Over, Chakladar,
  Prasetya, Jha, Ganapathy, Cong, Shroff, Saroufim, Miryoosefi, Hammad, Nasir,
  Xi, Gao, Maeng, Hora, Cheng, Haghani, Lewenberg, Lu, Matysiak, Raisinghani,
  Wang, Baugher, Sukthankar, Giang, Schultz, Fiedel, Chen, Lee, Dey, Zheng,
  Paul, Smith, Ly, Wang, Bansal, Perz, Ricco, Blank, Keshava, Sharma, Chow,
  Lad, Jalan, Osindero, Swanson, Scott, Ilić, Li, Jonnalagadda, Soudagar,
  Xiong, Batsaikhan, Jarrett, Kumar, Shah, Lawlor, Waters, Graham, May, Ramos,
  Lefdal, Cankara, Cano, O'Donoghue, Borovik, Liu, Grimstad, Alnahlawi,
  Tsihlas, Hudson, Grigorev, Jia, Huang, Igwe, Lebedev, Tang, Krivokon, Garcia,
  Tan, Jia, Stys, Vashishth, Liang, Venkatraman, Gu, Kementsietsidis, Zhu,
  Jung, Bai, Hosseini, Ahmed, Gupta, Yuan, Ashraf, Nigam, Vasudevan, Awasthi,
  Gilady, Mariet, Eskander, Li, Hu, Garrido, Schlattner, Zhang, Saxena, Dević,
  Muralidharan, Murthy, Zhou, Choi, Wongpanich, Wang, Shah, Xu, Huang, Spencer,
  Chen, Cohan, Wang, Tompson, Wu, Haroun, Li, Huergo, Yang, Yin, Wendt,
  Bendersky, Chaabouni, Snaider, Ferret, Jindal, Thompson, Xue, Bishop, Phal,
  Sharma, Sung, Radhakrishnan, Shomrat, Ingle, Vij, Gilmer, Istin, Sobell, Lu,
  Nottage, Sadigh, Willcock, Zhang, Xu, Brown, Lee, Wang, Zhu, Tay, Kim,
  Gutierrez, Sharma, Xian, Seo, Cui, Pochernina, Baetu, Jastrzębski, Ly,
  Elhawaty, Suh, Sezener, Wang, Yuen, Tucker, Cai, Yang, Wang, Muzio, Qian,
  Yoo, Lockhart, McKee, Guo, Mehrotra, Mendonça, Mehta, Ben, Tekur, Mu, Zhu,
  Krakovna, Lee, Maschinot, Cevey, Choe, Bai, Srinivasan, Gasaway, Young,
  Siegler, Holtmann-Rice, Piratla, Baumli, Yogev, Hofer, van Hasselt, Grant,
  Chervonyi, Silver, Hogue, Agarwal, Wang, Singh, Flynn, Lipschultz, David,
  Bellot, Yang, Le, Graziano, Olszewska, Hui, Maurya, Parotsidis, Chen,
  Oguntebi, Kelley, Baddepudi, Mauerer, Shaw, Siegman, Yang, Shetty, Roy, Song,
  Stokowiec, Burnell, Savant, Busa-Fekete, Miao, Ghosh, MacDermed, Lippe,
  Dektiarev, Behrman, Mentzer, Nguyen, Wei, Verma, Knutsen, Dasari, Yan,
  Mitrichev, Wang, Shejwalkar, Austin, Sunkara, Potti, Virin, Wright, Liu,
  Riva, Pot, Kochanski, Le, Balasubramaniam, Dhar, Liao, Bloniarz, Shukla,
  Cole, Lee, Zhang, Kafle, Vashishtha, Mahmoudieh, Chen, Hoffmann, Srinivasan,
  Lago, Shalom, Wang, Elabd, Sharma, Oh, Kothawade, Le, Monteiro, Yang,
  Alarakyia, Geirhos, Mincu, Garnes, Kobayashi, Mariooryad, Krasowiak, Zhixin,
  Lai, Mourad, Wang, Bu, Aharoni, Chen, Goyal, Zubov, Bapna, Dabir, Kothari,
  Lamerigts, Cao, Shar, Yew, Kulkarni, Mahaarachchi, Joshi, Zhu, Lichtarge,
  Zhou, Muckenhirn, Selo, Vinyals, Chen, Brohan, Mehta, Cogan, Wang, Geri, Ko,
  Chen, Viola, Shivam, Wang, Elish, Popa, Pereira, Liu, Koster, Kim, Zhang,
  Ebrahimi, Talukdar, Zheng, Poklukar, Mikhalap, Johnson, Vijayakumar,
  Omernick, Dibb, Dubey, Hu, Suman, Aggarwal, Kornakov, Xia, Lowe, Kolganov,
  Xiao, Nikolaev, Hemingray, Li, Iljazi, Rybiński, Sandhu, Lu, Luong,
  Jenatton, Govindaraj, Hui, Li, Dulac-Arnold, Park, Wang, Modi, Pouget-Abadie,
  Greller, Gupta, Berry, Ramachandran, Xie, McCafferty, Wang, Gupta, Lim,
  Bratanič, Brock, Akolzin, Sproch, Karliner, Kim, Goedeckemeyer, Shazeer,
  Schmid, Calandriello, Bhatia, Choromanski, Montgomery, Dua, Ramalho, King,
  Gao, Nguyen, Lindner, Pitta, Johnson, Salama, Ardila, Han, Farnese, Odoom,
  Wang, Ding, Rink, Smith, Lehri, Cohen, Vats, He, Gopavarapu, Paszke, Patel,
  Gansbeke, Loher, Castro, Voitovich, von Glehn, George, Niklaus, Eaton-Rosen,
  Rakićević, Jue, Perel, Zhang, Bahat, Pouget, Xing, Huot, Shenoy, Bos,
  Coriou, Richter, Noy, Wang, Ontanon, Qin, Makarchuk, Hassabis, Li, Sharma,
  Venkatesan, Kemaev, Daniel, Huang, Shah, Ponce, Warren, Chen, Faruqui, Wu,
  Andačić, Payrits, McDuff, Hume, Cao, Tessler, Wang, Wang, Rendulic,
  Agustsson, Johnson, Lando, Howard, Padmanabhan, Daswani, Banino, Kilgore,
  Heek, Ji, Caceres, Li, Kassner, Vlaskin, Liu, Grills, Hou, Sukkerd, Cheon,
  Shetty, Markeeva, Stanczyk, Iyer, Gong, Gao, Gopalakrishnan, Blyth, Reynolds,
  Bhoopchand, Bilenko, Gharibian, Zayats, Faust, Singh, Ma, Jiao,
  Vijayanarasimhan, Aroyo, Yadav, Chakera, Kakarla, Meshram, Gregor, Botea,
  Senter, Jia, Kovacs, Sharma, Baur, Kang, He, Zhuo, Kostelac, Laish, Peng,
  O'Bryan, Kasenberg, Rao, Leurent, Zhang, Stevens, Salazar, Zhang, Lobov,
  Walker, Porter, Redshaw, Ke, Rao, Lee, Lam, Moffitt, Kim, Qiao, Koo, Dadashi,
  Song, Sundararajan, Xu, Kawamoto, Zhong, Barbu, Reddy, Verzetti, Li,
  Papamakarios, Klimczak-Plucińska, Cassin, Kavukcuoglu, Swavely, Vaucher,
  Zhao, Hemsley, Tschannen, Ge, Menghani, Yu, Ha, He, Wu, Song, Sterneck,
  Zinke, Calian, Marsden, Ruiz, Hessel, Gueta, Lee, Farris, Gupta, Li, Saleh,
  Misra, Xiao, Mendolicchio, Buttimore, Krayvanova, Nayakanti, Wiethoff, Pande,
  Mirhoseini, Lao, Liu, Hua, Chen, Malkov, Kalashnikov, Gupta, Audhkhasi, Zhai,
  Kopalle, Jain, Ofek, Meyer, Baatarsukh, Strejček, Qian, Freedman, Figueira,
  Sokolik, Bachem, Lin, Kharrat, Hidey, Xu, Duan, Li, Ersoy, Everett, Cen,
  Santamaria-Fernandez, Taubenfeld, Mackinnon, Deng, Zablotskaia, Viswanadha,
  Goel, Yates, Deng, Choy, Chen, Sinha, Mossin, Wang, Szlam, Hao, Rubenstein,
  Toksoz-Exley, Aperghis, Zhong, Ahn, Isard, Lacombe, Luisier, Anastasiou,
  Kalley, Prabhu, Dunleavy, Bijwadia, Mao-Jones, Chen, Pasumarthi, Wood,
  Dostmohamed, Hurley, Simsa, Parrish, Pajarskas, Harvey, Skopek, Kochinski,
  Rey, Rieser, Zhou, Lee, Acharya, Li, Jiang, Zhang, Gipson, Mahintorabi,
  Gelmi, Khajehnouri, Yeh, Lee, Matthey, Baker, Pham, Fu, Pak, Gupta,
  Vasconcelos, Sadovsky, Walker, Hsiao, Zochbauer, Marzoca, Velan, Zeng,
  Baechler, Driess, Jain, Huang, Tao, Maggs, Levine, Schneider, Gemzer, Petit,
  Han, Fisher, Zelle, Biles, Ie, Fadeeva, Liu, Franco, Collister, Zhang, Wang,
  Zhao, Kieliger, Shuster, Zhu, Gong, Chan, Sun, Basu, Zimmermann, Hayes,
  Bapna, Snoek, Yang, Datta, Abdallah, Kilgour, Li, Mah, Jun, Rivière,
  Karmarkar, Spalink, Huang, Gonzalez, Tran, Nowak, Palowitch, Chadwick,
  Talius, Mehta, Sellam, Fränken, Nicosia, He, Kini, Amos, Basu, Jobe, Shaw,
  Xu, Evans, Ikeda, Yan, Jin, Wang, Yadav, Labzovsky, Sampath, Ma, Schumann,
  Siddhant, Shah, Youssef, Agarwal, Dabney, Tonioni, Ambar, Li, Guyon, Li,
  Soergel, Fang, Karadzhov, Udrescu, Trinh, Raunak, Noury, Guo, Gupta,
  Finkelstein, Petek, Liang, Billock, Sun, Wood, Song, Yu, Matejovicova, Cohen,
  Andra, D'Ambrosio, Deng, Nallatamby, Songhori, Dangovski, Lampinen, Botadra,
  Hillier, Cao, Baddi, Kuncoro, Yoshino, Bhagatwala, Ranzato, Schaeffer, Liu,
  Ye, Sarvana, Nham, Kuang, Gao, Baek, Mittal, Wahid, Gergely, Ni, Feldman,
  Muir, Lamblin, Macherey, Dyer, Kilpatrick, Campos, Bhutani, Fort, Ahmad,
  Severyn, Chatziprimou, Ferludin, Dimarco, Kusupati, Heyward, Bahir, Villela,
  Millican, Marcus, Bahargam, Unlu, Roth, Wei, Gopal, Ghoshal, Lee, Lin, Lees,
  Lee, Hosseini, Fan, Neel, Wu, Altun, Cai, Piqueras, Woodward, Bissacco,
  Haykal, Bordbar, Sundaram, Hodkinson, Toyama, Polovets, Myers, Sinha,
  Levinboim, Krishnakumar, Chhaparia, Sholokhova, Gundavarapu, Jawahar,
  Qureshi, Hu, Momchev, Rahtz, Wu, S, Dhamdhere, Guo, Gupta, Eslami, Schain,
  Blokzijl, Welling, Orr, Bolelli, Perez-Nieves, Sirotenko, Prasad, Kar, Pigem,
  Terzi, Weisz, Ghosh, Mavalankar, Madeka, Daugaard, Adam, Shah, Berman, Tran,
  Baker, Andrejczuk, Chole, Raboshchuk, Mirzazadeh, Kagohara, Wu, Schallhart,
  Orlando, Wang, Rrustemi, Xiong, Liu, Vezer, Ramsden, yiin Chang, Mudgal, Li,
  Vieillard, Hoshen, Ahmad, Slone, Hua, Potikha, Rossini, Stritar, Prakash,
  Wang, Dong, Nazari, Nehoran, Tekelioglu, Li, Badola, Funkhouser, Li, Yerram,
  Ganeshan, Formoso, Langner, Shi, Li, Yamamori, Panda, Saade, Scarpati,
  Breaux, Carey, Zhou, Hsieh, Bridgers, Butryna, Gupta, Tulsyan, Woo, Eltyshev,
  Grathwohl, Parks, Benjamin, Panigrahy, Dodhia, Freitas, Sauer, Song, Alet,
  Tolins, Paduraru, Zhou, Albert, Zhang, Shu, Bansal, Nguyen, Globerson, Xiao,
  Manyika, Hennigan, Rong, Matak, Bakalov, Sharma, Sinopalnikov, Pierson,
  Roller, Brown, Gao, Fukuzawa, Ghafouri, Vassigh, Barr, Wang, Korsun, Jayaram,
  Ren, Zaman, Khan, Lunts, Deutsch, Uthus, Katz, Samsikova, Khalifa, Sethi,
  Sun, Tang, Alon, Luo, Yu, Nayyar, Petrini, Truong, Hellendoorn, Chinaev,
  Alberti, Wang, Hu, Mirrokni, Balashankar, Aharon, Mehta, Iscen, Kready,
  Manning, Mohananey, Chen, Tripathi, Wu, Petrovski, Hwang, Baeuml,
  Chandrakaladharan, Liu, Coaguila, Chen, Ma, Tafti, Tatineni, Spitz, Ye,
  Vicol, Rosca, Puigdomènech, Yahav, Ghemawat, Lin, Kirk, Nabulsi, Brin,
  Bohnet, Caluwaerts, Veerubhotla, Zheng, Dai, Petrov, Xu, Mehran, Xu,
  Zintgraf, Choi, Hombaiah, Thoppilan, Reddi, Lew, Li, Webster, Sawhney,
  Lamprou, Shakeri, Lunayach, Chen, Bagri, Salcianu, Chen, Donchev, Magister,
  Nørly, Rodrigues, Izo, Noga, Zou, Köppe, Zhou, Lee, Long, Eisenbud, Chen,
  Schenck, To, Zhong, Taropa, Truong, Levy, Martins, Zhang, Semturs, Zhang,
  Yakubovich, Moreno, McConnaughey, Lu, Redmond, Weerts, Bitton, Refice,
  Lacasse, Conmy, Tallec, Odell, Forbes-Pollard, Socala, Hoech, Kohli, Walton,
  Wang, Sazanovich, Zhu, Kapishnikov, Galt, Denton, Murdoch, Sikora, Mohamed,
  Wei, First, McConnell, Cobo, Qin, Avrahami, Balle, Watanabe, Louis, Kraft,
  Ariafar, Gu, Rives, Yoon, Rusu, Cobon-Kerr, Hahn, Luo, Yuvein, Zhu, Ahuja,
  Benenson, Kaufman, Yu, Hightower, Zhang, Ni, Hendricks, Wang, Yona, Jain,
  Barrio, Bhupatiraju, Velusamy, Dafoe, Riedel, Thomas, Yuan, Bellaiche,
  Panthaplackel, Kloboves, Jauhari, Akbulut, Davchev, Gladchenko, Madras,
  Chuklin, Hill, Yuan, Madhavan, Leonhard, Scandinaro, Chen, Niu, Douillard,
  Damoc, Onoe, Pedregosa, Bertsch, Leichner, Pagadora, Malmaud, Ponda, Twigg,
  Duzhyi, Shen, Wang, Garg, Chen, Evci, Lee, Liu, Kojima, Yamaguchi, Rajendran,
  Piergiovanni, Rajendran, Fornoni, Ibagon, Ragan, Khan, Blitzer, Bunner, Sun,
  Kosakai, Lundberg, Elue, Guu, Park, Park, Narayanaswamy, Wu, Mudigonda, Cohn,
  Mu, Kumar, Graesser, Zhang, Killam, Zhuang, Giménez, Jishi, Ley-Wild, Zhai,
  Osawa, Cedillo, Liu, Upadhyay, Sieniek, Sharma, Paine, Angelova, Addepalli,
  Parada, Majumder, Lamp, Kumar, Deng, Myaskovsky, Sabolić, Dudek, York,
  de~Chaumont~Quitry, Nie, Cattle, Gunjan, Piot, Khawaja, Bang, Wang,
  Khodadadeh, R, Rawlani, Powell, Lee, Griesser, Oh, Magalhaes, Li, Tokumine,
  Vogel, Hsu, BC, Jindal, Cohen, Yang, Yuan, de~Cesare, Bruguier, Xu, Roy,
  Jacovi, Belov, Arya, Meadowlark, Cohen-Ganor, Ye, Morris-Suzuki, Banzal,
  Song, Ponnuramu, Zhang, Scrivener, Zaiem, Rochman, Han, Ghazi, Lee, Drath,
  Suo, Girgis, Shenoy, Nguyen, Eck, Gupta, Yan, Carreira, Gulati, Sang,
  Mirylenka, Cooney, Chou, Ling, Fan, Coleman, Tubone, Kumar, Baldridge,
  Hernandez-Campos, Lazaridou, Besley, Yona, Bulut, Wellens, Pierigiovanni,
  George, Green, Han, Tao, Clark, You, Abdolmaleki, Fu, Chen, Chaugule,
  Chandorkar, Rahman, Thompson, Koanantakool, Bernico, Ren, Vlasov,
  Vassilvitskii, Kula, Liang, Kim, Huang, Ye, Lepikhin, and
  Helmholz]{comanici2025gemini25pushingfrontier}
Gheorghe Comanici, Eric Bieber, Mike Schaekermann, Ice Pasupat, Noveen
  Sachdeva, Inderjit Dhillon, Marcel Blistein, Ori Ram, Dan Zhang, Evan Rosen,
  Luke Marris, Sam Petulla, Colin Gaffney, Asaf Aharoni, Nathan Lintz,
  Tiago~Cardal Pais, Henrik Jacobsson, Idan Szpektor, Nan-Jiang Jiang, Krishna
  Haridasan, Ahmed Omran, Nikunj Saunshi, Dara Bahri, Gaurav Mishra, Eric Chu,
  Toby Boyd, Brad Hekman, Aaron Parisi, Chaoyi Zhang, Kornraphop Kawintiranon,
  Tania Bedrax-Weiss, Oliver Wang, Ya~Xu, Ollie Purkiss, Uri Mendlovic, Ilaï
  Deutel, Nam Nguyen, Adam Langley, Flip Korn, Lucia Rossazza, Alexandre Ramé,
  Sagar Waghmare, Helen Miller, Nathan Byrd, Ashrith Sheshan, Raia Hadsell,
  Sangnie Bhardwaj, Pawel Janus, Tero Rissa, Dan Horgan, Alvin Abdagic, Lior
  Belenki, James Allingham, Anima Singh, Theo Guidroz, Srivatsan Srinivasan,
  Herman Schmit, Kristen Chiafullo, Andre Elisseeff, Nilpa Jha, Prateek Kolhar,
  Leonard Berrada, Frank Ding, Xiance Si, Shrestha~Basu Mallick, Franz Och,
  Sofia Erell, Eric Ni, Tejasi Latkar, Sherry Yang, Petar Sirkovic, Ziqiang
  Feng, Robert Leland, Rachel Hornung, Gang Wu, Charles Blundell, Hamidreza
  Alvari, Po-Sen Huang, Cathy Yip, Sanja Deur, Li~Liu, Gabriela Surita, Pablo
  Duque, Dima Damen, Johnson Jia, Arthur Guez, Markus Mircea, Animesh Sinha,
  Alberto Magni, Paweł Stradomski, Tal Marian, Vlado Galić, Wenhu Chen,
  Hisham Husain, Achintya Singhal, Dominik Grewe, François-Xavier Aubet,
  Shuang Song, Lorenzo Blanco, Leland Rechis, Lewis Ho, Rich Munoz, Kelvin
  Zheng, Jessica Hamrick, Kevin Mather, Hagai Taitelbaum, Eliza Rutherford, Yun
  Lei, Kuangyuan Chen, Anand Shukla, Erica Moreira, Eric Doi, Berivan Isik, Nir
  Shabat, Dominika Rogozińska, Kashyap Kolipaka, Jason Chang, Eugen Vušak,
  Srinivasan Venkatachary, Shadi Noghabi, Tarun Bharti, Younghoon Jun,
  Aleksandr Zaks, Simon Green, Jeshwanth Challagundla, William Wong, Muqthar
  Mohammad, Dean Hirsch, Yong Cheng, Iftekhar Naim, Lev Proleev, Damien
  Vincent, Aayush Singh, Maxim Krikun, Dilip Krishnan, Zoubin Ghahramani, Aviel
  Atias, Rajeev Aggarwal, Christo Kirov, Dimitrios Vytiniotis, Christy Koh,
  Alexandra Chronopoulou, Pawan Dogra, Vlad-Doru Ion, Gladys Tyen, Jason Lee,
  Felix Weissenberger, Trevor Strohman, Ashwin Balakrishna, Jack Rae, Marko
  Velic, Raoul de~Liedekerke, Oded Elyada, Wentao Yuan, Canoee Liu, Lior Shani,
  Sergey Kishchenko, Bea Alessio, Yandong Li, Richard Song, Sam Kwei, Orion
  Jankowski, Aneesh Pappu, Youhei Namiki, Yenai Ma, Nilesh Tripuraneni, Colin
  Cherry, Marissa Ikonomidis, Yu-Cheng Ling, Colin Ji, Beka Westberg, Auriel
  Wright, Da~Yu, David Parkinson, Swaroop Ramaswamy, Jerome Connor,
  Soheil~Hassas Yeganeh, Snchit Grover, George Kenwright, Lubo Litchev, Chris
  Apps, Alex Tomala, Felix Halim, Alex Castro-Ros, Zefei Li, Anudhyan Boral,
  Pauline Sho, Michal Yarom, Eric Malmi, David Klinghoffer, Rebecca Lin, Alan
  Ansell, Pradeep~Kumar S, Shubin Zhao, Siqi Zuo, Adam Santoro, Heng-Tze Cheng,
  Solomon Demmessie, Yuchi Liu, Nicole Brichtova, Allie Culp, Nathaniel Braun,
  Dan Graur, Will Ng, Nikhil Mehta, Aaron Phillips, Patrik Sundberg, Varun
  Godbole, Fangyu Liu, Yash Katariya, David Rim, Mojtaba Seyedhosseini, Sean
  Ammirati, Jonas Valfridsson, Mahan Malihi, Timothy Knight, Andeep Toor,
  Thomas Lampe, Abe Ittycheriah, Lewis Chiang, Chak Yeung, Alexandre
  Fréchette, Jinmeng Rao, Huisheng Wang, Himanshu Srivastava, Richard Zhang,
  Rocky Rhodes, Ariel Brand, Dean Weesner, Ilya Figotin, Felix Gimeno, Rachana
  Fellinger, Pierre Marcenac, José Leal, Eyal Marcus, Victor Cotruta, Rodrigo
  Cabrera, Sheryl Luo, Dan Garrette, Vera Axelrod, Sorin Baltateanu, David
  Barker, Dongkai Chen, Horia Toma, Ben Ingram, Jason Riesa, Chinmay Kulkarni,
  Yujing Zhang, Hongbin Liu, Chao Wang, Martin Polacek, Will Wu, Kai Hui,
  Adrian~N Reyes, Yi~Su, Megan Barnes, Ishaan Malhi, Anfal Siddiqui, Qixuan
  Feng, Mihai Damaschin, Daniele Pighin, Andreas Steiner, Samuel Yang,
  Ramya~Sree Boppana, Simeon Ivanov, Arun Kandoor, Aditya Shah, Asier Mujika,
  Da~Huang, Christopher~A. Choquette-Choo, Mohak Patel, Tianhe Yu, Toni
  Creswell, Jerry, Liu, Catarina Barros, Yasaman Razeghi, Aurko Roy, Phil
  Culliton, Binbin Xiong, Jiaqi Pan, Thomas Strohmann, Tolly Powell, Babi Seal,
  Doug DeCarlo, Pranav Shyam, Kaan Katircioglu, Xuezhi Wang, Cassidy Hardin,
  Immanuel Odisho, Josef Broder, Oscar Chang, Arun Nair, Artem Shtefan, Maura
  O'Brien, Manu Agarwal, Sahitya Potluri, Siddharth Goyal, Amit Jhindal,
  Saksham Thakur, Yury Stuken, James Lyon, Kristina Toutanova, Fangxiaoyu Feng,
  Austin Wu, Ben Horn, Alek Wang, Alex Cullum, Gabe Taubman, Disha Shrivastava,
  Chongyang Shi, Hamish Tomlinson, Roma Patel, Tao Tu, Ada~Maksutaj Oflazer,
  Francesco Pongetti, Mingyao Yang, Adrien~Ali Taïga, Vincent Perot, Nuo~Wang
  Pierse, Feng Han, Yoel Drori, Iñaki Iturrate, Ayan Chakrabarti, Legg Yeung,
  Dave Dopson, Yi~ting Chen, Apoorv Kulshreshtha, Tongfei Guo, Philip Pham, Tal
  Schuster, Junquan Chen, Alex Polozov, Jinwei Xing, Huanjie Zhou, Praneeth
  Kacham, Doron Kukliansky, Antoine Miech, Sergey Yaroshenko, Ed~Chi, Sholto
  Douglas, Hongliang Fei, Mathieu Blondel, Preethi Myla, Lior Madmoni, Xing Wu,
  Daniel Keysers, Kristian Kjems, Isabela Albuquerque, Lijun Yu, Joel D'sa,
  Michelle Plantan, Vlad Ionescu, Jaume~Sanchez Elias, Abhirut Gupta,
  Manish~Reddy Vuyyuru, Fred Alcober, Tong Zhou, Kaiyang Ji, Florian Hartmann,
  Subha Puttagunta, Hugo Song, Ehsan Amid, Anca Stefanoiu, Andrew Lee, Paul
  Pucciarelli, Emma Wang, Amit Raul, Slav Petrov, Isaac Tian, Valentin Anklin,
  Nana Nti, Victor Gomes, Max Schumacher, Grace Vesom, Alex Panagopoulos,
  Konstantinos Bousmalis, Daniel Andor, Josh Jacob, Yuan Zhang, Bill Rosgen,
  Matija Kecman, Matthew Tung, Alexandra Belias, Noah Goodman, Paul Covington,
  Brian Wieder, Nikita Saxena, Elnaz Davoodi, Muhuan Huang, Sharath Maddineni,
  Vincent Roulet, Folawiyo Campbell-Ajala, Pier~Giuseppe Sessa, Xintian, Wu,
  Guangda Lai, Paul Collins, Alex Haig, Vytenis Sakenas, Xiaowei Xu, Marissa
  Giustina, Laurent~El Shafey, Pichi Charoenpanit, Shefali Garg, Joshua
  Ainslie, Boone Severson, Montse~Gonzalez Arenas, Shreya Pathak, Sujee
  Rajayogam, Jie Feng, Michiel Bakker, Sheng Li, Nevan Wichers, Jamie Rogers,
  Xinyang Geng, Yeqing Li, Rolf Jagerman, Chao Jia, Nadav Olmert, David Sharon,
  Matthew Mauger, Sandeep Mariserla, Hongxu Ma, Megha Mohabey, Kyuyeun Kim,
  Alek Andreev, Scott Pollom, Juliette Love, Vihan Jain, Priyanka Agrawal,
  Yannick Schroecker, Alisa Fortin, Manfred Warmuth, Ji~Liu, Andrew Leach,
  Irina Blok, Ganesh~Poomal Girirajan, Roee Aharoni, Benigno Uria, Andrei
  Sozanschi, Dan Goldberg, Lucian Ionita, Marco~Tulio Ribeiro, Martin Zlocha,
  Vighnesh Birodkar, Sami Lachgar, Liangzhe Yuan, Himadri Choudhury, Matt
  Ginsberg, Fei Zheng, Gregory Dibb, Emily Graves, Swachhand Lokhande, Gabriel
  Rasskin, George-Cristian Muraru, Corbin Quick, Sandeep Tata, Pierre Sermanet,
  Aditya Chawla, Itay Karo, Yan Wang, Susan Zhang, Orgad Keller, Anca Dragan,
  Guolong Su, Ian Chou, Xi~Liu, Yiqing Tao, Shruthi Prabhakara, Marc Wilson,
  Ruibo Liu, Shibo Wang, Georgie Evans, David Du, Alfonso Castaño, Gautam
  Prasad, Mona~El Mahdy, Sebastian Gerlach, Machel Reid, Jarrod Kahn, Amir
  Zait, Thanumalayan~Sankaranarayana Pillai, Thatcher Ulrich, Guanyu Wang, Jan
  Wassenberg, Efrat Farkash, Kiran Yalasangi, Congchao Wang, Maria Bauza, Simon
  Bucher, Ting Liu, Jun Yan, Gary Leung, Vikas Sindhwani, Parker Barnes, Avi
  Singh, Ivan Jurin, Jichuan Chang, Niket~Kumar Bhumihar, Sivan Eiger, Gui
  Citovsky, Ben Withbroe, Zhang Li, Siyang Xue, Niccolò~Dal Santo, Georgi
  Stoyanov, Yves Raimond, Steven Zheng, Yilin Gao, Vít Listík, Sławek
  Kwasiborski, Rachel Saputro, Adnan Ozturel, Ganesh Mallya, Kushal Majmundar,
  Ross West, Paul Caron, Jinliang Wei, Lluis Castrejon, Sharad Vikram, Deepak
  Ramachandran, Nikhil Dhawan, Jiho Park, Sara Smoot, George van~den Driessche,
  Yochai Blau, Chase Malik, Wei Liang, Roy Hirsch, Cicero~Nogueira dos Santos,
  Eugene Weinstein, Aäron van~den Oord, Sid Lall, Nicholas FitzGerald, Zixuan
  Jiang, Xuan Yang, Dale Webster, Ali Elqursh, Aedan Pope, Georges Rotival,
  David Raposo, Wanzheng Zhu, Jeff Dean, Sami Alabed, Dustin Tran, Arushi
  Gupta, Zach Gleicher, Jessica Austin, Edouard Rosseel, Megh Umekar, Dipanjan
  Das, Yinghao Sun, Kai Chen, Karolis Misiunas, Xiang Zhou, Yixian Di, Alyssa
  Loo, Josh Newlan, Bo~Li, Vinay Ramasesh, Ying Xu, Alex Chen, Sudeep Gandhe,
  Radu Soricut, Nikita Gupta, Shuguang Hu, Seliem El-Sayed, Xavier Garcia, Idan
  Brusilovsky, Pu-Chin Chen, Andrew Bolt, Lu~Huang, Alex Gurney, Zhiying Zhang,
  Alexander Pritzel, Jarek Wilkiewicz, Bryan Seybold, Bhargav~Kanagal Shamanna,
  Felix Fischer, Josef Dean, Karan Gill, Ross Mcilroy, Abhishek Bhowmick,
  Jeremy Selier, Antoine Yang, Derek Cheng, Vladimir Magay, Jie Tan, Dhriti
  Varma, Christian Walder, Tomas Kocisky, Ryo Nakashima, Paul Natsev, Mike
  Kwong, Ionel Gog, Chiyuan Zhang, Sander Dieleman, Thomas Jimma, Andrey
  Ryabtsev, Siddhartha Brahma, David Steiner, Dayou Du, Ante Žužul, Mislav
  Žanić, Mukund Raghavachari, Willi Gierke, Zeyu Zheng, Dessie Petrova, Yann
  Dauphin, Yuchuan Liu, Ido Kessler, Steven Hand, Chris Duvarney, Seokhwan Kim,
  Hyo Lee, Léonard Hussenot, Jeffrey Hui, Josh Smith, Deepali Jain, Jiawei
  Xia, Gaurav~Singh Tomar, Keyvan Amiri, Du~Phan, Fabian Fuchs, Tobias Weyand,
  Nenad Tomasev, Alexandra Cordell, Xin Liu, Jonathan Mallinson, Pankaj Joshi,
  Andy Crawford, Arun Suggala, Steve Chien, Nick Fernando, Mariella
  Sanchez-Vargas, Duncan Williams, Phil Crone, Xiyang Luo, Igor Karpov, Jyn
  Shan, Terry Thurk, Robin Strudel, Paul Voigtlaender, Piyush Patil, Tim Dozat,
  Ali Khodaei, Sahil Singla, Piotr Ambroszczyk, Qiyin Wu, Yifan Chang, Brian
  Roark, Chaitra Hegde, Tianli Ding, Angelos Filos, Zhongru Wu, André~Susano
  Pinto, Shuang Liu, Saarthak Khanna, Aditya Pandey, Siobhan Mcloughlin, Qiujia
  Li, Sam Haves, Allan Zhou, Elena Buchatskaya, Isabel Leal, Peter de~Boursac,
  Nami Akazawa, Nina Anderson, Terry Chen, Krishna Somandepalli, Chen Liang,
  Sheela Goenka, Stephanie Winkler, Alexander Grushetsky, Yifan Ding, Jamie
  Smith, Fan Ye, Jordi Pont-Tuset, Eric Li, Ruichao Li, Tomer Golany, Dawid
  Wegner, Tao Jiang, Omer Barak, Yuan Shangguan, Eszter Vértes, Renee Wong,
  Jörg Bornschein, Alex Tudor, Michele Bevilacqua, Tom Schaul, Ankit~Singh
  Rawat, Yang Zhao, Kyriakos Axiotis, Lei Meng, Cory McLean, Jonathan Lai,
  Jennifer Beattie, Nate Kushman, Yaxin Liu, Blair Kutzman, Fiona Lang,
  Jingchen Ye, Praneeth Netrapalli, Pushkar Mishra, Myriam Khan, Megha Goel,
  Rob Willoughby, David Tian, Honglei Zhuang, JD~Chen, Zak Tsai, Tasos
  Kementsietsidis, Arjun Khare, James Keeling, Keyang Xu, Nathan Waters,
  Florent Altché, Ashok Popat, Bhavishya Mittal, David Saxton, Dalia~El
  Badawy, Michael Mathieu, Zheng Zheng, Hao Zhou, Nishant Ranka, Richard Shin,
  Qingnan Duan, Tim Salimans, Ioana Mihailescu, Uri Shaham, Ming-Wei Chang,
  Yannis Assael, Nishanth Dikkala, Martin Izzard, Vincent Cohen-Addad, Cat
  Graves, Vlad Feinberg, Grace Chung, DJ~Strouse, Danny Karmon, Sahand
  Sharifzadeh, Zoe Ashwood, Khiem Pham, Jon Blanton, Alex Vasiloff, Jarred
  Barber, Mark Geller, Aurick Zhou, Fedir Zubach, Tzu-Kuo Huang, Lei Zhang,
  Himanshu Gupta, Matt Young, Julia Proskurnia, Ronny Votel, Valentin Gabeur,
  Gabriel Barcik, Aditya Tripathi, Hongkun Yu, Geng Yan, Beer Changpinyo, Filip
  Pavetić, Amy Coyle, Yasuhisa Fujii, Jorge~Gonzalez Mendez, Tianhao Zhou,
  Harish Rajamani, Blake Hechtman, Eddie Cao, Da-Cheng Juan, Yi-Xuan Tan,
  Valentin Dalibard, Yilun Du, Natalie Clay, Kaisheng Yao, Wenhao Jia, Dimple
  Vijaykumar, Yuxiang Zhou, Xinyi Bai, Wei-Chih Hung, Steven Pecht, Georgi
  Todorov, Nikhil Khadke, Pramod Gupta, Preethi Lahoti, Arnaud Autef, Karthik
  Duddu, James Lee-Thorp, Alexander Bykovsky, Tautvydas Misiunas, Sebastian
  Flennerhag, Santhosh Thangaraj, Jed McGiffin, Zack Nado, Markus Kunesch,
  Andreas Noever, Amir Hertz, Marco Liang, Victor Stone, Evan Palmer, Samira
  Daruki, Arijit Pramanik, Siim Põder, Austin Kyker, Mina Khan, Evgeny
  Sluzhaev, Marvin Ritter, Avraham Ruderman, Wenlei Zhou, Chirag Nagpal, Kiran
  Vodrahalli, George Necula, Paul Barham, Ellie Pavlick, Jay Hartford, Izhak
  Shafran, Long Zhao, Maciej Mikuła, Tom Eccles, Hidetoshi Shimokawa, Kanav
  Garg, Luke Vilnis, Hanwen Chen, Ilia Shumailov, Kuang-Huei Lee, Abdelrahman
  Abdelhamed, Meiyan Xie, Vered Cohen, Ester Hlavnova, Dan Malkin, Chawin
  Sitawarin, James Lottes, Pauline Coquinot, Tianli Yu, Sandeep Kumar, Jingwei
  Zhang, Aroma Mahendru, Zafarali Ahmed, James Martens, Tao Chen, Aviel Boag,
  Daiyi Peng, Coline Devin, Arseniy Klimovskiy, Mary Phuong, Danny Vainstein,
  Jin Xie, Bhuvana Ramabhadran, Nathan Howard, Xinxin Yu, Gitartha Goswami,
  Jingyu Cui, Sam Shleifer, Mario Pinto, Chih-Kuan Yeh, Ming-Hsuan Yang, Sara
  Javanmardi, Dan Ethier, Chace Lee, Jordi Orbay, Suyog Kotecha, Carla
  Bromberg, Pete Shaw, James Thornton, Adi~Gerzi Rosenthal, Shane Gu, Matt
  Thomas, Ian Gemp, Aditya Ayyar, Asahi Ushio, Aarush Selvan, Joel Wee, Chenxi
  Liu, Maryam Majzoubi, Weiren Yu, Jake Abernethy, Tyler Liechty, Renke Pan,
  Hoang Nguyen, Qiong, Hu, Sarah Perrin, Abhinav Arora, Emily Pitler, Weiyi
  Wang, Kaushik Shivakumar, Flavien Prost, Ben Limonchik, Jing Wang, Yi~Gao,
  Timothee Cour, Shyamal Buch, Huan Gui, Maria Ivanova, Philipp Neubeck, Kelvin
  Chan, Lucy Kim, Huizhong Chen, Naman Goyal, Da-Woon Chung, Lu~Liu, Yao Su,
  Anastasia Petrushkina, Jiajun Shen, Armand Joulin, Yuanzhong Xu, Stein~Xudong
  Lin, Yana Kulizhskaya, Ciprian Chelba, Shobha Vasudevan, Eli Collins,
  Vasilisa Bashlovkina, Tony Lu, Doug Fritz, Jongbin Park, Yanqi Zhou, Chen Su,
  Richard Tanburn, Mikhail Sushkov, Mitchelle Rasquinha, Jinning Li, Jennifer
  Prendki, Yiming Li, Pallavi LV, Shriya Sharma, Hen Fitoussi, Hui Huang,
  Andrew Dai, Phuong Dao, Mike Burrows, Henry Prior, Danfeng Qin, Golan Pundak,
  Lars~Lowe Sjoesund, Art Khurshudov, Zhenkai Zhu, Albert Webson, Elizabeth
  Kemp, Tat Tan, Saurabh Agrawal, Susie Sargsyan, Liqun Cheng, Jim Stephan, Tom
  Kwiatkowski, David Reid, Arunkumar Byravan, Assaf~Hurwitz Michaely, Nicolas
  Heess, Luowei Zhou, Sonam Goenka, Viral Carpenter, Anselm Levskaya, Bo~Wang,
  Reed Roberts, Rémi Leblond, Sharat Chikkerur, Stav Ginzburg, Max Chang,
  Robert Riachi, Chuqiao, Xu, Zalán Borsos, Michael Pliskin, Julia Pawar,
  Morgane Lustman, Hannah Kirkwood, Ankit Anand, Aditi Chaudhary, Norbert Kalb,
  Kieran Milan, Sean Augenstein, Anna Goldie, Laurel Prince, Karthik Raman,
  Yanhua Sun, Vivian Xia, Aaron Cohen, Zhouyuan Huo, Josh Camp, Seher Ellis,
  Lukas Zilka, David~Vilar Torres, Lisa Patel, Sho Arora, Betty Chan, Jonas
  Adler, Kareem Ayoub, Jacky Liang, Fayaz Jamil, Jiepu Jiang, Simon
  Baumgartner, Haitian Sun, Yael Karov, Yaroslav Akulov, Hui Zheng, Irene Cai,
  Claudio Fantacci, James Rubin, Alex~Rav Acha, Mengchao Wang, Nina D'Souza,
  Rohit Sathyanarayana, Shengyang Dai, Simon Rowe, Andrey Simanovsky, Omer
  Goldman, Yuheng Kuang, Xiaoyue Pan, Andrew Rosenberg, Tania Rojas-Esponda,
  Praneet Dutta, Amy Zeng, Irina Jurenka, Greg Farquhar, Yamini Bansal, Shariq
  Iqbal, Becca Roelofs, Ga-Young Joung, Parker Beak, Changwan Ryu, Ryan Poplin,
  Yan Wu, Jean-Baptiste Alayrac, Senaka Buthpitiya, Olaf Ronneberger, Caleb
  Habtegebriel, Wei Li, Paul Cavallaro, Aurora Wei, Guy Bensky, Timo Denk,
  Harish Ganapathy, Jeff Stanway, Pratik Joshi, Francesco Bertolini, Jessica
  Lo, Olivia Ma, Zachary Charles, Geta Sampemane, Himanshu Sahni, Xu~Chen,
  Harry Askham, David Gaddy, Peter Young, Jiewen Tan, Matan Eyal, Arthur
  Bražinskas, Li~Zhong, Zhichun Wu, Mark Epstein, Kai Bailey, Andrew Hard,
  Kamyu Lee, Sasha Goldshtein, Alex Ruiz, Mohammed Badawi, Matthias
  Lochbrunner, JK~Kearns, Ashley Brown, Fabio Pardo, Theophane Weber, Haichuan
  Yang, Pan-Pan Jiang, Berkin Akin, Zhao Fu, Marcus Wainwright, Chi Zou, Meenu
  Gaba, Pierre-Antoine Manzagol, Wendy Kan, Yang Song, Karina Zainullina, Rui
  Lin, Jeongwoo Ko, Salil Deshmukh, Apoorv Jindal, James Svensson, Divya Tyam,
  Heri Zhao, Christine Kaeser-Chen, Scott Baird, Pooya Moradi, Jamie Hall,
  Qiuchen Guo, Vincent Tsang, Bowen Liang, Fernando Pereira, Suhas Ganesh, Ivan
  Korotkov, Jakub Adamek, Sridhar Thiagarajan, Vinh Tran, Charles Chen, Chris
  Tar, Sanil Jain, Ishita Dasgupta, Taylan Bilal, David Reitter, Kai Zhao,
  Giulia Vezzani, Yasmin Gehman, Pulkit Mehta, Lauren Beltrone, Xerxes
  Dotiwalla, Sergio Guadarrama, Zaheer Abbas, Stefani Karp, Petko Georgiev,
  Chun-Sung Ferng, Marc Brockschmidt, Liqian Peng, Christoph Hirnschall, Vikas
  Verma, Yingying Bi, Ying Xiao, Avigail Dabush, Kelvin Xu, Phil Wallis,
  Randall Parker, Qifei Wang, Yang Xu, Ilkin Safarli, Dinesh Tewari, Yin Zhang,
  Seungyeon Kim, Andrea Gesmundo, Mackenzie Thomas, Sergey Levi, Ahmed
  Chowdhury, Kanishka Rao, Peter Garst, Sam Conway-Rahman, Helen Ran, Kay
  McKinney, Zhisheng Xiao, Wenhao Yu, Rohan Agrawal, Axel Stjerngren, Catalin
  Ionescu, Jingjing Chen, Vivek Sharma, Justin Chiu, Fei Liu, Ken Franko,
  Clayton Sanford, Xingyu Cai, Paul Michel, Sanjay Ganapathy, Jane Labanowski,
  Zachary Garrett, Ben Vargas, Sean Sun, Bryan Gale, Thomas Buschmann,
  Guillaume Desjardins, Nimesh Ghelani, Palak Jain, Mudit Verma, Chulayuth
  Asawaroengchai, Julian Eisenschlos, Jitendra Harlalka, Hideto Kazawa, Don
  Metzler, Joshua Howland, Ying Jian, Jake Ades, Viral Shah, Tynan Gangwani,
  Seungji Lee, Roman Ring, Steven~M. Hernandez, Dean Reich, Amer Sinha,
  Ashutosh Sathe, Joe Kovac, Ashleah Gill, Ajay Kannan, Andrea D'olimpio,
  Martin Sevenich, Jay Whang, Been Kim, Khe~Chai Sim, Jilin Chen, Jiageng
  Zhang, Shuba Lall, Yossi Matias, Bill Jia, Abe Friesen, Sara Nasso, Ashish
  Thapliyal, Bryan Perozzi, Ting Yu, Anna Shekhawat, Safeen Huda, Peter
  Grabowski, Eric Wang, Ashwin Sreevatsa, Hilal Dib, Mehadi Hassen, Parker
  Schuh, Vedrana Milutinovic, Chris Welty, Michael Quinn, Ali Shah, Bangju
  Wang, Gabe Barth-Maron, Justin Frye, Natalie Axelsson, Tao Zhu, Yukun Ma,
  Irene Giannoumis, Hanie Sedghi, Chang Ye, Yi~Luan, Kevin Aydin, Bilva
  Chandra, Vivek Sampathkumar, Ronny Huang, Victor Lavrenko, Ahmed Eleryan, Zhi
  Hong, Steven Hansen, Sara~Mc Carthy, Bidisha Samanta, Domagoj Ćevid, Xin
  Wang, Fangtao Li, Michael Voznesensky, Matt Hoffman, Andreas Terzis, Vikash
  Sehwag, Gil Fidel, Luheng He, Mu~Cai, Yanzhang He, Alex Feng, Martin
  Nikoltchev, Samrat Phatale, Jason Chase, Rory Lawton, Ming Zhang, Tom Ouyang,
  Manuel Tragut, Mehdi~Hafezi Manshadi, Arjun Narayanan, Jiaming Shen, Xu~Gao,
  Tolga Bolukbasi, Nick Roy, Xin Li, Daniel Golovin, Liviu Panait, Zhen Qin,
  Guangxing Han, Thomas Anthony, Sneha Kudugunta, Viorica Patraucean, Aniket
  Ray, Xinyun Chen, Xiaochen Yang, Tanuj Bhatia, Pranav Talluri, Alex Morris,
  Andrija Ražnatović, Bethanie Brownfield, James An, Sheng Peng, Patrick
  Kane, Ce~Zheng, Nico Duduta, Joshua Kessinger, James Noraky, Siqi Liu, Keran
  Rong, Petar Veličković, Keith Rush, Alex Goldin, Fanny Wei, Shiva
  Mohan~Reddy Garlapati, Caroline Pantofaru, Okwan Kwon, Jianmo Ni, Eric
  Noland, Julia~Di Trapani, Françoise Beaufays, Abhijit~Guha Roy, Yinlam Chow,
  Aybuke Turker, Geoffrey Cideron, Lantao Mei, Jon Clark, Qingyun Dou, Matko
  Bošnjak, Ralph Leith, Yuqing Du, Amir Yazdanbakhsh, Milad Nasr, Chester
  Kwak, Suraj~Satishkumar Sheth, Alex Kaskasoli, Ankesh Anand, Balaji
  Lakshminarayanan, Sammy Jerome, David Bieber, Chun-Te Chu, Alexandre Senges,
  Tianxiao Shen, Mukund Sridhar, Ndaba Ndebele, Benjamin Beyret, Shakir
  Mohamed, Mia Chen, Markus Freitag, Jiaxian Guo, Luyang Liu, Paul Roit, Heng
  Chen, Shen Yan, Tom Stone, JD~Co-Reyes, Jeremy Cole, Salvatore Scellato,
  Shekoofeh Azizi, Hadi Hashemi, Alicia Jin, Anand Iyer, Marcella Valentine,
  András György, Arun Ahuja, Daniel~Hernandez Diaz, Chen-Yu Lee, Nathan
  Clement, Weize Kong, Drew Garmon, Ishaan Watts, Kush Bhatia, Khyatti Gupta,
  Matt Miecnikowski, Hugo Vallet, Ankur Taly, Edward Loper, Saket Joshi, James
  Atwood, Jo~Chick, Mark Collier, Fotis Iliopoulos, Ryan Trostle, Beliz Gunel,
  Ramiro Leal-Cavazos, Arnar~Mar Hrafnkelsson, Michael Guzman, Xiaoen Ju, Andy
  Forbes, Jesse Emond, Kushal Chauhan, Ben Caine, Li~Xiao, Wenjun Zeng,
  Alexandre Moufarek, Daniel Murphy, Maya Meng, Nitish Gupta, Felix Riedel,
  Anil Das, Elijah Lawal, Shashi Narayan, Tiberiu Sosea, James Swirhun, Linda
  Friso, Behnam Neyshabur, Jing Lu, Sertan Girgin, Michael Wunder, Edouard
  Yvinec, Aroonalok Pyne, Victor Carbune, Shruti Rijhwani, Yang Guo, Tulsee
  Doshi, Anton Briukhov, Max Bain, Ayal Hitron, Xuanhui Wang, Ashish Gupta,
  Ke~Chen, Cosmo Du, Weiyang Zhang, Dhruv Shah, Arjun Akula, Max Dylla, Ashyana
  Kachra, Weicheng Kuo, Tingting Zou, Lily Wang, Luyao Xu, Jifan Zhu, Justin
  Snyder, Sachit Menon, Orhan Firat, Igor Mordatch, Yuan Yuan, Natalia
  Ponomareva, Rory Blevins, Lawrence Moore, Weijun Wang, Phil Chen, Martin
  Scholz, Artur Dwornik, Jason Lin, Sicheng Li, Diego Antognini, Te~I, Xiaodan
  Song, Matt Miller, Uday Kalra, Adam Raveret, Oscar Akerlund, Felix Wu, Andrew
  Nystrom, Namrata Godbole, Tianqi Liu, Hannah DeBalsi, Jewel Zhao, Buhuang
  Liu, Avi Caciularu, Lauren Lax, Urvashi Khandelwal, Victoria Langston, Eric
  Bailey, Silvio Lattanzi, Yufei Wang, Neel Kovelamudi, Sneha Mondal, Guru
  Guruganesh, Nan Hua, Ofir Roval, Paweł Wesołowski, Rishikesh Ingale,
  Jonathan Halcrow, Tim Sohn, Christof Angermueller, Bahram Raad, Eli
  Stickgold, Eva Lu, Alec Kosik, Jing Xie, Timothy Lillicrap, Austin Huang,
  Lydia~Lihui Zhang, Dominik Paulus, Clement Farabet, Alex Wertheim, Bing Wang,
  Rishabh Joshi, Chu ling Ko, Yonghui Wu, Shubham Agrawal, Lily Lin, XiangHai
  Sheng, Peter Sung, Tyler Breland-King, Christina Butterfield, Swapnil Gawde,
  Sumeet Singh, Qiao Zhang, Raj Apte, Shilpa Shetty, Adrian Hutter, Tao Li,
  Elizabeth Salesky, Federico Lebron, Jonni Kanerva, Michela Paganini, Arthur
  Nguyen, Rohith Vallu, Jan-Thorsten Peter, Sarmishta Velury, David Kao, Jay
  Hoover, Anna Bortsova, Colton Bishop, Shoshana Jakobovits, Alessandro
  Agostini, Alekh Agarwal, Chang Liu, Charles Kwong, Sasan Tavakkol, Ioana
  Bica, Alex Greve, Anirudh GP, Jake Marcus, Le~Hou, Tom Duerig, Rivka
  Moroshko, Dave Lacey, Andy Davis, Julien Amelot, Guohui Wang, Frank Kim,
  Theofilos Strinopoulos, Hui Wan, Charline~Le Lan, Shankar Krishnan, Haotian
  Tang, Peter Humphreys, Junwen Bai, Idan~Heimlich Shtacher, Diego Machado,
  Chenxi Pang, Ken Burke, Dangyi Liu, Renga Aravamudhan, Yue Song, Ed~Hirst,
  Abhimanyu Singh, Brendan Jou, Liang Bai, Francesco Piccinno, Chuyuan~Kelly
  Fu, Robin Alazard, Barak Meiri, Daniel Winter, Charlie Chen, Mingda Zhang,
  Jens Heitkaemper, John Lambert, Jinhyuk Lee, Alexander Frömmgen, Sergey
  Rogulenko, Pranav Nair, Paul Niemczyk, Anton Bulyenov, Bibo Xu, Hadar
  Shemtov, Morteza Zadimoghaddam, Serge Toropov, Mateo Wirth, Hanjun Dai,
  Sreenivas Gollapudi, Daniel Zheng, Alex Kurakin, Chansoo Lee, Kalesha
  Bullard, Nicolas Serrano, Ivana Balazevic, Yang Li, Johan Schalkwyk, Mark
  Murphy, Mingyang Zhang, Kevin Sequeira, Romina Datta, Nishant Agrawal,
  Charles Sutton, Nithya Attaluri, Mencher Chiang, Wael Farhan, Gregory
  Thornton, Kate Lin, Travis Choma, Hung Nguyen, Kingshuk Dasgupta, Dirk
  Robinson, Iulia Comşa, Michael Riley, Arjun Pillai, Basil Mustafa, Ben
  Golan, Amir Zandieh, Jean-Baptiste Lespiau, Billy Porter, David Ross,
  Sujeevan Rajayogam, Mohit Agarwal, Subhashini Venugopalan, Bobak Shahriari,
  Qiqi Yan, Hao Xu, Taylor Tobin, Pavel Dubov, Hongzhi Shi, Adrià Recasens,
  Anton Kovsharov, Sebastian Borgeaud, Lucio Dery, Shanthal Vasanth, Elena
  Gribovskaya, Linhai Qiu, Mahdis Mahdieh, Wojtek Skut, Elizabeth Nielsen,
  CJ~Zheng, Adams Yu, Carrie~Grimes Bostock, Shaleen Gupta, Aaron Archer, Chris
  Rawles, Elinor Davies, Alexey Svyatkovskiy, Tomy Tsai, Yoni Halpern,
  Christian Reisswig, Bartek Wydrowski, Bo~Chang, Joan Puigcerver, Mor~Hazan
  Taege, Jian Li, Eva Schnider, Xinjian Li, Dragos Dena, Yunhan Xu, Umesh
  Telang, Tianze Shi, Heiga Zen, Kyle Kastner, Yeongil Ko, Neesha Subramaniam,
  Aviral Kumar, Pete Blois, Zhuyun Dai, John Wieting, Yifeng Lu, Yoel Zeldes,
  Tian Xie, Anja Hauth, Alexandru Ţifrea, Yuqi Li, Sam El-Husseini, Dan
  Abolafia, Howard Zhou, Wen Ding, Sahra Ghalebikesabi, Carlos Guía, Andrii
  Maksai, Ágoston Weisz, Sercan Arik, Nick Sukhanov, Aga Świetlik, Xuhui Jia,
  Luo Yu, Weiyue Wang, Mark Brand, Dawn Bloxwich, Sean Kirmani, Zhe Chen, Alec
  Go, Pablo Sprechmann, Nithish Kannen, Alen Carin, Paramjit Sandhu, Isabel
  Edkins, Leslie Nooteboom, Jai Gupta, Loren Maggiore, Javad Azizi, Yael
  Pritch, Pengcheng Yin, Mansi Gupta, Danny Tarlow, Duncan Smith, Desi Ivanov,
  Mohammad Babaeizadeh, Ankita Goel, Satish Kambala, Grace Chu, Matej Kastelic,
  Michelle Liu, Hagen Soltau, Austin Stone, Shivani Agrawal, Min Kim, Kedar
  Soparkar, Srinivas Tadepalli, Oskar Bunyan, Rachel Soh, Arvind Kannan,
  DY~Kim, Blake~JianHang Chen, Afief Halumi, Sudeshna Roy, Yulong Wang, Olcan
  Sercinoglu, Gena Gibson, Sijal Bhatnagar, Motoki Sano, Daniel von Dincklage,
  Qingchun Ren, Blagoj Mitrevski, Mirek Olšák, Jennifer She, Carl Doersch,
  Jilei, Wang, Bingyuan Liu, Qijun Tan, Tamar Yakar, Tris Warkentin, Alex
  Ramirez, Carl Lebsack, Josh Dillon, Rajiv Mathews, Tom Cobley, Zelin Wu,
  Zhuoyuan Chen, Jon Simon, Swaroop Nath, Tara Sainath, Alexei Bendebury, Ryan
  Julian, Bharath Mankalale, Daria Ćurko, Paulo Zacchello, Adam~R. Brown,
  Kiranbir Sodhia, Heidi Howard, Sergi Caelles, Abhinav Gupta, Gareth Evans,
  Anna Bulanova, Lesley Katzen, Roman Goldenberg, Anton Tsitsulin, Joe Stanton,
  Benoit Schillings, Vitaly Kovalev, Corey Fry, Rushin Shah, Kuo Lin, Shyam
  Upadhyay, Cheng Li, Soroush Radpour, Marcello Maggioni, Jing Xiong, Lukas
  Haas, Jenny Brennan, Aishwarya Kamath, Nikolay Savinov, Arsha Nagrani, Trevor
  Yacovone, Ryan Kappedal, Kostas Andriopoulos, Li~Lao, YaGuang Li, Grigory
  Rozhdestvenskiy, Kazuma Hashimoto, Andrew Audibert, Sophia Austin, Daniel
  Rodriguez, Anian Ruoss, Garrett Honke, Deep Karkhanis, Xi~Xiong, Qing Wei,
  James Huang, Zhaoqi Leng, Vittal Premachandran, Stan Bileschi, Georgios
  Evangelopoulos, Thomas Mensink, Jay Pavagadhi, Denis Teplyashin, Paul Chang,
  Linting Xue, Garrett Tanzer, Sally Goldman, Kaushal Patel, Shixin Li, Jeremy
  Wiesner, Ivy Zheng, Ian Stewart-Binks, Jie Han, Zhi Li, Liangchen Luo, Karel
  Lenc, Mario Lučić, Fuzhao Xue, Ryan Mullins, Alexey Guseynov, Chung-Ching
  Chang, Isaac Galatzer-Levy, Adam Zhang, Garrett Bingham, Grace Hu, Ale
  Hartman, Yue Ma, Jordan Griffith, Alex Irpan, Carey Radebaugh, Summer Yue,
  Lijie Fan, Victor Ungureanu, Christina Sorokin, Hannah Teufel, Peiran Li,
  Rohan Anil, Dimitris Paparas, Todd Wang, Chu-Cheng Lin, Hui Peng, Megan Shum,
  Goran Petrovic, Demetra Brady, Richard Nguyen, Klaus Macherey, Zhihao Li,
  Harman Singh, Madhavi Yenugula, Mariko Iinuma, Xinyi Chen, Kavya Kopparapu,
  Alexey Stern, Shachi Dave, Chandu Thekkath, Florence Perot, Anurag Kumar,
  Fangda Li, Yang Xiao, Matthew Bilotti, Mohammad~Hossein Bateni, Isaac Noble,
  Lisa Lee, Amelio Vázquez-Reina, Julian Salazar, Xiaomeng Yang, Boyu Wang,
  Ela Gruzewska, Anand Rao, Sindhu Raghuram, Zheng Xu, Eyal Ben-David, Jieru
  Mei, Sid Dalmia, Zhaoyi Zhang, Yuchen Liu, Gagan Bansal, Helena Pankov,
  Steven Schwarcz, Andrea Burns, Christine Chan, Sumit Sanghai, Ricky Liang,
  Ethan Liang, Antoine He, Amy Stuart, Arun Narayanan, Yukun Zhu, Christian
  Frank, Bahar Fatemi, Amit Sabne, Oran Lang, Indro Bhattacharya, Shane Settle,
  Maria Wang, Brendan McMahan, Andrea Tacchetti, Livio~Baldini Soares, Majid
  Hadian, Serkan Cabi, Timothy Chung, Nikita Putikhin, Gang Li, Jeremy Chen,
  Austin Tarango, Henryk Michalewski, Mehran Kazemi, Hussain Masoom, Hila
  Sheftel, Rakesh Shivanna, Archita Vadali, Ramona Comanescu, Doug Reid, Joss
  Moore, Arvind Neelakantan, Michaël Sander, Jonathan Herzig, Aviv Rosenberg,
  Mostafa Dehghani, JD~Choi, Michael Fink, Reid Hayes, Eric Ge, Shitao Weng,
  Chia-Hua Ho, John Karro, Kalpesh Krishna, Lam~Nguyen Thiet, Amy Skerry-Ryan,
  Daniel Eppens, Marco Andreetto, Navin Sarma, Silvano Bonacina, Burcu~Karagol
  Ayan, Megha Nawhal, Zhihao Shan, Mike Dusenberry, Shantanu Thakoor, Sagar
  Gubbi, Duc~Dung Nguyen, Reut Tsarfaty, Samuel Albanie, Jovana Mitrović, Meet
  Gandhi, Bo-Juen Chen, Alessandro Epasto, Georgi Stephanov, Ye~Jin, Samuel
  Gehman, Aida Amini, Jack Weber, Feryal Behbahani, Shawn Xu, Miltos Allamanis,
  Xi~Chen, Myle Ott, Claire Sha, Michal Jastrzebski, Hang Qi, David Greene,
  Xinyi Wu, Abodunrinwa Toki, Daniel Vlasic, Jane Shapiro, Ragha Kotikalapudi,
  Zhe Shen, Takaaki Saeki, Sirui Xie, Albin Cassirer, Shikhar Bharadwaj,
  Tatsuya Kiyono, Srinadh Bhojanapalli, Elan Rosenfeld, Sam Ritter, Jieming
  Mao, João~Gabriel Oliveira, Zoltan Egyed, Bernd Bandemer, Emilio Parisotto,
  Keisuke Kinoshita, Juliette Pluto, Petros Maniatis, Steve Li, Yaohui Guo,
  Golnaz Ghiasi, Jean Tarbouriech, Srimon Chatterjee, Julie Jin, Katrina, Xu,
  Jennimaria Palomaki, Séb Arnold, Madhavi Sewak, Federico Piccinini, Mohit
  Sharma, Ben Albrecht, Sean Purser-haskell, Ashwin Vaswani, Chongyan Chen,
  Matheus Wisniewski, Qin Cao, John Aslanides, Nguyet~Minh Phu, Maximilian
  Sieb, Lauren Agubuzu, Anne Zheng, Daniel Sohn, Marco Selvi, Anders
  Andreassen, Krishan Subudhi, Prem Eruvbetine, Oliver Woodman, Tomas Mery,
  Sebastian Krause, Xiaoqi Ren, Xiao Ma, Jincheng Luo, Dawn Chen, Wei Fan,
  Henry Griffiths, Christian Schuler, Alice Li, Shujian Zhang, Jean-Michel
  Sarr, Shixin Luo, Riccardo Patana, Matthew Watson, Dani Naboulsi, Michael
  Collins, Sailesh Sidhwani, Emiel Hoogeboom, Sharon Silver, Emily Caveness,
  Xiaokai Zhao, Mikel Rodriguez, Maxine Deines, Libin Bai, Patrick Griffin,
  Marco Tagliasacchi, Emily Xue, Spandana~Raj Babbula, Bo~Pang, Nan Ding,
  Gloria Shen, Elijah Peake, Remi Crocker, Shubha~Srinivas Raghvendra, Danny
  Swisher, Woohyun Han, Richa Singh, Ling Wu, Vladimir Pchelin, Tsendsuren
  Munkhdalai, Dana Alon, Geoff Bacon, Efren Robles, Jannis Bulian, Melvin
  Johnson, George Powell, Felipe~Tiengo Ferreira, Yaoyiran Li, Frederik
  Benzing, Mihajlo Velimirović, Hubert Soyer, William Kong, Tony, Nguyên,
  Zhen Yang, Jeremiah Liu, Joost van Amersfoort, Daniel Gillick, Baochen Sun,
  Nathalie Rauschmayr, Katie Zhang, Serena Zhan, Tao Zhou, Alexey Frolov,
  Chengrun Yang, Denis Vnukov, Louis Rouillard, Hongji Li, Amol Mandhane, Nova
  Fallen, Rajesh Venkataraman, Clara~Huiyi Hu, Jennifer Brennan, Jenny Lee,
  Jerry Chang, Martin Sundermeyer, Zhufeng Pan, Rosemary Ke, Simon Tong, Alex
  Fabrikant, William Bono, Jindong Gu, Ryan Foley, Yiran Mao, Manolis Delakis,
  Dhruva Bhaswar, Roy Frostig, Nick Li, Avital Zipori, Cath Hope, Olga Kozlova,
  Swaroop Mishra, Josip Djolonga, Craig Schiff, Majd~Al Merey, Eleftheria
  Briakou, Peter Morgan, Andy Wan, Avinatan Hassidim, RJ~Skerry-Ryan, Kuntal
  Sengupta, Mary Jasarevic, Praveen Kallakuri, Paige Kunkle, Hannah Brennan,
  Tom Lieber, Hassan Mansoor, Julian Walker, Bing Zhang, Annie Xie, Goran
  Žužić, Adaeze Chukwuka, Alex Druinsky, Donghyun Cho, Rui Yao, Ferjad
  Naeem, Shiraz Butt, Eunyoung Kim, Zhipeng Jia, Mandy Jordan, Adam Lelkes,
  Mark Kurzeja, Sophie Wang, James Zhao, Andrew Over, Abhishek Chakladar,
  Marcel Prasetya, Neha Jha, Sriram Ganapathy, Yale Cong, Prakash Shroff, Carl
  Saroufim, Sobhan Miryoosefi, Mohamed Hammad, Tajwar Nasir, Weijuan Xi, Yang
  Gao, Young Maeng, Ben Hora, Chin-Yi Cheng, Parisa Haghani, Yoad Lewenberg,
  Caden Lu, Martin Matysiak, Naina Raisinghani, Huiyu Wang, Lexi Baugher, Rahul
  Sukthankar, Minh Giang, John Schultz, Noah Fiedel, Minmin Chen, Cheng-Chun
  Lee, Tapomay Dey, Hao Zheng, Shachi Paul, Celine Smith, Andy Ly, Yicheng
  Wang, Rishabh Bansal, Bartek Perz, Susanna Ricco, Stasha Blank, Vaishakh
  Keshava, Deepak Sharma, Marvin Chow, Kunal Lad, Komal Jalan, Simon Osindero,
  Craig Swanson, Jacob Scott, Anastasija Ilić, Xiaowei Li, Siddhartha~Reddy
  Jonnalagadda, Afzal~Shama Soudagar, Yan Xiong, Bat-Orgil Batsaikhan, Daniel
  Jarrett, Naveen Kumar, Maulik Shah, Matt Lawlor, Austin Waters, Mark Graham,
  Rhys May, Sabela Ramos, Sandra Lefdal, Zeynep Cankara, Nacho Cano, Brendan
  O'Donoghue, Jed Borovik, Frederick Liu, Jordan Grimstad, Mahmoud Alnahlawi,
  Katerina Tsihlas, Tom Hudson, Nikolai Grigorev, Yiling Jia, Terry Huang,
  Tobenna~Peter Igwe, Sergei Lebedev, Xiaodan Tang, Igor Krivokon, Frankie
  Garcia, Melissa Tan, Eric Jia, Peter Stys, Shikhar Vashishth, Yu~Liang,
  Balaji Venkatraman, Chenjie Gu, Anastasios Kementsietsidis, Chen Zhu,
  Junehyuk Jung, Yunfei Bai, Mohammad~Javad Hosseini, Faruk Ahmed, Aditya
  Gupta, Xin Yuan, Shereen Ashraf, Shitij Nigam, Gautam Vasudevan, Pranjal
  Awasthi, Adi~Mayrav Gilady, Zelda Mariet, Ramy Eskander, Haiguang Li, Hexiang
  Hu, Guillermo Garrido, Philippe Schlattner, George Zhang, Rohun Saxena, Petar
  Dević, Kritika Muralidharan, Ashwin Murthy, Yiqian Zhou, Min Choi, Arissa
  Wongpanich, Zhengdong Wang, Premal Shah, Yuntao Xu, Yiling Huang, Stephen
  Spencer, Alice Chen, James Cohan, Junjie Wang, Jonathan Tompson, Junru Wu,
  Ruba Haroun, Haiqiong Li, Blanca Huergo, Fan Yang, Tongxin Yin, James Wendt,
  Michael Bendersky, Rahma Chaabouni, Javier Snaider, Johan Ferret, Abhishek
  Jindal, Tara Thompson, Andrew Xue, Will Bishop, Shubham~Milind Phal, Archit
  Sharma, Yunhsuan Sung, Prabakar Radhakrishnan, Mo~Shomrat, Reeve Ingle,
  Roopali Vij, Justin Gilmer, Mihai~Dorin Istin, Sam Sobell, Yang Lu, Emily
  Nottage, Dorsa Sadigh, Jeremiah Willcock, Tingnan Zhang, Steve Xu, Sasha
  Brown, Katherine Lee, Gary Wang, Yun Zhu, Yi~Tay, Cheolmin Kim, Audrey
  Gutierrez, Abhanshu Sharma, Yongqin Xian, Sungyong Seo, Claire Cui, Elena
  Pochernina, Cip Baetu, Krzysztof Jastrzębski, Mimi Ly, Mohamed Elhawaty, Dan
  Suh, Eren Sezener, Pidong Wang, Nancy Yuen, George Tucker, Jiahao Cai,
  Zuguang Yang, Cindy Wang, Alex Muzio, Hai Qian, Jae Yoo, Derek Lockhart,
  Kevin~R. McKee, Mandy Guo, Malika Mehrotra, Artur Mendonça, Sanket~Vaibhav
  Mehta, Sherry Ben, Chetan Tekur, Jiaqi Mu, Muye Zhu, Victoria Krakovna,
  Hongrae Lee, AJ~Maschinot, Sébastien Cevey, HyunJeong Choe, Aijun Bai, Hansa
  Srinivasan, Derek Gasaway, Nick Young, Patrick Siegler, Dan Holtmann-Rice,
  Vihari Piratla, Kate Baumli, Roey Yogev, Alex Hofer, Hado van Hasselt,
  Svetlana Grant, Yuri Chervonyi, David Silver, Andrew Hogue, Ayushi Agarwal,
  Kathie Wang, Preeti Singh, Four Flynn, Josh Lipschultz, Robert David,
  Lizzetth Bellot, Yao-Yuan Yang, Long Le, Filippo Graziano, Kate Olszewska,
  Kevin Hui, Akanksha Maurya, Nikos Parotsidis, Weijie Chen, Tayo Oguntebi, Joe
  Kelley, Anirudh Baddepudi, Johannes Mauerer, Gregory Shaw, Alex Siegman, Lin
  Yang, Shravya Shetty, Subhrajit Roy, Yunting Song, Wojciech Stokowiec, Ryan
  Burnell, Omkar Savant, Robert Busa-Fekete, Jin Miao, Samrat Ghosh, Liam
  MacDermed, Phillip Lippe, Mikhail Dektiarev, Zach Behrman, Fabian Mentzer,
  Kelvin Nguyen, Meng Wei, Siddharth Verma, Chris Knutsen, Sudeep Dasari,
  Zhipeng Yan, Petr Mitrichev, Xingyu Wang, Virat Shejwalkar, Jacob Austin,
  Srinivas Sunkara, Navneet Potti, Yan Virin, Christian Wright, Gaël Liu,
  Oriana Riva, Etienne Pot, Greg Kochanski, Quoc Le, Gargi Balasubramaniam,
  Arka Dhar, Yuguo Liao, Adam Bloniarz, Divyansh Shukla, Elizabeth Cole, Jong
  Lee, Sheng Zhang, Sushant Kafle, Siddharth Vashishtha, Parsa Mahmoudieh,
  Grace Chen, Raphael Hoffmann, Pranesh Srinivasan, Agustin~Dal Lago, Yoav~Ben
  Shalom, Zi~Wang, Michael Elabd, Anuj Sharma, Junhyuk Oh, Suraj Kothawade,
  Maigo Le, Marianne Monteiro, Shentao Yang, Kaiz Alarakyia, Robert Geirhos,
  Diana Mincu, Håvard Garnes, Hayato Kobayashi, Soroosh Mariooryad, Kacper
  Krasowiak, Zhixin, Lai, Shibl Mourad, Mingqiu Wang, Fan Bu, Ophir Aharoni,
  Guanjie Chen, Abhimanyu Goyal, Vadim Zubov, Ankur Bapna, Elahe Dabir, Nisarg
  Kothari, Kay Lamerigts, Nicola~De Cao, Jeremy Shar, Christopher Yew, Nitish
  Kulkarni, Dre Mahaarachchi, Mandar Joshi, Zhenhai Zhu, Jared Lichtarge,
  Yichao Zhou, Hannah Muckenhirn, Vittorio Selo, Oriol Vinyals, Peter Chen,
  Anthony Brohan, Vaibhav Mehta, Sarah Cogan, Ruth Wang, Ty~Geri, Wei-Jen Ko,
  Wei Chen, Fabio Viola, Keshav Shivam, Lisa Wang, Madeleine~Clare Elish,
  Raluca~Ada Popa, Sébastien Pereira, Jianqiao Liu, Raphael Koster, Donnie
  Kim, Gufeng Zhang, Sayna Ebrahimi, Partha Talukdar, Yanyan Zheng, Petra
  Poklukar, Ales Mikhalap, Dale Johnson, Anitha Vijayakumar, Mark Omernick,
  Matt Dibb, Ayush Dubey, Qiong Hu, Apurv Suman, Vaibhav Aggarwal, Ilya
  Kornakov, Fei Xia, Wing Lowe, Alexey Kolganov, Ted Xiao, Vitaly Nikolaev,
  Steven Hemingray, Bonnie Li, Joana Iljazi, Mikołaj Rybiński, Ballie Sandhu,
  Peggy Lu, Thang Luong, Rodolphe Jenatton, Vineetha Govindaraj, Hui, Li,
  Gabriel Dulac-Arnold, Wonpyo Park, Henry Wang, Abhinit Modi, Jean
  Pouget-Abadie, Kristina Greller, Rahul Gupta, Robert Berry, Prajit
  Ramachandran, Jinyu Xie, Liam McCafferty, Jianling Wang, Kilol Gupta,
  Hyeontaek Lim, Blaž Bratanič, Andy Brock, Ilia Akolzin, Jim Sproch, Dan
  Karliner, Duhyeon Kim, Adrian Goedeckemeyer, Noam Shazeer, Cordelia Schmid,
  Daniele Calandriello, Parul Bhatia, Krzysztof Choromanski, Ceslee Montgomery,
  Dheeru Dua, Ana Ramalho, Helen King, Yue Gao, Lynn Nguyen, David Lindner,
  Divya Pitta, Oleaser Johnson, Khalid Salama, Diego Ardila, Michael Han, Erin
  Farnese, Seth Odoom, Ziyue Wang, Xiangzhuo Ding, Norman Rink, Ray Smith,
  Harshal~Tushar Lehri, Eden Cohen, Neera Vats, Tong He, Parthasarathy
  Gopavarapu, Adam Paszke, Miteyan Patel, Wouter~Van Gansbeke, Lucia Loher,
  Luis Castro, Maria Voitovich, Tamara von Glehn, Nelson George, Simon Niklaus,
  Zach Eaton-Rosen, Nemanja Rakićević, Erik Jue, Sagi Perel, Carrie Zhang,
  Yuval Bahat, Angéline Pouget, Zhi Xing, Fantine Huot, Ashish Shenoy, Taylor
  Bos, Vincent Coriou, Bryan Richter, Natasha Noy, Yaqing Wang, Santiago
  Ontanon, Siyang Qin, Gleb Makarchuk, Demis Hassabis, Zhuowan Li, Mandar
  Sharma, Kumaran Venkatesan, Iurii Kemaev, Roxanne Daniel, Shiyu Huang, Saloni
  Shah, Octavio Ponce, Warren, Chen, Manaal Faruqui, Jialin Wu, Slavica
  Andačić, Szabolcs Payrits, Daniel McDuff, Tom Hume, Yuan Cao, MH~Tessler,
  Qingze Wang, Yinan Wang, Ivor Rendulic, Eirikur Agustsson, Matthew Johnson,
  Tanya Lando, Andrew Howard, Sri Gayatri~Sundara Padmanabhan, Mayank Daswani,
  Andrea Banino, Michael Kilgore, Jonathan Heek, Ziwei Ji, Alvaro Caceres,
  Conglong Li, Nora Kassner, Alexey Vlaskin, Zeyu Liu, Alex Grills, Yanhan Hou,
  Roykrong Sukkerd, Gowoon Cheon, Nishita Shetty, Larisa Markeeva, Piotr
  Stanczyk, Tejas Iyer, Yuan Gong, Shawn Gao, Keerthana Gopalakrishnan, Tim
  Blyth, Malcolm Reynolds, Avishkar Bhoopchand, Misha Bilenko, Dero Gharibian,
  Vicky Zayats, Aleksandra Faust, Abhinav Singh, Min Ma, Hongyang Jiao,
  Sudheendra Vijayanarasimhan, Lora Aroyo, Vikas Yadav, Sarah Chakera, Ashwin
  Kakarla, Vilobh Meshram, Karol Gregor, Gabriela Botea, Evan Senter, Dawei
  Jia, Geza Kovacs, Neha Sharma, Sebastien Baur, Kai Kang, Yifan He, Lin Zhuo,
  Marija Kostelac, Itay Laish, Songyou Peng, Louis O'Bryan, Daniel Kasenberg,
  Girish~Ramchandra Rao, Edouard Leurent, Biao Zhang, Sage Stevens, Ana
  Salazar, Ye~Zhang, Ivan Lobov, Jake Walker, Allen Porter, Morgan Redshaw, Han
  Ke, Abhishek Rao, Alex Lee, Hoi Lam, Michael Moffitt, Jaeyoun Kim, Siyuan
  Qiao, Terry Koo, Robert Dadashi, Xinying Song, Mukund Sundararajan, Peng Xu,
  Chizu Kawamoto, Yan Zhong, Clara Barbu, Apoorv Reddy, Mauro Verzetti, Leon
  Li, George Papamakarios, Hanna Klimczak-Plucińska, Mary Cassin, Koray
  Kavukcuoglu, Rigel Swavely, Alain Vaucher, Jeffrey Zhao, Ross Hemsley,
  Michael Tschannen, Heming Ge, Gaurav Menghani, Yang Yu, Natalie Ha, Wei He,
  Xiao Wu, Maggie Song, Rachel Sterneck, Stefan Zinke, Dan~A. Calian, Annie
  Marsden, Alejandro~Cruzado Ruiz, Matteo Hessel, Almog Gueta, Benjamin Lee,
  Brian Farris, Manish Gupta, Yunjie Li, Mohammad Saleh, Vedant Misra, Kefan
  Xiao, Piermaria Mendolicchio, Gavin Buttimore, Varvara Krayvanova, Nigamaa
  Nayakanti, Matthew Wiethoff, Yash Pande, Azalia Mirhoseini, Ni~Lao, Jasmine
  Liu, Yiqing Hua, Angie Chen, Yury Malkov, Dmitry Kalashnikov, Shubham Gupta,
  Kartik Audhkhasi, Yuexiang Zhai, Sudhindra Kopalle, Prateek Jain, Eran Ofek,
  Clemens Meyer, Khuslen Baatarsukh, Hana Strejček, Jun Qian, James Freedman,
  Ricardo Figueira, Michal Sokolik, Olivier Bachem, Raymond Lin, Dia Kharrat,
  Chris Hidey, Pingmei Xu, Dennis Duan, Yin Li, Muge Ersoy, Richard Everett,
  Kevin Cen, Rebeca Santamaria-Fernandez, Amir Taubenfeld, Ian Mackinnon, Linda
  Deng, Polina Zablotskaia, Shashank Viswanadha, Shivanker Goel, Damion Yates,
  Yunxiao Deng, Peter Choy, Mingqing Chen, Abhishek Sinha, Alex Mossin, Yiming
  Wang, Arthur Szlam, Susan Hao, Paul~Kishan Rubenstein, Metin Toksoz-Exley,
  Miranda Aperghis, Yin Zhong, Junwhan Ahn, Michael Isard, Olivier Lacombe,
  Florian Luisier, Chrysovalantis Anastasiou, Yogesh Kalley, Utsav Prabhu, Emma
  Dunleavy, Shaan Bijwadia, Justin Mao-Jones, Kelly Chen, Rama Pasumarthi,
  Emily Wood, Adil Dostmohamed, Nate Hurley, Jiri Simsa, Alicia Parrish, Mantas
  Pajarskas, Matt Harvey, Ondrej Skopek, Yony Kochinski, Javier Rey, Verena
  Rieser, Denny Zhou, Sun~Jae Lee, Trilok Acharya, Guowang Li, Joe Jiang,
  Xiaofan Zhang, Bryant Gipson, Ethan Mahintorabi, Marco Gelmi, Nima
  Khajehnouri, Angel Yeh, Kayi Lee, Loic Matthey, Leslie Baker, Trang Pham, Han
  Fu, Alex Pak, Prakhar Gupta, Cristina Vasconcelos, Adam Sadovsky, Brian
  Walker, Sissie Hsiao, Patrik Zochbauer, Andreea Marzoca, Noam Velan, Junhao
  Zeng, Gilles Baechler, Danny Driess, Divya Jain, Yanping Huang, Lizzie Tao,
  John Maggs, Nir Levine, Jon Schneider, Erika Gemzer, Samuel Petit, Shan Han,
  Zach Fisher, Dustin Zelle, Courtney Biles, Eugene Ie, Asya Fadeeva, Casper
  Liu, Juliana~Vicente Franco, Adrian Collister, Hao Zhang, Renshen Wang,
  Ruizhe Zhao, Leandro Kieliger, Kurt Shuster, Rui Zhu, Boqing Gong, Lawrence
  Chan, Ruoxi Sun, Sujoy Basu, Roland Zimmermann, Jamie Hayes, Abhishek Bapna,
  Jasper Snoek, Weel Yang, Puranjay Datta, Jad~Al Abdallah, Kevin Kilgour,
  Lu~Li, SQ~Mah, Yennie Jun, Morgane Rivière, Abhijit Karmarkar, Tammo
  Spalink, Tao Huang, Lucas Gonzalez, Duc-Hieu Tran, Averi Nowak, John
  Palowitch, Martin Chadwick, Ellie Talius, Harsh Mehta, Thibault Sellam,
  Philipp Fränken, Massimo Nicosia, Kyle He, Aditya Kini, David Amos, Sugato
  Basu, Harrison Jobe, Eleni Shaw, Qiantong Xu, Colin Evans, Daisuke Ikeda,
  Chaochao Yan, Larry Jin, Lun Wang, Sachin Yadav, Ilia Labzovsky, Ramesh
  Sampath, Ada Ma, Candice Schumann, Aditya Siddhant, Rohin Shah, John Youssef,
  Rishabh Agarwal, Natalie Dabney, Alessio Tonioni, Moran Ambar, Jing Li,
  Isabelle Guyon, Benny Li, David Soergel, Boya Fang, Georgi Karadzhov,
  Cristian Udrescu, Trieu Trinh, Vikas Raunak, Seb Noury, Dee Guo, Sonal Gupta,
  Mara Finkelstein, Denis Petek, Lihao Liang, Greg Billock, Pei Sun, David
  Wood, Yiwen Song, Xiaobin Yu, Tatiana Matejovicova, Regev Cohen, Kalyan
  Andra, David D'Ambrosio, Zhiwei Deng, Vincent Nallatamby, Ebrahim Songhori,
  Rumen Dangovski, Andrew Lampinen, Pankil Botadra, Adam Hillier, Jiawei Cao,
  Nagabhushan Baddi, Adhi Kuncoro, Toshihiro Yoshino, Ankit Bhagatwala,
  Marcáurelio Ranzato, Rylan Schaeffer, Tianlin Liu, Shuai Ye, Obaid Sarvana,
  John Nham, Chenkai Kuang, Isabel Gao, Jinoo Baek, Shubham Mittal, Ayzaan
  Wahid, Anita Gergely, Bin Ni, Josh Feldman, Carrie Muir, Pascal Lamblin,
  Wolfgang Macherey, Ethan Dyer, Logan Kilpatrick, Víctor Campos, Mukul
  Bhutani, Stanislav Fort, Yanif Ahmad, Aliaksei Severyn, Kleopatra
  Chatziprimou, Oleksandr Ferludin, Mason Dimarco, Aditya Kusupati, Joe
  Heyward, Dan Bahir, Kevin Villela, Katie Millican, Dror Marcus, Sanaz
  Bahargam, Caglar Unlu, Nicholas Roth, Zichuan Wei, Siddharth Gopal, Deepanway
  Ghoshal, Edward Lee, Sharon Lin, Jennie Lees, Dayeong Lee, Anahita Hosseini,
  Connie Fan, Seth Neel, Marcus Wu, Yasemin Altun, Honglong Cai, Enrique
  Piqueras, Josh Woodward, Alessandro Bissacco, Salem Haykal, Mahyar Bordbar,
  Prasha Sundaram, Sarah Hodkinson, Daniel Toyama, George Polovets, Austin
  Myers, Anu Sinha, Tomer Levinboim, Kashyap Krishnakumar, Rachita Chhaparia,
  Tatiana Sholokhova, Nitesh~Bharadwaj Gundavarapu, Ganesh Jawahar, Haroon
  Qureshi, Jieru Hu, Nikola Momchev, Matthew Rahtz, Renjie Wu, Aishwarya~P S,
  Kedar Dhamdhere, Meiqi Guo, Umang Gupta, Ali Eslami, Mariano Schain, Michiel
  Blokzijl, David Welling, Dave Orr, Levent Bolelli, Nicolas Perez-Nieves,
  Mikhail Sirotenko, Aman Prasad, Arjun Kar, Borja De~Balle Pigem, Tayfun
  Terzi, Gellért Weisz, Dipankar Ghosh, Aditi Mavalankar, Dhruv Madeka, Kaspar
  Daugaard, Hartwig Adam, Viraj Shah, Dana Berman, Maggie Tran, Steven Baker,
  Ewa Andrejczuk, Grishma Chole, Ganna Raboshchuk, Mahdi Mirzazadeh, Thais
  Kagohara, Shimu Wu, Christian Schallhart, Bernett Orlando, Chen Wang, Alban
  Rrustemi, Hao Xiong, Hao Liu, Arpi Vezer, Nolan Ramsden, Shuo yiin Chang,
  Sidharth Mudgal, Yan Li, Nino Vieillard, Yedid Hoshen, Farooq Ahmad, Ambrose
  Slone, Amy Hua, Natan Potikha, Mirko Rossini, Jon Stritar, Sushant Prakash,
  Zifeng Wang, Xuanyi Dong, Alireza Nazari, Efrat Nehoran, Kaan Tekelioglu,
  Yinxiao Li, Kartikeya Badola, Tom Funkhouser, Yuanzhen Li, Varun Yerram,
  Ramya Ganeshan, Daniel Formoso, Karol Langner, Tian Shi, Huijian Li, Yumeya
  Yamamori, Amayika Panda, Alaa Saade, Angelo~Scorza Scarpati, Chris Breaux,
  CJ~Carey, Zongwei Zhou, Cho-Jui Hsieh, Sophie Bridgers, Alena Butryna,
  Nishesh Gupta, Vaibhav Tulsyan, Sanghyun Woo, Evgenii Eltyshev, Will
  Grathwohl, Chanel Parks, Seth Benjamin, Rina Panigrahy, Shenil Dodhia,
  Daniel~De Freitas, Chris Sauer, Will Song, Ferran Alet, Jackson Tolins,
  Cosmin Paduraru, Xingyi Zhou, Brian Albert, Zizhao Zhang, Lei Shu, Mudit
  Bansal, Sarah Nguyen, Amir Globerson, Owen Xiao, James Manyika, Tom Hennigan,
  Rong Rong, Josip Matak, Anton Bakalov, Ankur Sharma, Danila Sinopalnikov,
  Andrew Pierson, Stephen Roller, Geoff Brown, Mingcen Gao, Toshiyuki Fukuzawa,
  Amin Ghafouri, Kenny Vassigh, Iain Barr, Zhicheng Wang, Anna Korsun, Rajesh
  Jayaram, Lijie Ren, Tim Zaman, Samira Khan, Yana Lunts, Dan Deutsch, Dave
  Uthus, Nitzan Katz, Masha Samsikova, Amr Khalifa, Nikhil Sethi, Jiao Sun,
  Luming Tang, Uri Alon, Xianghong Luo, Dian Yu, Abhishek Nayyar, Bryce
  Petrini, Will Truong, Vincent Hellendoorn, Nikolai Chinaev, Chris Alberti,
  Wei Wang, Jingcao Hu, Vahab Mirrokni, Ananth Balashankar, Avia Aharon, Aahil
  Mehta, Ahmet Iscen, Joseph Kready, Lucas Manning, Anhad Mohananey, Yuankai
  Chen, Anshuman Tripathi, Allen Wu, Igor Petrovski, Dawsen Hwang, Martin
  Baeuml, Shreyas Chandrakaladharan, Yuan Liu, Rey Coaguila, Maxwell Chen,
  Sally Ma, Pouya Tafti, Susheel Tatineni, Terry Spitz, Jiayu Ye, Paul Vicol,
  Mihaela Rosca, Adrià Puigdomènech, Zohar Yahav, Sanjay Ghemawat, Hanzhao
  Lin, Phoebe Kirk, Zaid Nabulsi, Sergey Brin, Bernd Bohnet, Ken Caluwaerts,
  Aditya~Srikanth Veerubhotla, Dan Zheng, Zihang Dai, Petre Petrov, Yichong Xu,
  Ramin Mehran, Zhuo Xu, Luisa Zintgraf, Jiho Choi, Spurthi~Amba Hombaiah,
  Romal Thoppilan, Sashank Reddi, Lukasz Lew, Li~Li, Kellie Webster,
  KP~Sawhney, Lampros Lamprou, Siamak Shakeri, Mayank Lunayach, Jianmin Chen,
  Sumit Bagri, Alex Salcianu, Ying Chen, Yani Donchev, Charlotte Magister,
  Signe Nørly, Vitor Rodrigues, Tomas Izo, Hila Noga, Joe Zou, Thomas Köppe,
  Wenxuan Zhou, Kenton Lee, Xiangzhu Long, Danielle Eisenbud, Anthony Chen,
  Connor Schenck, Chi~Ming To, Peilin Zhong, Emanuel Taropa, Minh Truong, Omer
  Levy, Danilo Martins, Zhiyuan Zhang, Christopher Semturs, Kelvin Zhang, Alex
  Yakubovich, Pol Moreno, Lara McConnaughey, Di~Lu, Sam Redmond, Lotte Weerts,
  Yonatan Bitton, Tiziana Refice, Nicolas Lacasse, Arthur Conmy, Corentin
  Tallec, Julian Odell, Hannah Forbes-Pollard, Arkadiusz Socala, Jonathan
  Hoech, Pushmeet Kohli, Alanna Walton, Rui Wang, Mikita Sazanovich, Kexin Zhu,
  Andrei Kapishnikov, Rich Galt, Matthew Denton, Ben Murdoch, Caitlin Sikora,
  Kareem Mohamed, Wei Wei, Uri First, Tim McConnell, Luis~C. Cobo, James Qin,
  Thi Avrahami, Daniel Balle, Yu~Watanabe, Annie Louis, Adam Kraft, Setareh
  Ariafar, Yiming Gu, Eugénie Rives, Charles Yoon, Andrei Rusu, James
  Cobon-Kerr, Chris Hahn, Jiaming Luo, Yuvein, Zhu, Niharika Ahuja, Rodrigo
  Benenson, Raphaël~Lopez Kaufman, Honglin Yu, Lloyd Hightower, Junlin Zhang,
  Darren Ni, Lisa~Anne Hendricks, Gabby Wang, Gal Yona, Lalit Jain, Pablo
  Barrio, Surya Bhupatiraju, Siva Velusamy, Allan Dafoe, Sebastian Riedel, Tara
  Thomas, Zhe Yuan, Mathias Bellaiche, Sheena Panthaplackel, Klemen Kloboves,
  Sarthak Jauhari, Canfer Akbulut, Todor Davchev, Evgeny Gladchenko, David
  Madras, Aleksandr Chuklin, Tyrone Hill, Quan Yuan, Mukundan Madhavan, Luke
  Leonhard, Dylan Scandinaro, Qihang Chen, Ning Niu, Arthur Douillard, Bogdan
  Damoc, Yasumasa Onoe, Fabian Pedregosa, Fred Bertsch, Chas Leichner, Joseph
  Pagadora, Jonathan Malmaud, Sameera Ponda, Andy Twigg, Oleksii Duzhyi,
  Jingwei Shen, Miaosen Wang, Roopal Garg, Jing Chen, Utku Evci, Jonathan Lee,
  Leon Liu, Koji Kojima, Masa Yamaguchi, Arunkumar Rajendran, AJ~Piergiovanni,
  Vinodh~Kumar Rajendran, Marco Fornoni, Gabriel Ibagon, Harry Ragan, Sadh~MNM
  Khan, John Blitzer, Andrew Bunner, Guan Sun, Takahiro Kosakai, Scott
  Lundberg, Ndidi Elue, Kelvin Guu, SK~Park, Jane Park, Arunachalam
  Narayanaswamy, Chengda Wu, Jayaram Mudigonda, Trevor Cohn, Hairong Mu, Ravi
  Kumar, Laura Graesser, Yichi Zhang, Richard Killam, Vincent Zhuang, Mai
  Giménez, Wael~Al Jishi, Ruy Ley-Wild, Alex Zhai, Kazuki Osawa, Diego
  Cedillo, Jialu Liu, Mayank Upadhyay, Marcin Sieniek, Roshan Sharma, Tom
  Paine, Anelia Angelova, Sravanti Addepalli, Carolina Parada, Kingshuk
  Majumder, Avery Lamp, Sanjiv Kumar, Xiang Deng, Artiom Myaskovsky, Tea
  Sabolić, Jeffrey Dudek, Sarah York, Félix de~Chaumont~Quitry, Jiazhong Nie,
  Dee Cattle, Alok Gunjan, Bilal Piot, Waleed Khawaja, Seojin Bang, Simon Wang,
  Siavash Khodadadeh, Raghavender R, Praynaa Rawlani, Richard Powell, Kevin
  Lee, Johannes Griesser, GS~Oh, Cesar Magalhaes, Yujia Li, Simon Tokumine,
  Hadas~Natalie Vogel, Dennis Hsu, Arturo BC, Disha Jindal, Matan Cohen,
  Zi~Yang, Junwei Yuan, Dario de~Cesare, Tony Bruguier, Jun Xu, Monica Roy,
  Alon Jacovi, Dan Belov, Rahul Arya, Phoenix Meadowlark, Shlomi Cohen-Ganor,
  Wenting Ye, Patrick Morris-Suzuki, Praseem Banzal, Gan Song, Pranavaraj
  Ponnuramu, Fred Zhang, George Scrivener, Salah Zaiem, Alif~Raditya Rochman,
  Kehang Han, Badih Ghazi, Kate Lee, Shahar Drath, Daniel Suo, Antonious
  Girgis, Pradeep Shenoy, Duy Nguyen, Douglas Eck, Somit Gupta, Le~Yan, Joao
  Carreira, Anmol Gulati, Ruoxin Sang, Daniil Mirylenka, Emma Cooney, Edward
  Chou, Mingyang Ling, Cindy Fan, Ben Coleman, Guilherme Tubone, Ravin Kumar,
  Jason Baldridge, Felix Hernandez-Campos, Angeliki Lazaridou, James Besley,
  Itay Yona, Neslihan Bulut, Quentin Wellens, AJ~Pierigiovanni, Jasmine George,
  Richard Green, Pu~Han, Connie Tao, Geoff Clark, Chong You, Abbas Abdolmaleki,
  Justin Fu, Tongzhou Chen, Ashwin Chaugule, Angad Chandorkar, Altaf Rahman,
  Will Thompson, Penporn Koanantakool, Mike Bernico, Jie Ren, Andrey Vlasov,
  Sergei Vassilvitskii, Maciej Kula, Yizhong Liang, Dahun Kim, Yangsibo Huang,
  Chengxi Ye, Dmitry Lepikhin, and Wesley Helmholz.
\newblock Gemini 2.5: Pushing the frontier with advanced reasoning,
  multimodality, long context, and next generation agentic capabilities, 2025.
\newblock URL \url{https://arxiv.org/abs/2507.06261}.

\bibitem[C{\^o}t{\'e} et~al.(2018)C{\^o}t{\'e}, K{\'a}d{\'a}r, Yuan,
  et~al.]{cote2018textworld}
Marc-Alexandre C{\^o}t{\'e}, {\'A}kos K{\'a}d{\'a}r, Xingdi Yuan, et~al.
\newblock {TextWorld}: A learning environment for text-based games.
\newblock In \emph{Computer Games Workshop at IJCAI}, 2018.

\bibitem[{DeepSeek-AI}(2026{\natexlab{a}})]{deepseek2026flash0731}
{DeepSeek-AI}.
\newblock {DeepSeek-V4-Flash-0731}.
\newblock Official model repository, July 2026{\natexlab{a}}.
\newblock URL \url{https://huggingface.co/deepseek-ai/DeepSeek-V4-Flash-0731}.

\bibitem[{DeepSeek-AI}(2026{\natexlab{b}})]{deepseek2026pro0813}
{DeepSeek-AI}.
\newblock {DeepSeek-V4-Pro-0813}.
\newblock Official model repository, August 2026{\natexlab{b}}.
\newblock URL \url{https://huggingface.co/deepseek-ai/DeepSeek-V4-Pro-0813}.

\bibitem[{DylCat}(2025)]{dylcat2025supertux}
{DylCat}.
\newblock {SuperTux} -- welcome to antarctica in 26.98.
\newblock Speedrun.com, 2025.
\newblock URL \url{https://www.speedrun.com/st/runs/yljjn7km}.
\newblock Verified run, category 0.6.x; real-time attempt (RTA).

\bibitem[{EiP25}(2026)]{eip25_2026_sml1}
{EiP25}.
\newblock Super mario land -- any\% in 12:21.183.
\newblock Speedrun.com, 2026.
\newblock URL \url{https://www.speedrun.com/sml1/runs/m7j0kgem}.
\newblock Verified run, 1st place, Super Game Boy 2 (JPN/NTSC).

\bibitem[Fan et~al.(2022)Fan, Wang, Jiang, Mandlekar, Yang, Zhu, Tang, Huang,
  Zhu, and Anandkumar]{fan2022minedojo}
Linxi Fan, Guanzhi Wang, Yunfan Jiang, Ajay Mandlekar, Yuncong Yang, Haoyi Zhu,
  Andrew Tang, De-An Huang, Yuke Zhu, and Anima Anandkumar.
\newblock {MineDojo}: Building open-ended embodied agents with internet-scale
  knowledge.
\newblock In \emph{Advances in Neural Information Processing Systems (NeurIPS)
  Datasets and Benchmarks Track}, 2022.

\bibitem[Francis(2016)]{francis2016speedrunners}
Bryant Francis.
\newblock 6 ways to make your game more appealing to speedrunners.
\newblock Game Developer, 2016.
\newblock URL
  \url{https://www.gamedeveloper.com/business/6-ways-to-make-your-game-more-appealing-to-speedrunners}.

\bibitem[{Google DeepMind}(2026)]{deepmind2026gemini37flash}
{Google DeepMind}.
\newblock Gemini 3.7 {Flash} {Model} {Card}.
\newblock
  \url{https://storage.googleapis.com/deepmind-media/Model-Cards/Gemini-3-7-Flash-Model-Card.pdf},
  August 2026.
\newblock Accessed: 2026-08-24.

\bibitem[Guertler et~al.(2025)Guertler, Cheng, Yu, Liu, Choshen, and
  Tan]{guertler2025textarena}
Leon Guertler, Bobby Cheng, Simon Yu, Bo~Liu, Leshem Choshen, and Cheston Tan.
\newblock {TextArena}.
\newblock \emph{arXiv preprint arXiv:2504.11442}, 2025.

\bibitem[Guruprasad et~al.(2026)Guruprasad, Rivera, Lu, Jain, Ren, and
  Sikka]{multinet_v2_preview_technical_report_2026}
Pranav Guruprasad, Sean Rivera, Helen Lu, Arushi Jain, Hangliang Ren, and
  Harshvardhan Sikka.
\newblock Multinet 2.0 preview: Goal progress decays with task horizon for
  frontier vlms in interactive 2d environments, 2026.
\newblock URL \url{https://www.fig.inc/blog/multinet-v2-preview/}.
\newblock MultiNet 2.0 Preview: Interactive 2D Mazes.

\bibitem[Hafner(2022)]{hafner2022crafter}
Danijar Hafner.
\newblock Benchmarking the spectrum of agent capabilities.
\newblock In \emph{International Conference on Learning Representations
  (ICLR)}, 2022.

\bibitem[Hafner et~al.(2023)Hafner, Pasukonis, Ba, and
  Lillicrap]{hafner2023dreamerv3}
Danijar Hafner, Jurgis Pasukonis, Jimmy Ba, and Timothy Lillicrap.
\newblock Mastering diverse domains through world models.
\newblock \emph{arXiv preprint arXiv:2301.04104}, 2023.

\bibitem[Hausknecht et~al.(2020)Hausknecht, Ammanabrolu, C{\^o}t{\'e}, and
  Yuan]{hausknecht2020jericho}
Matthew Hausknecht, Prithviraj Ammanabrolu, Marc-Alexandre C{\^o}t{\'e}, and
  Xingdi Yuan.
\newblock Interactive fiction games: A colossal adventure.
\newblock In \emph{Proceedings of the AAAI Conference on Artificial
  Intelligence}, 2020.

\bibitem[Hearn \& Demaine(2009)Hearn and Demaine]{hearn2009games}
Robert~A. Hearn and Erik~D. Demaine.
\newblock \emph{Games, Puzzles, and Computation}.
\newblock A K Peters, 2009.

\bibitem[Hu et~al.(2026)Hu, Huo, Zhang, Yu, Xing, Stoica, Rosing, Jin, and
  Zhang]{hu2026lmgamebench}
Lanxiang Hu, Mingjia Huo, Yuxuan Zhang, Haoyang Yu, Eric~P. Xing, Ion Stoica,
  Tajana Rosing, Haojian Jin, and Hao Zhang.
\newblock lmgame-bench: How good are {LLM}s at playing games?
\newblock In \emph{The Fourteenth International Conference on Learning
  Representations}, 2026.
\newblock URL \url{https://openreview.net/forum?id=qeziG97WUZ}.

\bibitem[Hu et~al.(2024)Hu, Huang, and
  Liu]{hu2024pokellmonhumanparityagentpokemon}
Sihao Hu, Tiansheng Huang, and Ling Liu.
\newblock Pokellmon: A human-parity agent for pokemon battles with large
  language models, 2024.
\newblock URL \url{https://arxiv.org/abs/2402.01118}.

\bibitem[Kamahori et~al.(2026)Kamahori, Li, Peter, and
  Kasikci]{kamahori2026vibeserveaiagentsbuild}
Keisuke Kamahori, Shihang Li, Simon Peter, and Baris Kasikci.
\newblock Vibeserve: Can ai agents build bespoke llm serving systems?, 2026.
\newblock URL \url{https://arxiv.org/abs/2605.06068}.

\bibitem[Karpathy(2026)]{karpathy2026autoresearch}
Andrej Karpathy.
\newblock {autoresearch}: An autonomous {LLM} research loop.
\newblock \url{https://github.com/karpathy/autoresearch}, 2026.
\newblock Accessed: 2026-05-06.

\bibitem[Karten et~al.(2026{\natexlab{a}})Karten, Grigsby, Jr, Bae, Hong,
  Jeong, Jung, Kerdthaisong, Kim, Kim, Kim, Kwon, Liu, Mariglia, Park, Schink,
  Shi, Sistilli, Twin, Urdu, Urdu, Wang, Wu, Zhang, Zhou, Milani, Vodrahalli,
  Zhang, Fang, Zhu, and
  Jin]{karten2026pokeagentchallengecompetitivelongcontext}
Seth Karten, Jake Grigsby, Tersoo~Upaa Jr, Junik Bae, Seonghun Hong, Hyunyoung
  Jeong, Jaeyoon Jung, Kun Kerdthaisong, Gyungbo Kim, Hyeokgi Kim, Yujin Kim,
  Eunju Kwon, Dongyu Liu, Patrick Mariglia, Sangyeon Park, Benedikt Schink,
  Xianwei Shi, Anthony Sistilli, Joseph Twin, Arian Urdu, Matin Urdu, Qiao
  Wang, Ling Wu, Wenli Zhang, Kunsheng Zhou, Stephanie Milani, Kiran
  Vodrahalli, Amy Zhang, Fei Fang, Yuke Zhu, and Chi Jin.
\newblock The pokeagent challenge: Competitive and long-context learning at
  scale, 2026{\natexlab{a}}.
\newblock URL \url{https://arxiv.org/abs/2603.15563}.

\bibitem[Karten et~al.(2026{\natexlab{b}})Karten, Zhang, Jr, Feng, Li, Shi,
  Jin, and Vodrahalli]{karten2026continualharnessonlineadaptation}
Seth Karten, Joel Zhang, Tersoo~Upaa Jr, Ruirong Feng, Wenzhe Li, Chengshuai
  Shi, Chi Jin, and Kiran Vodrahalli.
\newblock Continual harness: Online adaptation for self-improving foundation
  agents, 2026{\natexlab{b}}.
\newblock URL \url{https://arxiv.org/abs/2605.09998}.

\bibitem[Krentel(1988)]{krentel1988optimization}
Mark~W. Krentel.
\newblock The complexity of optimization problems.
\newblock \emph{Journal of Computer and System Sciences}, 36\penalty0
  (3):\penalty0 490--509, 1988.
\newblock \doi{10.1016/0022-0000(88)90039-6}.

\bibitem[Kurach et~al.(2020)Kurach, Raichuk, Sta{\'n}czyk,
  et~al.]{kurach2020football}
Karol Kurach, Anton Raichuk, Piotr Sta{\'n}czyk, et~al.
\newblock Google research football: A novel reinforcement learning environment.
\newblock In \emph{Proceedings of the AAAI Conference on Artificial
  Intelligence}, 2020.

\bibitem[K{\"u}ttler et~al.(2020)K{\"u}ttler, Nardelli, Miller,
  et~al.]{kuttler2020nethack}
Heinrich K{\"u}ttler, Nantas Nardelli, Alexander~H. Miller, et~al.
\newblock The {NetHack} learning environment.
\newblock In \emph{Advances in Neural Information Processing Systems
  (NeurIPS)}, 2020.

\bibitem[Machado et~al.(2018)Machado, Bellemare, Talvitie, Veness, Hausknecht,
  and Bowling]{machado2018ale}
Marlos~C. Machado, Marc~G. Bellemare, Erik Talvitie, Joel Veness, Matthew
  Hausknecht, and Michael Bowling.
\newblock Revisiting the arcade learning environment: Evaluation protocols and
  open problems for general agents.
\newblock \emph{Journal of Artificial Intelligence Research}, 61:\penalty0
  523--562, 2018.

\bibitem[Mandhane et~al.(2022)Mandhane, Zhernov, Rauh,
  et~al.]{mandhane2022muzero}
Amol Mandhane, Anton Zhernov, Maribeth Rauh, et~al.
\newblock {MuZero} with self-competition for rate control in {VP9} video
  compression.
\newblock \emph{arXiv preprint arXiv:2202.06626}, 2022.

\bibitem[Mnih et~al.(2015)Mnih, Kavukcuoglu, Silver, et~al.]{mnih2015dqn}
Volodymyr Mnih, Koray Kavukcuoglu, David Silver, et~al.
\newblock Human-level control through deep reinforcement learning.
\newblock \emph{Nature}, 518:\penalty0 529--533, 2015.
\newblock \doi{10.1038/nature14236}.

\bibitem[{OpenAI}(2026)]{openai2026gpt56}
{OpenAI}.
\newblock Gpt-5.6 system card, 2026.
\newblock URL \url{https://deploymentsafety.openai.com/gpt-5-6}.

\bibitem[{OpenCode contributors}(2026)]{opencode_software}
{OpenCode contributors}.
\newblock Opencode: The open source coding agent.
\newblock GitHub repository, 2026.
\newblock URL \url{https://github.com/anomalyco/opencode}.
\newblock Accessed: 2026-08-25.

\bibitem[Ouyang et~al.(2026)Ouyang, Hu, Lin, Ng, and
  Shou]{ouyang2026gameworldstandardizedverifiableevaluation}
Mingyu Ouyang, Siyuan Hu, Kevin~Qinghong Lin, Hwee~Tou Ng, and Mike~Zheng Shou.
\newblock Gameworld: Towards standardized and verifiable evaluation of
  multimodal game agents, 2026.
\newblock URL \url{https://arxiv.org/abs/2604.07429}.

\bibitem[Paglieri et~al.(2025)Paglieri, Cupia{\l}, Coward,
  et~al.]{paglieri2025balrog}
Davide Paglieri, Bart{\l}omiej Cupia{\l}, Samuel Coward, et~al.
\newblock {BALROG}: Benchmarking agentic {LLM} and {VLM} reasoning on games.
\newblock In \emph{International Conference on Learning Representations
  (ICLR)}, 2025.

\bibitem[Reed et~al.(2022)Reed, Zolna, Parisotto, et~al.]{reed2022gato}
Scott Reed, Konrad Zolna, Emilio Parisotto, et~al.
\newblock A generalist agent.
\newblock \emph{Transactions on Machine Learning Research}, 2022.

\bibitem[Samuel(1959)]{samuel1959checkers}
Arthur~L. Samuel.
\newblock Some studies in machine learning using the game of checkers.
\newblock \emph{IBM Journal of Research and Development}, 3\penalty0
  (3):\penalty0 210--229, 1959.
\newblock \doi{10.1147/rd.33.0210}.

\bibitem[Schrittwieser et~al.(2020)Schrittwieser, Antonoglou, Hubert,
  et~al.]{schrittwieser2020muzero}
Julian Schrittwieser, Ioannis Antonoglou, Thomas Hubert, et~al.
\newblock Mastering {Atari}, {Go}, chess and shogi by planning with a learned
  model.
\newblock \emph{Nature}, 588:\penalty0 604--609, 2020.
\newblock \doi{10.1038/s41586-020-03051-4}.

\bibitem[Scully-Blaker(2014)]{scullyblaker2014practiced}
Rainforest Scully-Blaker.
\newblock A practiced practice: Speedrunning through space with de {Certeau}
  and {Virilio}.
\newblock \emph{Game Studies}, 14\penalty0 (1), 2014.
\newblock URL \url{http://gamestudies.org/1401/articles/scullyblaker}.

\bibitem[Scully-Blaker(2018)]{scullyblaker2018museum}
Rainforest Scully-Blaker.
\newblock The speedrunning museum of accidents.
\newblock \emph{Kinephanos}, pp.\  71--88, 2018.
\newblock URL
  \url{https://www.kinephanos.ca/Revue_files/2018_Scully_Blaker.pdf}.
\newblock Special issue: Preserving Play.

\bibitem[Silver et~al.(2016)Silver, Huang, Maddison, Guez, Sifre, van~den
  Driessche, Schrittwieser, Antonoglou, Panneershelvam, Lanctot, Dieleman,
  Grewe, Nham, Kalchbrenner, Sutskever, Lillicrap, Leach, Kavukcuoglu, Graepel,
  and Hassabis]{silver2016alphago}
David Silver, Aja Huang, Chris~J. Maddison, Arthur Guez, Laurent Sifre, George
  van~den Driessche, Julian Schrittwieser, Ioannis Antonoglou, Veda
  Panneershelvam, Marc Lanctot, Sander Dieleman, Dominik Grewe, John Nham, Nal
  Kalchbrenner, Ilya Sutskever, Timothy Lillicrap, Madeleine Leach, Koray
  Kavukcuoglu, Thore Graepel, and Demis Hassabis.
\newblock Mastering the game of {Go} with deep neural networks and tree search.
\newblock \emph{Nature}, 529\penalty0 (7587):\penalty0 484--489, 2016.

\bibitem[Silver et~al.(2017)Silver, Schrittwieser, Simonyan,
  et~al.]{silver2017alphagozero}
David Silver, Julian Schrittwieser, Karen Simonyan, et~al.
\newblock Mastering the game of {Go} without human knowledge.
\newblock \emph{Nature}, 550:\penalty0 354--359, 2017.
\newblock \doi{10.1038/nature24270}.

\bibitem[Silver et~al.(2018)Silver, Hubert, Schrittwieser,
  et~al.]{silver2018alphazero}
David Silver, Thomas Hubert, Julian Schrittwieser, et~al.
\newblock A general reinforcement learning algorithm that masters chess, shogi,
  and {Go} through self-play.
\newblock \emph{Science}, 362\penalty0 (6419):\penalty0 1140--1144, 2018.
\newblock \doi{10.1126/science.aar6404}.

\bibitem[{SlNNED}(2025)]{sinned2025pkmnredblue}
{SlNNED}.
\newblock Pok\'emon red/blue -- any\% glitchless in 1:43:40.
\newblock Speedrun.com, 2025.
\newblock URL \url{https://www.speedrun.com/pkmnredblue/runs/y4n7endm}.
\newblock Verified run, 1st place, GBA emulator.

\bibitem[{SpaceXAI}(2026)]{spacexai2026grok46}
{SpaceXAI}.
\newblock Introducing {Grok} 4.6, 2026.
\newblock URL \url{https://x.ai/news/grok-4-6}.

\bibitem[Taesiri et~al.(2024)Taesiri, Feng, Nguyen, and
  Bezemer]{taesiri2024glitchbenchlargemultimodalmodels}
Mohammad~Reza Taesiri, Tianjun Feng, Anh Nguyen, and Cor-Paul Bezemer.
\newblock Glitchbench: Can large multimodal models detect video game glitches?,
  2024.
\newblock URL \url{https://arxiv.org/abs/2312.05291}.

\bibitem[Team et~al.(2026)Team, Bai, Bai, Bao, C., Cai, Cai, Cao, Cao, Chai,
  Charles, Che, Chen, Chen, Chen, Chen, Chen, Chen, Chen, Chen, Chen, Chen,
  Chen, Chen, Chen, Chen, Chen, Chen, Chen, Chen, Chen, Chen, Cheng, Cheng,
  Cui, Cui, Dai, Deng, Ding, Ding, Ding, Dong, Dong, Dong, Dong, Du, Du, Du,
  Du, Du, Fan, Feng, Feng, Feng, Fu, Fu, Gao, Gao, Gao, Gao, Gao, Geng, Gong,
  Gong, Gong, Gong, Gu, Gu, Guan, Guo, Guo, Guo, Guo, Hao, Hao, Hao, He, He,
  He, He, He, He, He, He, He, Hong, Hong, Hu, Hu, Hu, Hu, Hu, Hu, Hua, Huang,
  Huang, Huang, Huang, Huang, Huang, Huang, Huang, Hui, Jia, Jiang, Jiang,
  Jiang, Jin, Jin, Jing, Kong, Lai, Li, Li, Li, Li, Li, Li, Li, Li, Li, Li, Li,
  Li, Li, Li, Li, Li, Li, Li, Li, Li, Li, Li, Li, Li, Lin, Lin, Lin, Lin, Lin,
  Liu, Liu, Liu, Liu, Liu, Liu, Liu, Liu, Liu, Liu, Liu, Liu, Liu, Liu, Liu,
  Lu, Lu, Lu, Lu, Lu, Luo, Luo, Luo, Luo, Lyu, Lyu, Mao, Mei, Men, Ni, Niu,
  Pan, Peng, Qi, Qin, Qin, Qin, Qiu, Qiu, Qiu, Qu, Qu, Shang, Shao, Shen, Shi,
  Shi, Shi, Shi, Siu, Song, Song, Su, Su, Su, Sui, Sun, Sun, Sun, Sun, Sun,
  Sun, Tai, Tang, Tang, Tang, Tang, Tian, Tian, Tian, Tu, Wang, Wang, Wang,
  Wang, Wang, Wang, Wang, Wang, Wang, Wang, Wang, Wang, Wang, Wang, Wang, Wang,
  Wang, Wang, Wang, Wang, Wang, Wang, Wang, Wang, Wang, Wang, Wang, Wang, Wang,
  Wang, Wang, Wang, Wang, Wang, Wang, Wang, Wang, Wang, Wang, Wang, Wang, Wei,
  Wei, Wei, Wen, Wu, Wu, Wu, Wu, Wu, Wu, Wu, Wu, Wu, Xian, Xiang, Xiang, Xiao,
  Xiao, Xiao, Xie, Xie, Xie, Xie, Xing, Xiong, Xu, Xu, Xu, Xu, Xu, Xu, Xu, Xu,
  Xu, Xu, Xu, Xu, Xu, Xu, Xu, Xu, Xu, Xu, Xue, Yan, Yan, Yang, Yang, Yang,
  Yang, Yang, Yang, Yang, Yang, Yang, Yang, Yang, Yang, Yang, Yang, Yang, Yang,
  Yao, Ye, Ye, Ye, Ye, Yin, Yin, Yin, Yu, Yu, Yu, Yu, Yu, Yu, Yuan, Yuan, Yue,
  Yue, Yue, Zha, Zhan, Zhang, Zhang, Zhang, Zhang, Zhang, Zhang, Zhang, Zhang,
  Zhang, Zhang, Zhang, Zhang, Zhang, Zhang, Zhang, Zhang, Zhang, Zhang, Zhang,
  Zhang, Zhang, Zhang, Zhang, Zhang, Zhang, Zhang, Zhang, Zhang, Zhang, Zhang,
  Zhang, Zhao, Zhao, Zhao, Zhao, Zhao, Zhao, Zhao, Zhao, Zhao, Zhao, Zhao,
  Zheng, Zheng, Zheng, Zheng, Zheng, Zhong, Zhong, Zhong, Zhou, Zhou, Zhou,
  Zhou, Zhou, Zhou, Zhou, Zhu, Zhu, Zhu, Zhu, Zhu, Zhu, Zhuang, Zhuang, and
  Zu]{kimiteam2026kimik3openfrontier}
Kimi Team, Tongtong Bai, Yifan Bai, Yiping Bao, M.~C., Jianfeng Cai, Xinyuan
  Cai, Peizhou Cao, Yuxuan Cao, Ziwei Chai, Y.~Charles, H.~S. Che, Guanduo
  Chen, Guangyu Chen, Guanzheng Chen, Huarong Chen, Jia Chen, Jianlong Chen,
  Jun Chen, Kexin Chen, Peng Chen, Ruijue Chen, Wentao Chen, Xin Chen, Yang
  Chen, Yanru Chen, Yifei Chen, Yingjiang Chen, Yuankun Chen, Yujie Chen,
  Yutian Chen, Zhirong Chen, Dazhi Cheng, Yean Cheng, Jialei Cui, Jingbing Cui,
  Anqi Dai, Jiaqi Deng, Hao Ding, Rui Ding, Shaofeng Ding, Mengfan Dong,
  Mengnan Dong, Yuhao Dong, Yuxin Dong, Angang Du, Chenzhuang Du, Dikang Du,
  Jusen Du, Yulun Du, Yu~Fan, Jing Feng, Qiulin Feng, Yichen Feng, Kelin Fu,
  Qiang Fu, Fuxuan Gao, Hongcheng Gao, Jingyue Gao, Tong Gao, Weijia Gao,
  Shangyi Geng, Jie Gong, Linhu Gong, Shengao Gong, Xiaochen Gong, Qizheng Gu,
  Yicheng Gu, Shuhao Guan, Haiqing Guo, Shiqi Guo, Xiang Guo, Zhengyan Guo,
  Beixi Hao, Wenxin Hao, Xiaoru Hao, Dailan He, Haotian He, Lehan He, Qi~He,
  Weiran He, Xinran He, Xinyi He, Yibo He, Yunjia He, Chao Hong, Tiange Hong,
  Hao Hu, Jiaxi Hu, Ruikun Hu, Weiming Hu, Yangyang Hu, Zhenxing Hu, Liang Hua,
  Jinbin Huang, Ke~Huang, Ruiyuan Huang, Siying Huang, Weixiao Huang, Yan
  Huang, Zhengjie Huang, Zhiqi Huang, Yulong Hui, Chaobo Jia, Yutong Jiang,
  Zhejun Jiang, Zuoyou Jiang, Wenyi Jin, Xinyi Jin, Yu~Jing, Huanjun Kong,
  Guokun Lai, Aidi Li, Cheng Li, Chengyuan Li, Cong Li, Fang Li, Guanyu Li,
  Haoyang Li, Jia Li, Junxiong Li, Lei Li, Letian Li, Lincan Li, Weihong Li,
  Wentao Li, Xintong Li, Yang Li, Yishen Li, Yiwei Li, Yuxiao Li, Zhaowei Li,
  Zhaoxi Li, Zheming Li, Zhengxiao Li, Zhiyuan Li, Jiawei Lin, Xiaohan Lin,
  Yibo Lin, Zichao Lin, Ziyan Lin, Bill Liu, Boxiao Liu, Chuan Liu, Liang Liu,
  Shaowei Liu, Shudong Liu, Shuran Liu, Tianwei Liu, Weizhou Liu, Yangyang Liu,
  Yanming Liu, Yibo Liu, Yipeng Liu, Zhengying Liu, Zhiheng Liu, Enzhe Lu,
  Haoyu Lu, Linqiang Lu, Tingzhan Lu, Zhiyuan Lu, Aotian Luo, G.~Luo, Junyu
  Luo, Yifan Luo, B.~Lyu, Wenzhou Lyu, Shaoguang Mao, Yuan Mei, Xin Men,
  Minqing Ni, Yixuan Niu, Siyuan Pan, Shujun Peng, Zhangyang Qi, Ruoyu Qin,
  ZeChao Qin, Zeyu Qin, Haiquan Qiu, Jianxin Qiu, Jiezhong Qiu, Bowen Qu, Yuhao
  Qu, Zeyu Shang, Youbo Shao, Han Shen, Jincheng Shi, Juanfeng Shi, Lidong Shi,
  Shengyuan Shi, Wingchun Siu, Pengwei Song, Xiaoxi Song, Jianlin Su, Yunfeng
  Su, Zhaochen Su, Lin Sui, Jingsong Sun, Junyao Sun, Shaoning Sun, Shuzhe Sun,
  Tongyu Sun, Yujun Sun, Yunpeng Tai, Chuning Tang, Heyi Tang, Sirui Tang,
  Zecheng Tang, Chaoran Tian, Rongpeng Tian, Yu~Tian, Wei Tu, Chensi Wang,
  Chuang Wang, Chunjie Wang, Dinglu Wang, Feng Wang, Hailong Wang, Haiming
  Wang, Hao Wang, Hao Wang, Huaqing Wang, Hui Wang, Jiayi Wang, Jinglong Wang,
  Jinhong Wang, Jiuzheng Wang, Linian Wang, Shaobo Wang, Shenzhi Wang, Shuyi
  Wang, Si~Wang, Siyuan Wang, Tianfu Wang, Wenjue Wang, Xingran Wang, Xinmei
  Wang, Xinyuan Wang, Xusheng Wang, Yalin Wang, Yangkun Wang, Yao Wang, Yaoyu
  Wang, Yejie Wang, Yiqin Wang, Yucheng Wang, Yuzhi Wang, Zhaoji Wang, Zhaowei
  Wang, Zhengtao Wang, Zhenhao Wang, Zhongsheng Wang, Zifan Wang, Chu Wei, Ming
  Wei, Shouxin Wei, Zichen Wen, Fan Wu, Haoning Wu, Rucong Wu, Wenhao Wu,
  Xiaoxue Wu, Yingcong Wu, Yongqi Wu, Yuxin Wu, Zijian Wu, Xinglang Xian,
  Chenxuan Xiang, Yuye Xiang, Bocheng Xiao, Chenjun Xiao, Xin Xiao, Jin Xie,
  Xiaotong Xie, Yifeng Xie, Zhe Xie, Bowei Xing, Yiming Xiong, Baosheng Xu,
  Boyu Xu, Jiale Xu, Jianfan Xu, Jing Xu, Jinjing Xu, L.~H. Xu, Qingtao Xu,
  Shuyao Xu, Suting Xu, Tiantian Xu, Tianxiang Xu, Weixin Xu, Xinran Xu,
  Yangchuan Xu, Ye~Xu, Yueni Xu, Ziyao Xu, Haonan Xue, Junjie Yan, Yaoyao Yan,
  Fan Yang, Guangyao Yang, Hao Yang, Junwei Yang, Ruoyu Yang, Wenjie Yang,
  Xiaofei Yang, Xinyu Yang, Yi~Yang, Yiling Yang, Ying Yang, Yuchen Yang, Zhen
  Yang, Zhilin Yang, Zian Yang, Zuhao Yang, Haotian Yao, Dan Ye, Haoran Ye,
  Wenjie Ye, Zhanbo Ye, Bohong Yin, Haoxiang Yin, Xietong Yin, Chengzhen Yu,
  Haozhen Yu, Longhui Yu, Shengnan Yu, Shuying Yu, Tianxiang Yu, Enming Yuan,
  Mengjie Yuan, Tongtian Yue, Wei Yue, Yang Yue, Dunyuan Zha, Haobing Zhan,
  B.~H. Zhang, Dehao Zhang, Fei Zhang, Hao Zhang, Haoyuan Zhang, Huanyu Zhang,
  Jiapei Zhang, Jiaxuan Zhang, Jin Zhang, Kaiyi Zhang, Miaozhen Zhang, Puqi
  Zhang, Qinglei Zhang, Rong Zhang, Rui Zhang, Shaoshuai Zhang, Shiyi Zhang,
  Xiaobin Zhang, Xiaoyun Zhang, Y.~Zhang, Yangkun Zhang, Ye~Zhang, Yichi Zhang,
  Yikun Zhang, Yizhi Zhang, Yongting Zhang, Yu~Zhang, Yutao Zhang, Yutong
  Zhang, Zheng Zhang, Zijing Zhang, Bin Zhao, Chenguang Zhao, Feifan Zhao,
  Jinglun Zhao, Jinxiang Zhao, Shuai Zhao, Wenshuo Zhao, Xiangyu Zhao, Xuanle
  Zhao, Yikai Zhao, Zijia Zhao, Haozhi Zheng, Huabin Zheng, Ruihan Zheng,
  Shaojie Zheng, Tengyang Zheng, Haofeng Zhong, Lei Zhong, Longguang Zhong,
  M.~Zhou, Qiankang Zhou, Runjie Zhou, Ruozhang Zhou, Xinyu Zhou, Yiqiao Zhou,
  Zaida Zhou, Jinguo Zhu, Liya Zhu, Xinhao Zhu, Yangjunfeng Zhu, Yuxuan Zhu,
  Zhen Zhu, Chen Zhuang, Weiyu Zhuang, and Xinxing Zu.
\newblock Kimi {K}3: {O}pen frontier intelligence, 2026.
\newblock URL \url{https://arxiv.org/abs/2607.24653}.

\bibitem[Tesauro(1995)]{tesauro1995tdgammon}
Gerald Tesauro.
\newblock Temporal difference learning and {TD-Gammon}.
\newblock \emph{Communications of the ACM}, 38\penalty0 (3):\penalty0 58--68,
  1995.
\newblock \doi{10.1145/203330.203343}.

\bibitem[{The SuperTux Team}(2021)]{supertux2021}
{The SuperTux Team}.
\newblock {SuperTux} 0.6.3, December 2021.
\newblock URL \url{https://github.com/SuperTux/supertux/releases/tag/v0.6.3}.
\newblock Open-source 2D sidescrolling platform game.

\bibitem[{theodorepringle}(2021)]{pringle2021supertuxkart}
{theodorepringle}.
\newblock {SuperTuxKart} -- black forest in 2:09.220.
\newblock Speedrun.com, 2021.
\newblock URL \url{https://www.speedrun.com/stk/runs/y8dr38dz}.
\newblock Verified run, category 1.x; real-time attempt (RTA).

\bibitem[{Thinking Machines Lab}(2026)]{thinkingmachines2026inklingsmall}
{Thinking Machines Lab}.
\newblock Inkling-small model card, 2026.
\newblock URL \url{https://thinkingmachines.ai/model-card/inkling-small/}.

\bibitem[Toromanoff et~al.(2019)Toromanoff, Wirbel, and
  Moutarde]{toromanoff2019superhuman}
Marin Toromanoff, Emilie Wirbel, and Fabien Moutarde.
\newblock Is deep reinforcement learning really superhuman on {Atari}?
  {Leveling} the playing field.
\newblock \emph{arXiv preprint arXiv:1908.04683}, 2019.

\bibitem[{Tuxemon Contributors}(2026)]{tuxemon}
{Tuxemon Contributors}.
\newblock {Tuxemon}: Open source monster-fighting rpg.
\newblock GitHub repository, 2026.
\newblock URL \url{https://github.com/Tuxemon/Tuxemon}.

\bibitem[Vinyals et~al.(2017)Vinyals, Ewalds, Bartunov,
  et~al.]{vinyals2017sc2le}
Oriol Vinyals, Timo Ewalds, Sergey Bartunov, et~al.
\newblock {StarCraft II}: A new challenge for reinforcement learning.
\newblock \emph{arXiv preprint arXiv:1708.04782}, 2017.

\bibitem[Vinyals et~al.(2019)Vinyals, Babuschkin, Czarnecki, Mathieu, Dudzik,
  Chung, Choi, Powell, Ewalds, Georgiev, Oh, Horgan, Kroiss, Danihelka, Huang,
  Sifre, Cai, Agapiou, Jaderberg, Vezhnevets, Leblond, Pohlen, Dalibard,
  Budden, Sulsky, Molloy, Paine, Gulcehre, Wang, Pfaff, Wu, Ring, Yogatama,
  W{\"u}nsch, McKinney, Smith, Schaul, Lillicrap, Kavukcuoglu, Hassabis, Apps,
  and Silver]{vinyals2019alphastar}
Oriol Vinyals, Igor Babuschkin, Wojciech~M. Czarnecki, Micha{\"e}l Mathieu,
  Andrew Dudzik, Junyoung Chung, David~H. Choi, Richard Powell, Timo Ewalds,
  Petko Georgiev, Junhyuk Oh, Dan Horgan, Manuel Kroiss, Ivo Danihelka, Aja
  Huang, Laurent Sifre, Trevor Cai, John~P. Agapiou, Max Jaderberg,
  Alexander~S. Vezhnevets, R{\'e}mi Leblond, Tobias Pohlen, Valentin Dalibard,
  David Budden, Yury Sulsky, James Molloy, Tom~L. Paine, Caglar Gulcehre, Ziyu
  Wang, Tobias Pfaff, Yuhuai Wu, Roman Ring, Dani Yogatama, Dario W{\"u}nsch,
  Katrina McKinney, Oliver Smith, Tom Schaul, Timothy Lillicrap, Koray
  Kavukcuoglu, Demis Hassabis, Chris Apps, and David Silver.
\newblock Grandmaster level in {StarCraft II} using multi-agent reinforcement
  learning.
\newblock \emph{Nature}, 575\penalty0 (7782):\penalty0 350--354, 2019.

\bibitem[Wang et~al.(2024)Wang, Xie, Jiang, Mandlekar, Xiao, Zhu, Fan, and
  Anandkumar]{wang2024voyager}
Guanzhi Wang, Yuqi Xie, Yunfan Jiang, Ajay Mandlekar, Chaowei Xiao, Yuke Zhu,
  Linxi Fan, and Anima Anandkumar.
\newblock Voyager: An open-ended embodied agent with large language models.
\newblock \emph{Transactions on Machine Learning Research}, 2024.

\bibitem[Wu et~al.(2024)Wu, Tang, Mitchell, and Li]{wu2024smartplay}
Yue Wu, Xuan Tang, Tom Mitchell, and Yuanzhi Li.
\newblock {SmartPlay}: A benchmark for {LLMs} as intelligent agents.
\newblock In \emph{International Conference on Learning Representations
  (ICLR)}, 2024.

\bibitem[{Z.ai}(2026{\natexlab{a}})]{zai2026glm52}
{Z.ai}.
\newblock {GLM-5.2}: Built for long-horizon tasks, June 2026{\natexlab{a}}.
\newblock URL \url{https://z.ai/blog/glm-5.2}.

\bibitem[{Z.ai}(2026{\natexlab{b}})]{zai2026glm53}
{Z.ai}.
\newblock {GLM-5.3}: Frontier coding with emergent cyber capabilities, August
  2026{\natexlab{b}}.
\newblock URL \url{https://z.ai/blog/glm-5.3}.

\bibitem[Zhang et~al.(2025)Zhang, Griffiths, Narasimhan, and
  Press]{zhang2025videogamebench}
Alex~L. Zhang, Thomas~L. Griffiths, Karthik~R. Narasimhan, and Ofir Press.
\newblock {VideoGameBench}: Can vision-language models complete popular video
  games?
\newblock \emph{arXiv preprint arXiv:2505.18134}, 2025.

\bibitem[Zhang(2025)]{zhang2025gemini}
Joel Zhang.
\newblock The making of {G}emini {P}lays {P}ok\'emon.
\newblock Blog post, Joel's Developer Blog, August 2025.
\newblock URL \url{https://blog.jcz.dev/the-making-of-gemini-plays-pokemon}.
\newblock Accessed: 2026-08-24.

\bibitem[Zhao et~al.(2025)Zhao, Magka, Jiang, Li, Raileanu, Shavrina,
  Gagnon-Audet, Niu, Sodhani, Shvartsman, Lupu, Lupidi, Hambardzumyan,
  Josifoski, Toledo, Foster, Kun, Dunfield, Charnalia, Miller, Aodha, Foerster,
  and Bachrach]{zhao2025automated}
Bingchen Zhao, Despoina Magka, Minqi Jiang, Xian Li, Roberta Raileanu, Tatiana
  Shavrina, Jean-Christophe Gagnon-Audet, Kelvin Niu, Shagun Sodhani, Michael
  Shvartsman, Andrei Lupu, Alisia Lupidi, Karen Hambardzumyan, Martin
  Josifoski, Edan Toledo, Thomas Foster, Lucia~Cipolina Kun, Derek Dunfield,
  Abhishek Charnalia, Alexander Miller, Oisin~Mac Aodha, Jakob Foerster, and
  Yoram Bachrach.
\newblock The automated {LLM} speedrunning benchmark: Reproducing nano{GPT}
  improvements.
\newblock In \emph{Advances in Neural Information Processing Systems 38,
  Datasets and Benchmarks Track}, 2025.
\newblock \doi{10.52202/085713-0928}.
\newblock URL
  \url{https://proceedings.nips.cc/paper_files/paper/2025/hash/27e7f21d16fb840f3720ac87ad896220-Abstract-Datasets_and_Benchmarks_Track.html}.

\bibitem[Zhao et~al.(2023)Zhao, Kumar, Levine, and Finn]{Zhao-RSS-23}
Tony~Z. Zhao, Vikash Kumar, Sergey Levine, and Chelsea Finn.
\newblock {Learning Fine-Grained Bimanual Manipulation with Low-Cost Hardware}.
\newblock In \emph{Proceedings of Robotics: Science and Systems}, Daegu,
  Republic of Korea, July 2023.
\newblock \doi{10.15607/RSS.2023.XIX.016}.

\end{thebibliography}
